\documentclass[10pt,twocolumn,letterpaper]{article}

\usepackage{cvpr}
\usepackage{times}
\usepackage{epsfig}
\usepackage{graphicx}
\usepackage{amsmath}
\usepackage{amssymb}
\usepackage{booktabs} 
\usepackage[toc,page,header]{appendix}
\usepackage[breaklinks=false,bookmarks=false]{hyperref}

\usepackage[numbers,sort,compress]{natbib}
\usepackage[ruled]{algorithm2e} 
\usepackage{soul} 

\usepackage{commath} 
\usepackage{float}
\usepackage{blindtext}
\usepackage{hyperref}
\usepackage{authblk}

\SetAlFnt{\small}
\SetAlCapFnt{\small}
\SetAlCapNameFnt{\small}
\SetAlCapHSkip{0pt}
\usepackage{amsmath}

\cvprfinalcopy

\begin{document}

\title{Pivotal Tuning for Latent-based Editing of Real Images}

\author[1]{Daniel Roich}
\author[1]{Ron Mokady}
\author[1]{Amit H. Bermano}
\author[1]{Daniel Cohen-Or}
\affil[1]{The Blavatnik School of Computer Science, Tel Aviv University}


\newcommand{\teaserSize}{0.12}

\maketitle

\begin{abstract}
\vspace{-0.695cm}
Recently, a surge of advanced facial editing techniques have been proposed that leverage the generative power of a pre-trained StyleGAN. To successfully edit an image this way, one must first project (or invert) the image into the pre-trained generator's domain. As it turns out, however, StyleGAN's latent space induces an inherent tradeoff between distortion and editability, i.e. between maintaining the original appearance and convincingly altering some of its attributes. Practically, this means it is still challenging to apply ID-preserving facial latent-space editing to faces which are out of the generator's domain.
In this paper, we present an approach to bridge this gap. Our technique slightly alters the generator, so that an out-of-domain image is faithfully mapped into an in-domain latent code. The key idea is \textit{pivotal tuning} --- a brief training process that preserves the editing quality of an in-domain latent region, while changing its portrayed identity and appearance. 
In Pivotal Tuning Inversion (PTI), an initial inverted latent code serves as a pivot, around which the generator is fined-tuned. At the same time, a regularization term keeps nearby identities intact, to locally contain the effect. This surgical training process ends up altering appearance features that represent mostly identity, without affecting editing capabilities.
To supplement this, we further show that pivotal tuning can also adjust the generator to accommodate a multitude of faces, while introducing negligible distortion on the rest of the domain. 
We validate our technique through inversion and editing metrics, and show preferable scores to state-of-the-art methods. We further qualitatively demonstrate our technique by applying advanced edits (such as pose, age, or expression) to numerous images of well-known and recognizable identities.
Finally, we demonstrate resilience to harder cases, including heavy make-up, elaborate hairstyles and/or headwear, which otherwise could not have been successfully inverted and edited by state-of-the-art methods. Source code can be found at: \url{https://github.com/danielroich/PTI}.

\quad
\end{abstract}

\begin{figure}[h]

\centering
\setlength{\tabcolsep}{1pt}
\begin{tabular}{cccc}
Input & Smile & Afro & Pose \\ 
\raisebox{-.5\totalheight}{\includegraphics[width=\teaserSize\textwidth]{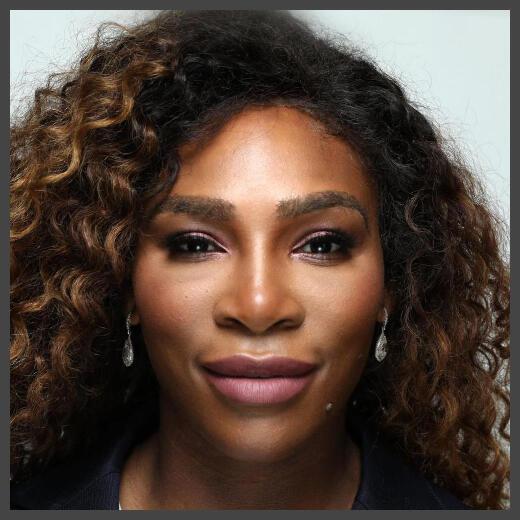}} & 
\raisebox{-.5\totalheight}{\includegraphics[width=\teaserSize\textwidth]{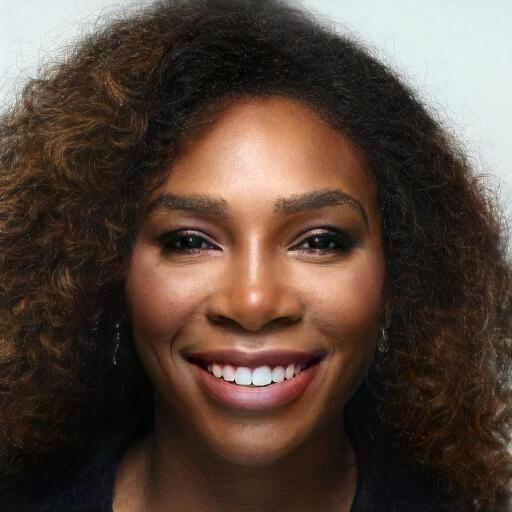}} &
\raisebox{-.5\totalheight}{\includegraphics[width=\teaserSize\textwidth]{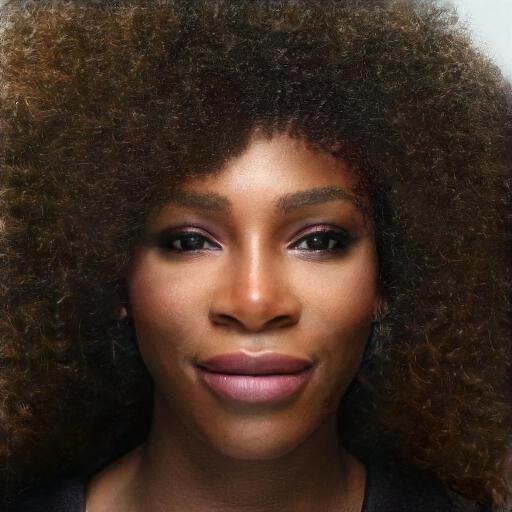}}
&
\raisebox{-.5\totalheight}{\includegraphics[width=\teaserSize\textwidth]{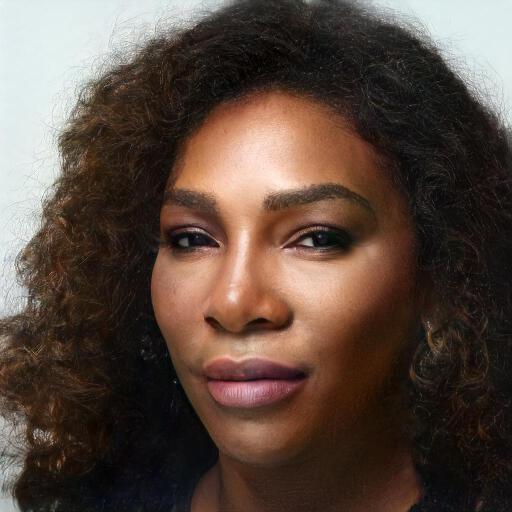}} \\

Input & Age & Mouth & No Beard \\

\raisebox{-.5\totalheight}{\includegraphics[width=\teaserSize\textwidth]{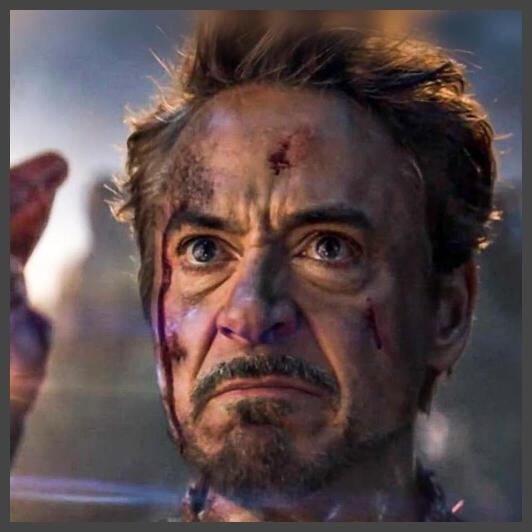}} & 
\raisebox{-.5\totalheight}{\includegraphics[width=\teaserSize\textwidth]{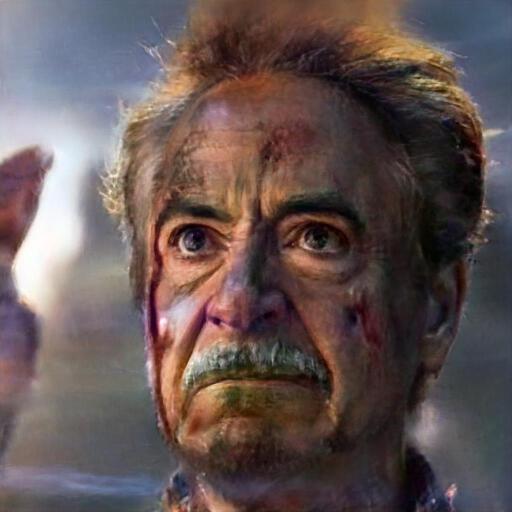}} &
\raisebox{-.5\totalheight}{\includegraphics[width=\teaserSize\textwidth]{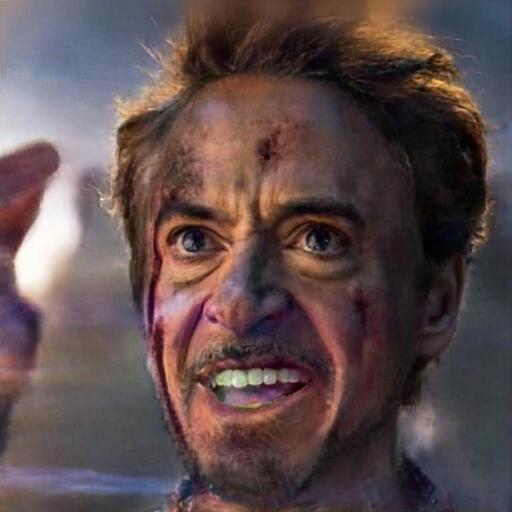}}
&
\raisebox{-.5\totalheight}{\includegraphics[width=\teaserSize\textwidth]{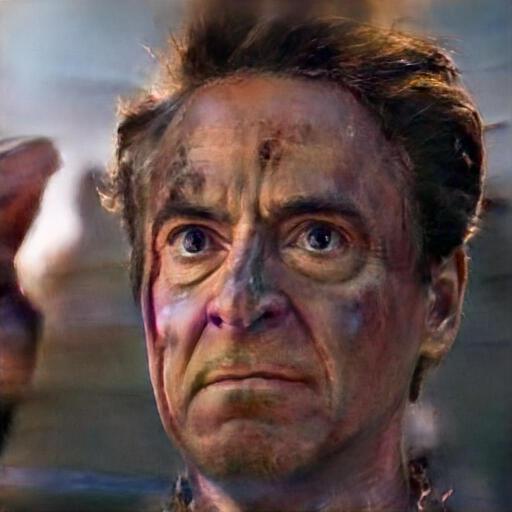}}
\\

Input & Smile & Age & Pose \\
\raisebox{-.5\totalheight}{\includegraphics[width=\teaserSize\textwidth]{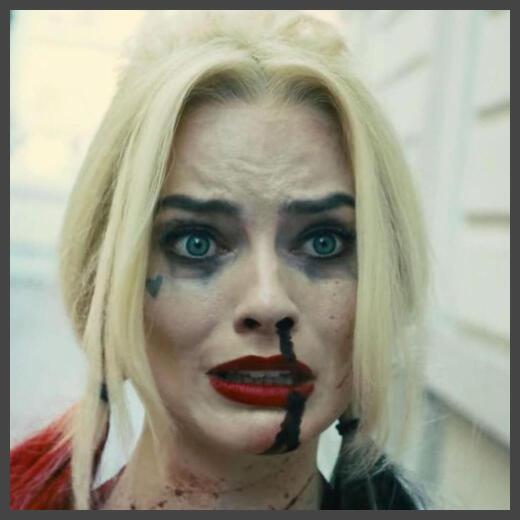}} & 
\raisebox{-.5\totalheight}{\includegraphics[width=\teaserSize\textwidth]{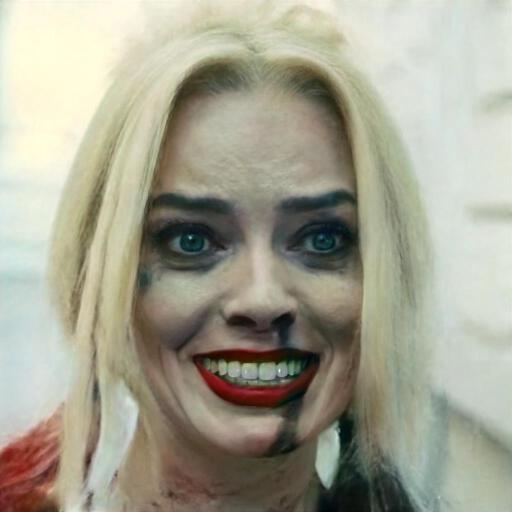}} &
\raisebox{-.5\totalheight}{\includegraphics[width=\teaserSize\textwidth]{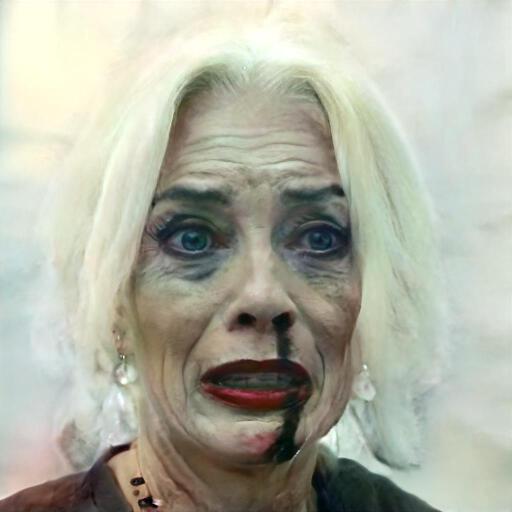}}
&
\raisebox{-.5\totalheight}{\includegraphics[width=\teaserSize\textwidth]{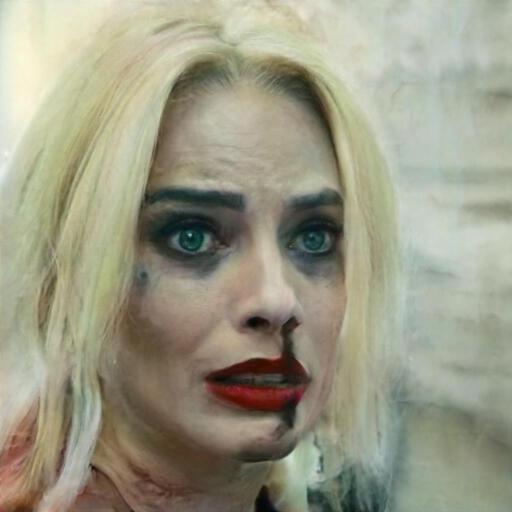}} \\

 Input & Smile & Age & Pose \\
\raisebox{-.5\totalheight}{\includegraphics[width=\teaserSize\textwidth]{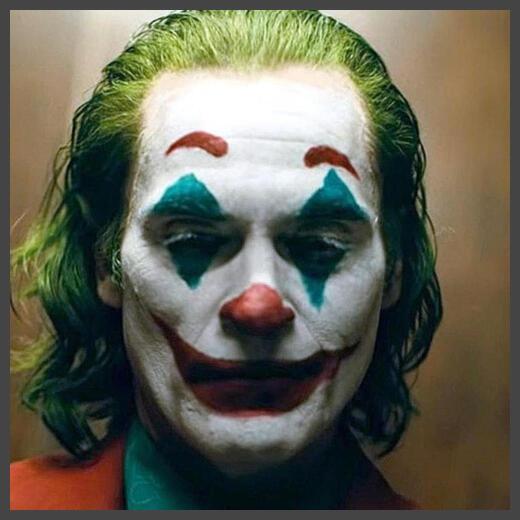}} & 

\raisebox{-.5\totalheight}{\includegraphics[width=\teaserSize\textwidth]{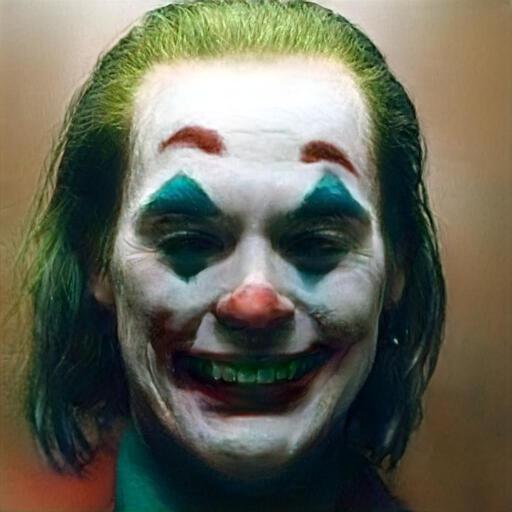}} &
\raisebox{-.5\totalheight}{\includegraphics[width=\teaserSize\textwidth]{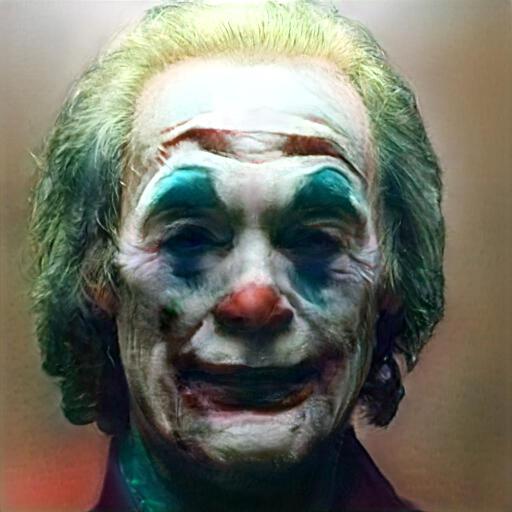}} &
\raisebox{-.5\totalheight}{\includegraphics[width=\teaserSize\textwidth]{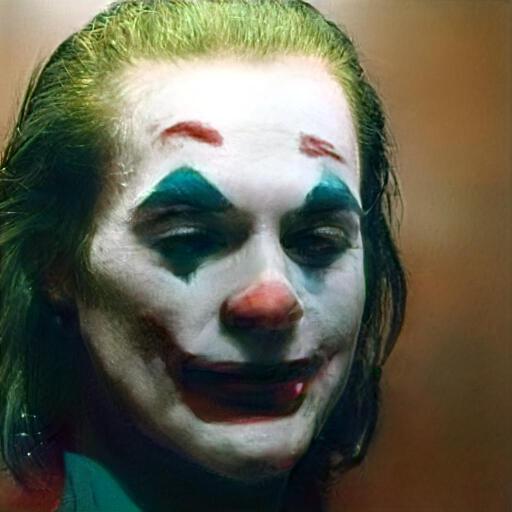}}
\\

\end{tabular}
\vspace{0.1cm}
\caption{Pivotal Tuning Inversion (PTI) enables employing off-the-shelf latent-based semantic editing techniques on real images using StyleGAN. PTI excels in identity preserving edits, portrayed through recognizable figures --- Serena Williams and Robert Downey Jr. (top), and in handling faces which are clearly out-of-domain, e.g., due to heavy makeup (bottom).}

\label{fig:teaser}

\end{figure}

\begin{figure*}

\centering
\begin{tabular}{c}
\raisebox{-.5\totalheight}{\includegraphics[width=.99\textwidth]{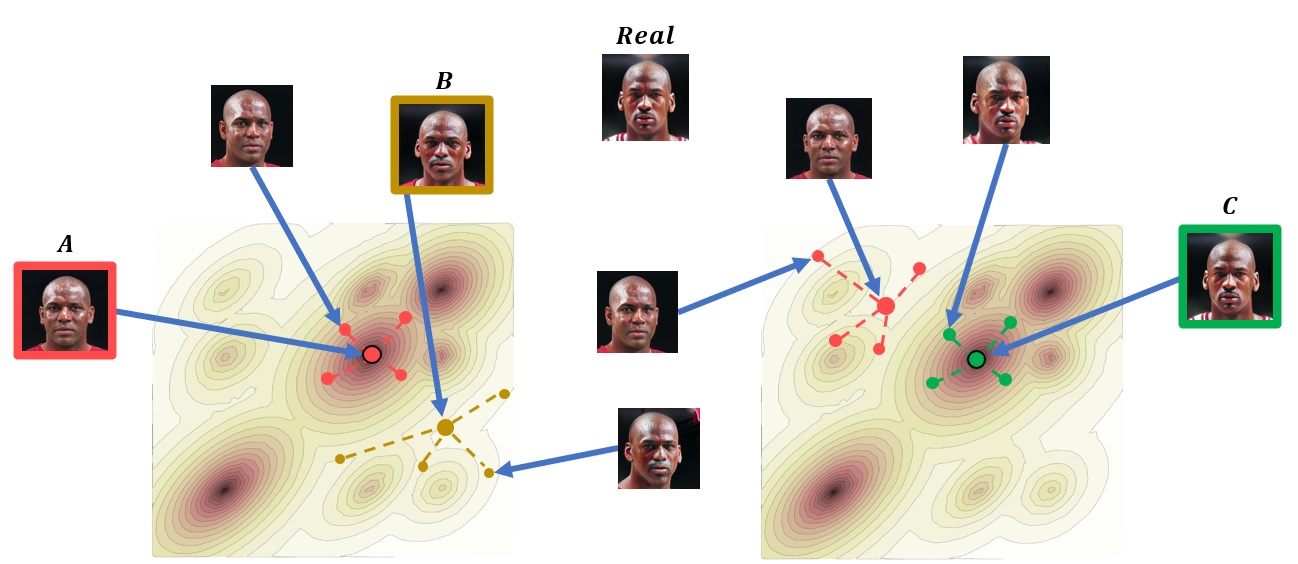}} \\
\hspace{1.0cm}  Before PTI \hspace{7cm} After PTI \hspace{1.0cm} \\
\end{tabular}
\vspace{0.1cm}
\caption{An illustration of the PTI method. StyleGAN's latent space is portrayed in two dimensions (see Tov et al.~\cite{tov2021designing}), where the warmer colors indicate higher densities of $W$, i.e. regions of higher editability. On the left, we illustrate the generated samples before pivotal tuning. We can see the Editability-Distortion trade-off. A choice must be made between Identity $"A"$ and Identity $"B"$. $"A"$ resides in a more editable region but does not resemble the $"Real"$ image. $"B"$ resides in a less editable region, which causes artifacts, but induces less distortion. On the right - After the pivotal tuning procedure. $"C"$ maintains the same high editing capabilities of $"A"$, while achieving even better similarity to $"Real"$ compared to $"B"$.}

\label{fig:illustrate}
\end{figure*}

\section{Introduction}
In recent years, unconditional image synthesis has made huge progress with the emergence of Generative Adversarial Networks (GANs) ~\cite{Goodfellow2014GenerativeAN}. In essence, GANs learn the domain (or manifold) of the desired image set and produce new samples from the same distribution. In particular, StyleGAN ~\cite{karras2019style, karras2020analyzing, Karras2020ada} is one of the most popular choices for this task. Not only does it achieve state-of-the-art visual fidelity and diversity, but it also demonstrates fantastic editing capabilities due to an organically formed disentangled latent space.
Using this property, many methods demonstrate realistic editing abilities over StylGAN's latent space ~\cite{collins2020editing, harkonen2020ganspace, shen2020interpreting, tewari2020stylerig, abdal2020styleflow, wu2020stylespace, patashnik2021styleclip}, such as changing facial orientations, expressions, or age, by traversing the learned manifold.

While impressive, these edits are performed strictly in the generator's latent space, and cannot be applied to real images that are out of its domain. Hence, editing a real image starts with finding its latent representation. This process, called GAN \textit{inversion}, has recently drawn considerable attention \cite{abdal2019image2stylegan, zhu2020domain, menon2020pulse, richardson2020encoding, tov2021designing, alaluf-restyle}.
Early attempts inverted the image to $\mathcal{W}$ --- StyleGAN's native latent space.
However, Abdal et al.~\cite{abdal2019image2stylegan}  have shown that inverting real images to this space results in distortion, i.e. a dissimilarity between the given and generated images, causing artifacts such as identity loss, or an unnatural appearance. 
Therefore, current inversion methods employ an extended latent space, often denoted as $\mathcal{W}+$, which is more expressive and induces significantly less distortion \cite{abdal2019image2stylegan}.

However, even though employing codes from $\mathcal{W}+$ potentially produces great visual quality even for out-of-domain images, these codes suffer from weaker editability, since they are not from the generator's trained domain. Tov et al.~\cite{tov2021designing} define this conflict as the distortion-editability tradeoff, and show that the closer the codes are to $\mathcal{W}$, the better their editability is. Indeed, recent works \cite{zhu2020improved, tov2021designing, alaluf-restyle} suggest a compromise between edibility and distortion, by picking latent codes in $\mathcal{W}+$ which are more editable.

In this paper, we introduce a novel approach to mitigate the distortion-editability trade-off, allowing convincing edits on real images that are out-of-distribution. Instead of projecting the input image into the learned manifold, we augment the manifold to include the image by slightly altering the generator, in a process we call \textit{pivotal tuning}. This adjustment is analogous to shooting a dart and then shifting the board itself to compensate for a near hit.

Since StyleGAN training is expensive and the generator achieves unprecedented visual quality, the popular approach is to keep the generator frozen. In contrast, we propose producing a personalized version of the generator, that accommodates the desired input image or images. Our approach consists of two main steps. First, we invert the input image to an editable latent code, using off-the-shelf inversion techniques. This, of course, yields an image that is similar to the original, but not necessarily identical. In the second step, we perform  \textit{Pivotal Tuning} --- we lightly tune the pretrained StyleGAN, such that the input image is generated when using the pivot latent code found in the previous step (see Figure ~\ref{fig:illustrate} for an illustration.). The key idea is that even though the generator is slightly modified, the latent code keeps its editing qualities. As can be seen in our experiments, the modified generator retains the editing capabilities of the pivot code, while achieving unprecedented reconstruction quality. As we demonstrate, the pivotal tuning is a \textit{local} operation in the latent space, shifting the identity of the pivotal region to the desired one with minimal repercussions. To minimize side-effects even further, we introduce a regularization term, enforcing only a surgical adaptation of the latent space. This yields a version of the generator StyleGAN that can edit multiple target identities without interference.

In essence, our method extends the high quality editing capabilities of the pretrained StyleGAN to images that are out of its distribution, as demonstrated in Figure~\ref{fig:teaser}. We validate our approach through quantitative and qualitative results, and demonstrate that our method achieves state-of-the-art results for the task of StyleGAN inversion and real image editing. In Section~\ref{sec:experiments}, we show that not only do we achieve better reconstruction, but also superior editability. 
We show this through the utilization of several existing editing techniques, and achieve realistic editing even on challenging images. Furthermore, we confirm that using our regularization restricts the pivotal tuning side effect to be local, with negligible effect on distant latent codes, and that pivotal tuning can be applied for multiple images simultaneously to incorporate several identities into the same model (see Figure \ref{fig:leaders_multi_id1}). Finally, we show through numerous challenging examples that our pivotal tuning-based inversion approach achieves completely automatic, fast, faithful, and powerful editing capabilities.

\section{Related Work}

\subsection{Latent Space Manipulation}

Most real-life applications require control over the generated image. Such control can be obtained in the unconditional setting, by first learning the manifold, and then realizing image editing through latent space traversal. Many works have examined semantic directions in the latent spaces of pre-trained GANs. Some using full-supervision in the form of semantic labels \cite{denton2019detecting, goetschalckx2019ganalyze, shen2020interpreting}, others ~\cite{jahanian2019steerability, spingarn2020gan, plumerault2020controlling} find meaningful directions in a self-supervised fashion, and finally recent works present unsupervised methods to achieve the same goal \cite{voynov2020unsupervised, harkonen2020ganspace, wang2021a}, requiring no manual annotations.

More specifically for StyleGAN, Shen et al.~\cite{shen2020interpreting} use supervision in the form of facial attribute labels to find meaningful linear directions in the latent space. Similar labels are used by Abdal et al.~\cite{abdal2020styleflow} to train a mapping network conditioned on these labels. Harkonen et al.~\cite{harkonen2020ganspace} identify latent directions based on Principal Component Analysis (PCA). Shen et al.~\cite{shen2020closedform} perform eigenvector decomposition on the generator’s weights to find edit directions without additional supervision. Collins et al. ~\cite{collins2020editing} borrow parts of the latent code of other samples to produce local and semantically aware edits.  Wu et al. \cite{wu2020stylespace} discover disentangled editing controls in the space of channel wise style parameters. 
Other works \cite{tewari2020pie, tewari2020stylerig} focus on facial editing, as they utilize a prior in the form of a 3D morphable face model.
Most recently, Patashnik et al. ~\cite{patashnik2021styleclip} utilize a contrastive language-image pre-training (CLIP) models ~\cite{radford2021learning} to explore new editing capabilities. 
In this paper, we demonstrate our inversion approach by utilizing these editing methods as downstream tasks. As seen in Section~\ref{sec:experiments}, our PTI process induces higher visual quality for several of these popular approaches.

\subsection{GAN inversion}

As previously mentioned, in order to edit a real image using latent manipulation, one must perform \textit{GAN inversion} ~\cite{zhu2016generative}, meaning one must find a latent vector from which the generator would generate the input image.  Inversion methods can typically be divided into optimization-based ones --- which directly optimize the latent code using a single sample ~\cite{lipton2017precise, creswell2018inverting, abdal2019image2stylegan, karras2020analyzing}, or encoder-based ones --- which train an encoder over a large number of samples ~\cite{perarnau2016invertible, luo2017learning, guan2020collaborative}. 
Many works consider specifically the task of StyleGAN inversion, aiming at leveraging the high visual quality and editability of this generator. 
Abdal et al.~\cite{abdal2020image2stylegan++} demonstrate that it is not feasible to invert images to StyleGAN's native latent space $\mathcal{W}$ without significant artifacts. Instead, it has been shown that the extended $\mathcal{W}+$ is much more expressive, and enables better image preservation. Menon et al.~\cite{menon2020pulse} use direct optimization for the task of super-resolution by inverting a low-resolution image to $\mathcal{W+}$ space. Zhu et al.~\cite{zhu2020domain} use a hybrid approach: first, an encoder is trained, then a direct optimization is performed. Richardson et al.~\cite{richardson2020encoding} were the first to train an encoder for $\mathcal{W+}$ inversion which was demonstrated to solve a variety of image-to-image translation tasks.

\subsection{Distortion-editability tradeoff}

Even though $\mathcal{W+}$ inversion achieves minimal distortion, it has been shown that the results of latent manipulations over $\mathcal{W+}$ inversions are inferior compared to the same manipulations over latent codes from StyleGAN's native space $\mathcal{W}$. Tov et al.~\cite{tov2021designing} define this as the distortion-editability tradeoff, and design an encoder that attempts to find a "sweet-spot" in this trade-off.    

Similarly, the tradeoff was also demonstrated by Zhu et al.~\cite{zhu2020improved}, who suggests an improved embedding algorithm using a novel regularization method. StyleFlow~\cite{abdal2020styleflow} also concludes that real image editing produces significant artifacts compared to images generated by StyleGAN. Both Zhu et al. and Tov et al. achieve better editability compared to previous methods but also suffer from more distortion. In contrast, our method combines the editing quality of $\mathcal{W}$ inversions with highly accurate reconstructions, thus mitigating the distortion-editability tradeoff.

\subsection{Generator Tuning}
 
Typically, editing methods avoid altering StyleGAN, in order to preserve its excellent performance. Some works, however, do take the approach we adopt as well, and tune the generator. Pidhorskyi et al.~\cite{pidhorskyi2020adversarial} train both the encoder and the StyleGAN generator, but their reconstruction results suffer from significant distortion, as the StyleGAN tuning step is too extensive. Bau et al.~\cite{bau2020semantic} propose a method for interactive editing which tunes the generator proposed by Karras et al.~\cite{karras2017progressive} to reconstruct the input image. They claim, however, that directly updating the weights results in sensitivity to small changes in the input, which induces unrealistic artifacts. In contrast, we show that after directly updating the weights, our generator keeps its editing capabilities, and demonstrate this over a variety of editing techniques. Pan et al.~\cite{pan2020exploiting} invert images to BigGAN's ~\cite{brock2018large} latent space by optimizing a random noise vector and tuning the generator simultaneously. Nonetheless, as we demonstrate in Section~\ref{sec:experiments}, optimizing a random vector decreases reconstruction and editability quality significantly for StyleGAN.

\begin{figure*}[h]
\centering
\begin{tabular}{ccccccc}
\rotatebox[origin=t]{90}{Real Image} & \raisebox{-.5\totalheight}{\includegraphics[width=0.13\textwidth]{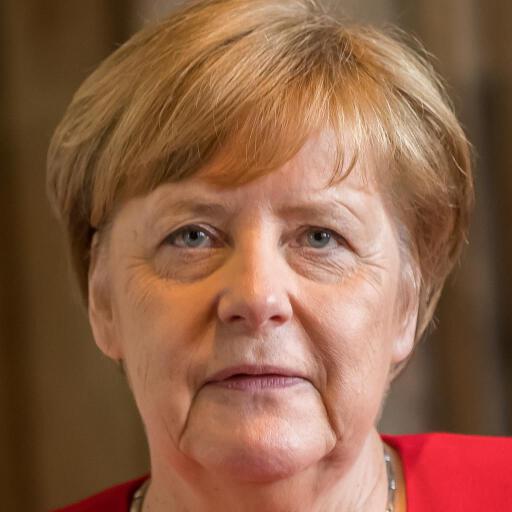}} &
\raisebox{-.5\totalheight}{\includegraphics[width=0.13\textwidth]{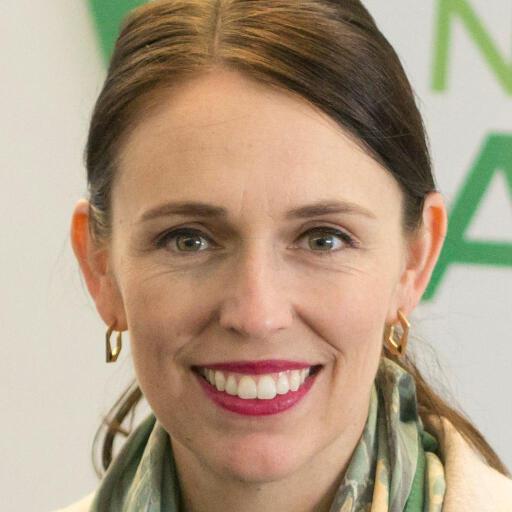}} & 
\raisebox{-.5\totalheight}{\includegraphics[width=0.13\textwidth]{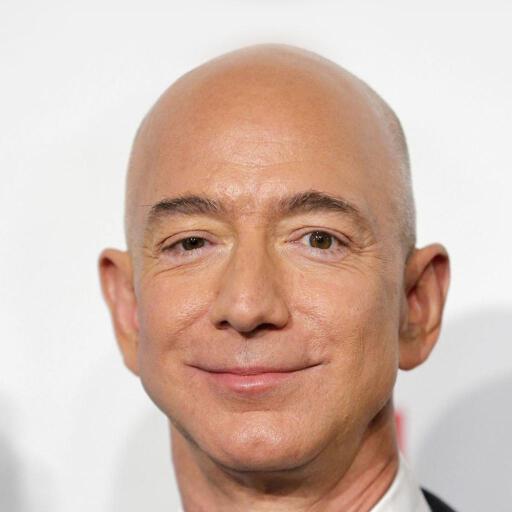}} & 
\raisebox{-.5\totalheight}{\includegraphics[width=0.13\textwidth]{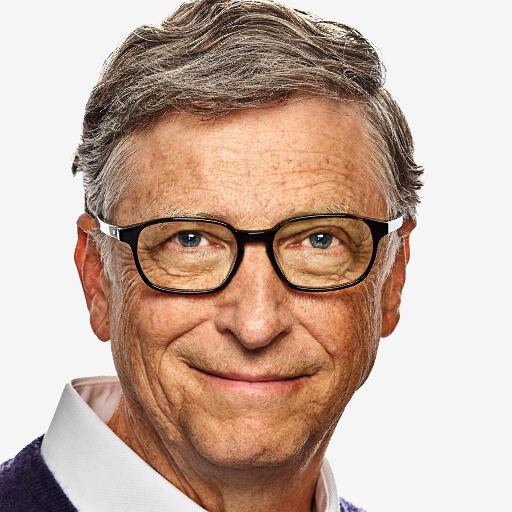}} & 
\raisebox{-.5\totalheight}{\includegraphics[width=0.13\textwidth]{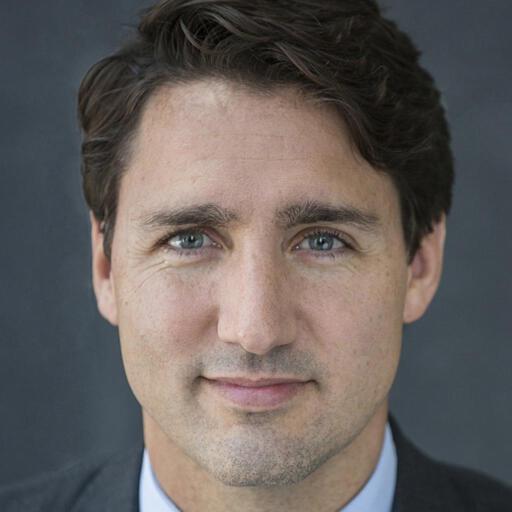}} &
\raisebox{-.5\totalheight}{\includegraphics[width=0.13\textwidth]{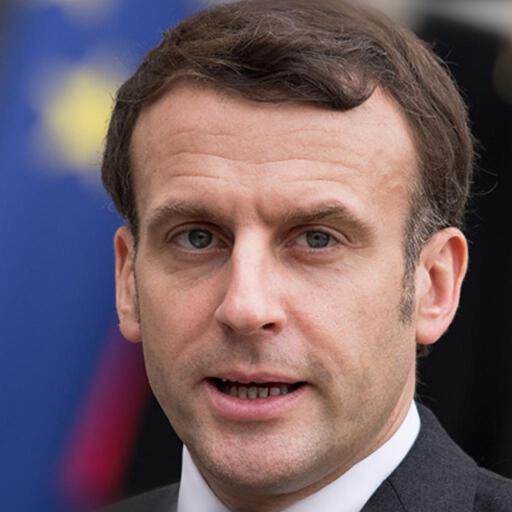}} \\
\noalign{\vskip 1mm}
\rotatebox[origin=t]{90}{Inversion} &
\raisebox{-.5\totalheight}{\includegraphics[width=0.13\textwidth]{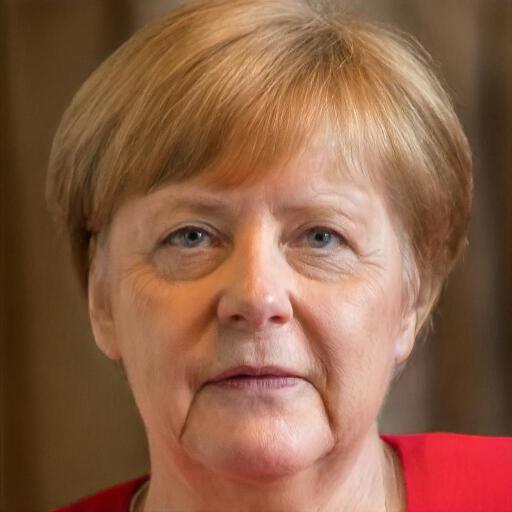}} &
\raisebox{-.5\totalheight}{\includegraphics[width=0.13\textwidth]{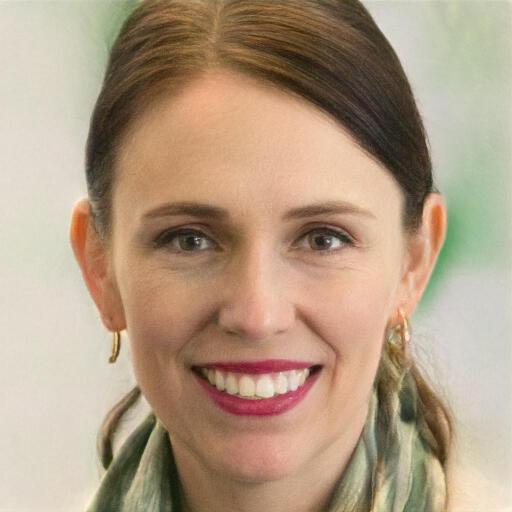}} & 
\raisebox{-.5\totalheight}{\includegraphics[width=0.13\textwidth]{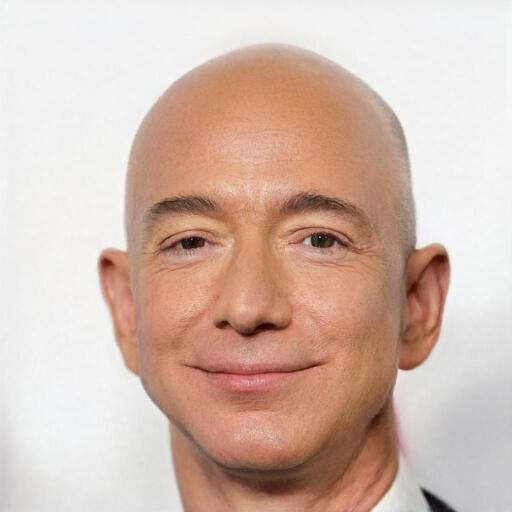}} & 
\raisebox{-.5\totalheight}{\includegraphics[width=0.13\textwidth]{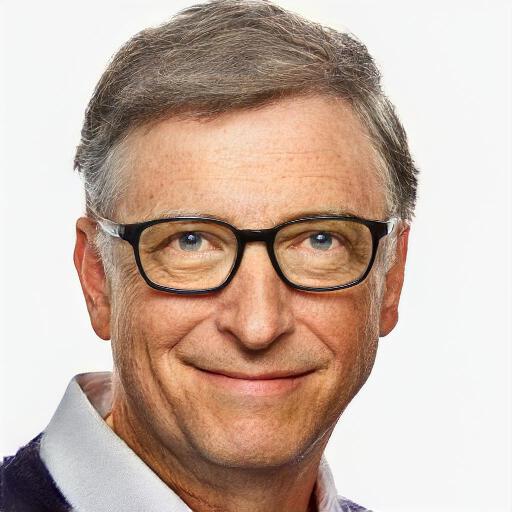}} & 
\raisebox{-.5\totalheight}{\includegraphics[width=0.13\textwidth]{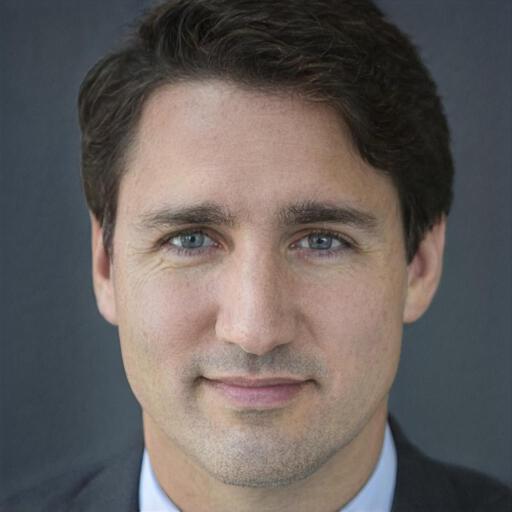}} &
\raisebox{-.5\totalheight}{\includegraphics[width=0.13\textwidth]{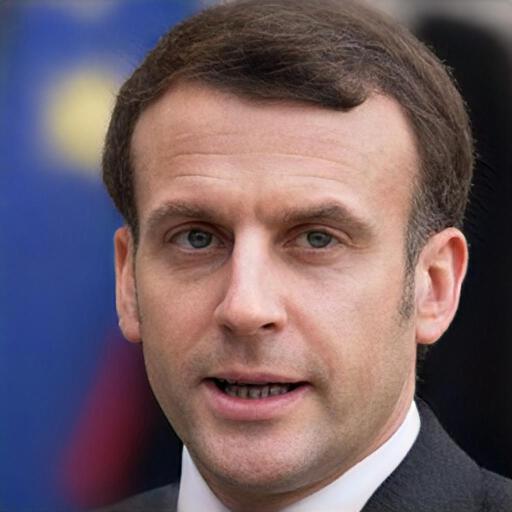}} \\
\noalign{\vskip 0.1cm}
\rotatebox[origin=t]{90}{$\pm$Age} &
\raisebox{-.5\totalheight}{\includegraphics[width=0.13\textwidth]{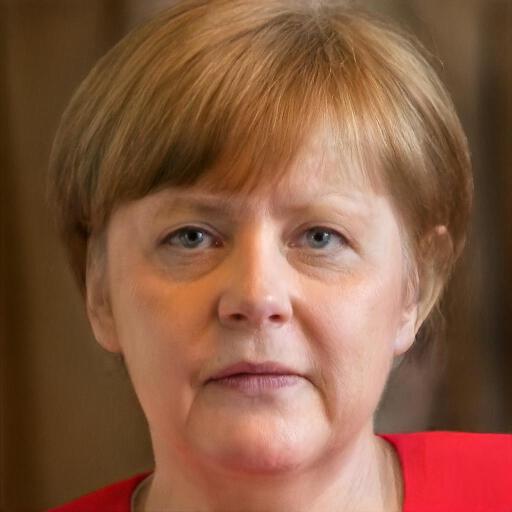}} &
\raisebox{-.5\totalheight}{\includegraphics[width=0.13\textwidth]{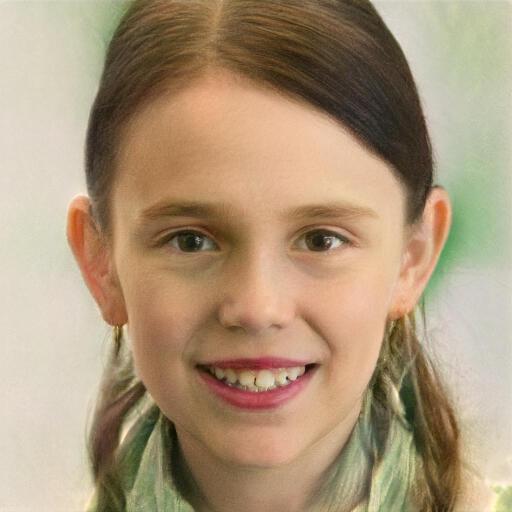}} & 
\raisebox{-.5\totalheight}{\includegraphics[width=0.13\textwidth]{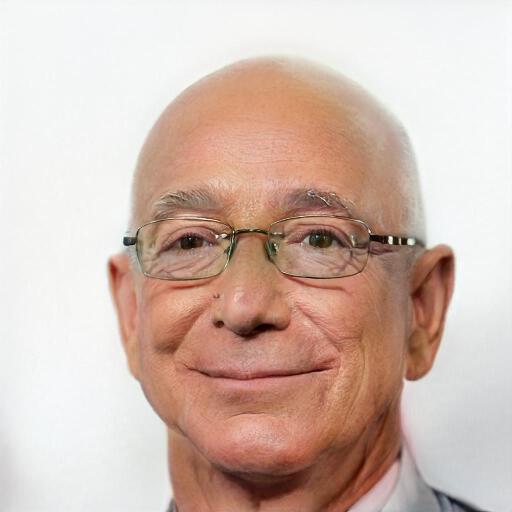}} & 
\raisebox{-.5\totalheight}{\includegraphics[width=0.13\textwidth]{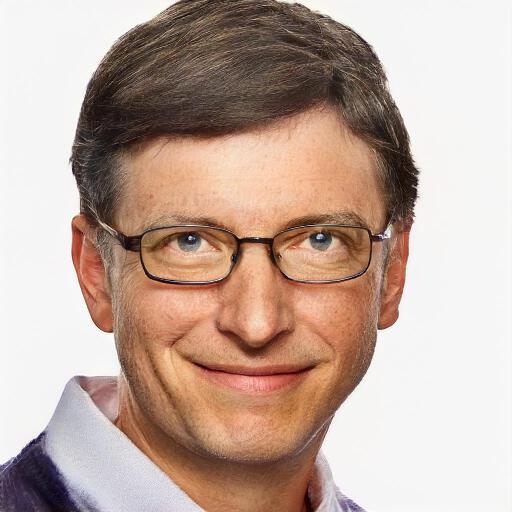}} & 
\raisebox{-.5\totalheight}{\includegraphics[width=0.13\textwidth]{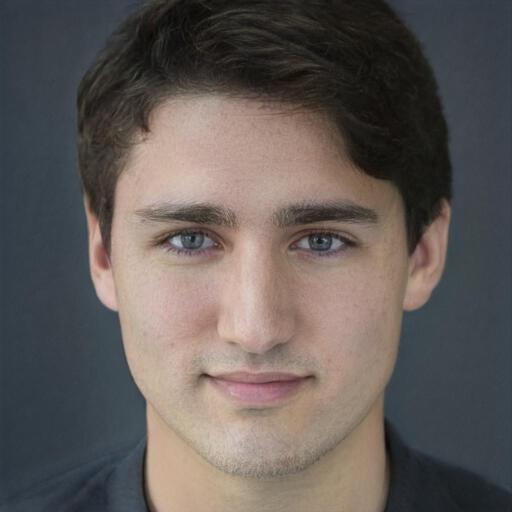}} &
\raisebox{-.5\totalheight}{\includegraphics[width=0.13\textwidth]{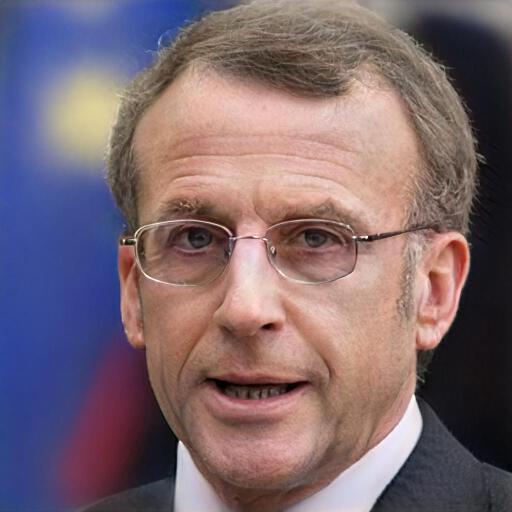}} \\
\noalign{\vskip 0.1cm}
\rotatebox[origin=t]{90}{$\pm$Smile} &
\raisebox{-.5\totalheight}{\includegraphics[width=0.13\textwidth]{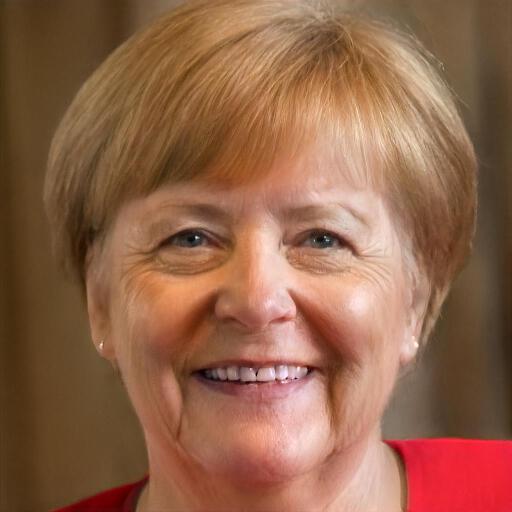}} &
\raisebox{-.5\totalheight}{\includegraphics[width=0.13\textwidth]{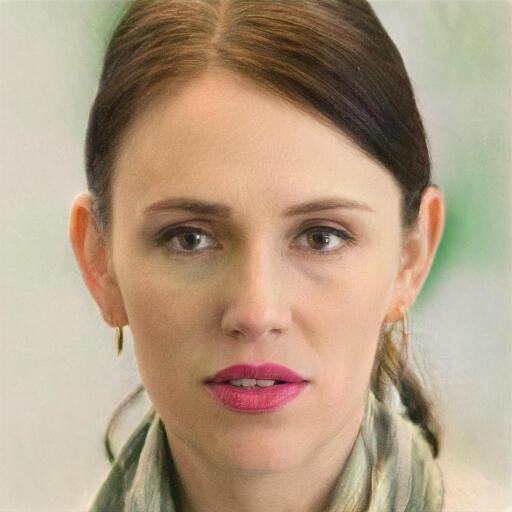}} & 
\raisebox{-.5\totalheight}{\includegraphics[width=0.13\textwidth]{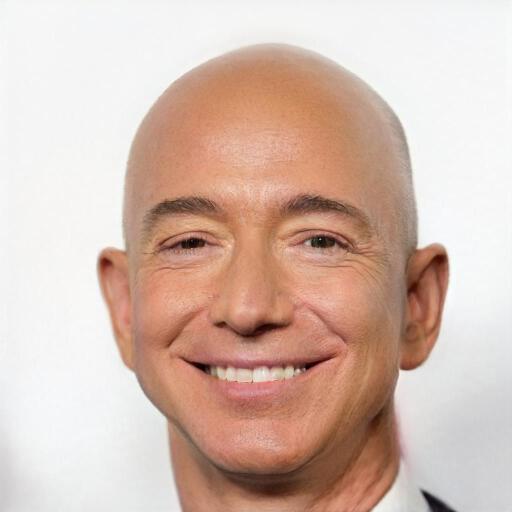}} & 
\raisebox{-.5\totalheight}{\includegraphics[width=0.13\textwidth]{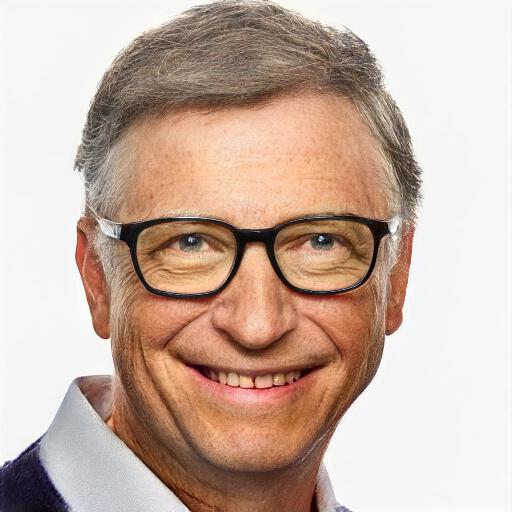}} & 
\raisebox{-.5\totalheight}{\includegraphics[width=0.13\textwidth]{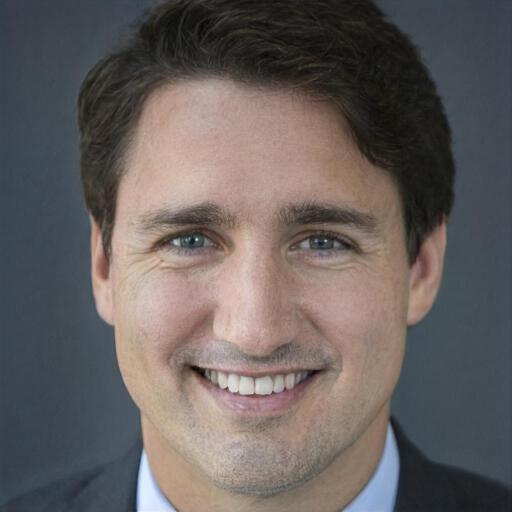}} &
\raisebox{-.5\totalheight}{\includegraphics[width=0.13\textwidth]{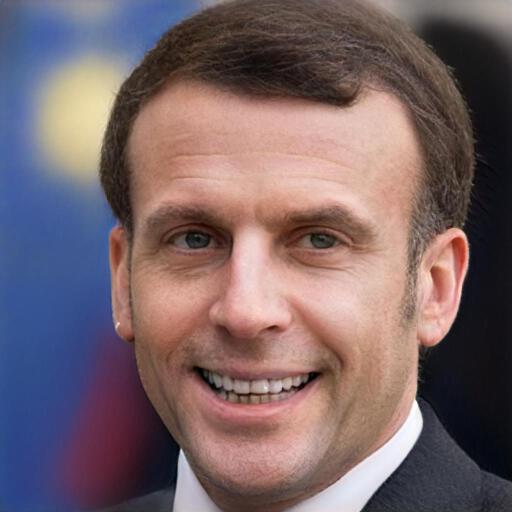}} \\
\noalign{\vskip 0.1cm}
\rotatebox[origin=t]{90}{Rotation} &
\raisebox{-.5\totalheight}{\includegraphics[width=0.13\textwidth]{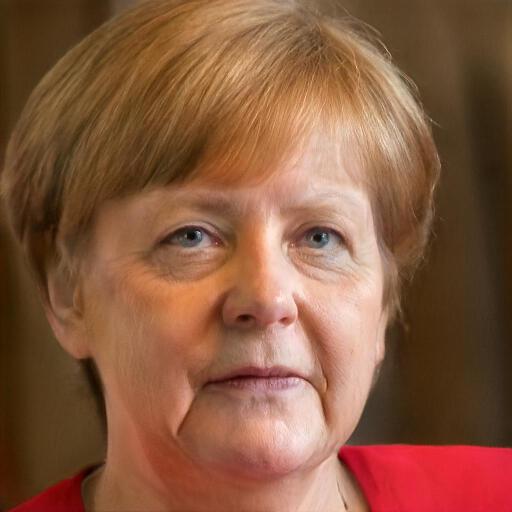}} &
\raisebox{-.5\totalheight}{\includegraphics[width=0.13\textwidth]{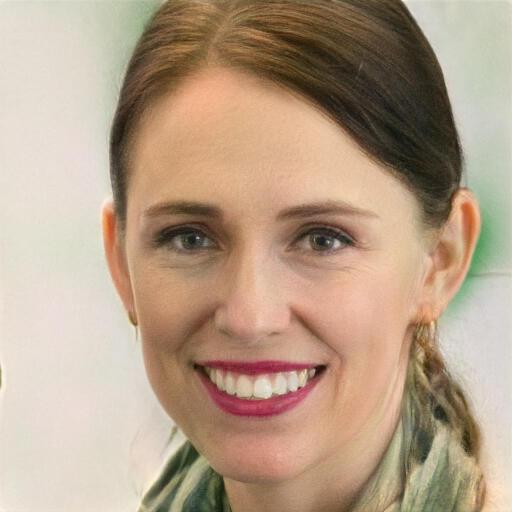}} & 
\raisebox{-.5\totalheight}{\includegraphics[width=0.13\textwidth]{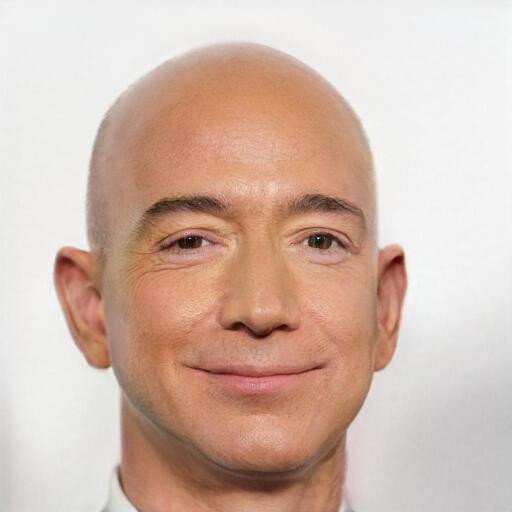}} & 
\raisebox{-.5\totalheight}{\includegraphics[width=0.13\textwidth]{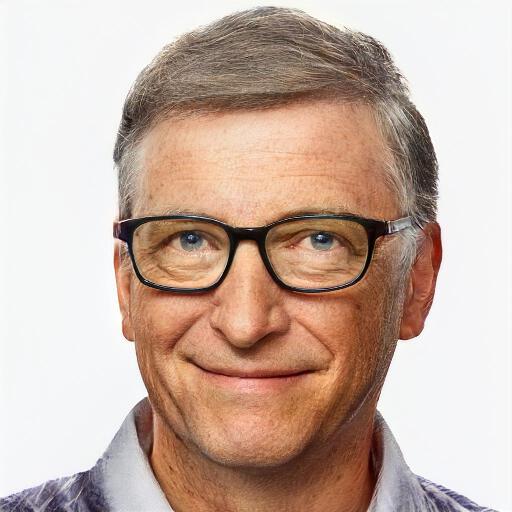}} & 
\raisebox{-.5\totalheight}{\includegraphics[width=0.13\textwidth]{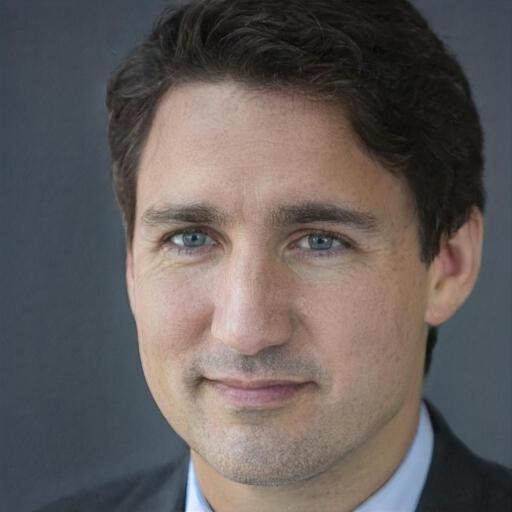}} &
\raisebox{-.5\totalheight}{\includegraphics[width=0.13\textwidth]{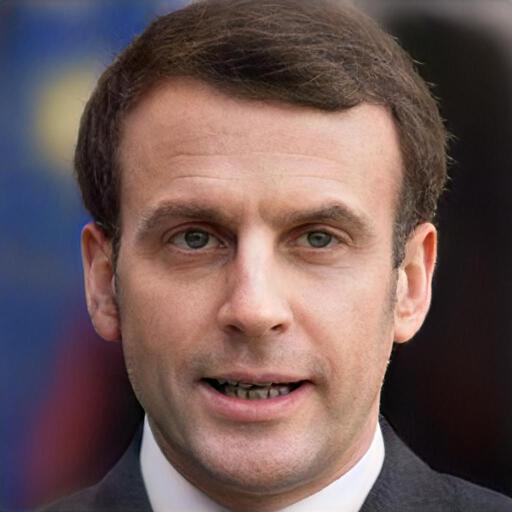}} \\
\end{tabular}
\vspace{0.2cm}
\caption{Real Images editing example using a Multi-ID Personalized StyleGAN. All depicted images are generated by the same model, fine-tuned on political and industrial world leaders. As can be seen, applying various edit operations on these newly introduced, highly recognizable identities preserves them well.}
\label{fig:leaders_multi_id1}
\end{figure*}

\section{Method}

Our method seeks to provide high quality editing for a real image using StyleGAN. The key idea of our approach is that due to StyleGAN's disentangled nature, slight and local changes to its produced appearance can be applied without damaging its powerful editing capabilities. Hence, given an image, possibly is out-of-distribution in terms of appearance (e.g., real identities, extreme lighting conditions, heavy makeup, and/or extravagant hair and headwear), we propose finding its closest editable point within the generator's domain. This pivotal point can then be pulled toward the target, with only minimal effect in its neighborhood, and negligible effect elsewhere.
In this section, we present a two-step method for inverting real images to highly editable latent codes. First, we invert the given input to $w_p$ in the native latent space of StyleGAN, $\mathcal{W}$. Then, we apply a \textit{Pivotal Tuning} on this \textit{pivot code} $w_p$ to tune the pretrained StyleGAN to produce the desired image for input $w_p$. The driving intuition here is that since $w_p$ is close enough, training the generator to produce the input image from the pivot can be achieved through augmenting appearance-related weights only, without affecting the well-behaved structure of StyleGAN's latent space. 

\subsection{Inversion}

The purpose of the inversion step is to provide a convenient starting point for the Pivotal Tuning one (Section~\ref{sec:pivotalTuning}). As previously stated, StyleGAN's native latent space $\mathcal{W}$ provides the best editability. Due to this and since the distortion is diminished during Pivotal Tuning, we opted to invert the given input image $x$ to this space, instead of the more popular $\mathcal{W}+$ extension. We use an off-the-shelf inversion method, as proposed by Karras et al.~\cite{karras2020analyzing}. In essence, a direct optimization is applied to optimize both latent code $w$ and noise vector $n$ to reconstruct the input image $x$, measured by the LPIPS perceptual loss function \cite{zhang2018unreasonable}. As described in \cite{karras2020analyzing}, optimizing the noise vector $n$ using a noise regularization term improves the inversion significantly, as the noise regularization prevents the noise vector from containing vital information. This means that once $w_p$ has been determined, the $n$ values play a minor role in the final visual appearance. Overall, the optimization defined as the following objective:
\begin{align}
w_p, n = \underset{w, n}{\arg\min} \mathcal{L}_{\text{LPIPS}}(x, G(w, n; \theta)) + \lambda_{n}\mathcal{L}_{n}(n),
\end{align}
where $G(w, n, \theta)$ is the generated image using a generator $G$ with weights $\theta$. Note that we do not use StyleGAN's mapping network (converting from $\mathcal{Z}$ to $\mathcal{W}$). $\mathcal{L}_{\text{LPIPS}}$ denotes the perceptual loss, $\mathcal{L}_{n}$ is a noise regularization term and $\lambda_{n}$ is a hyperparameter. At this step, the generator remains frozen.

\subsection{Pivotal Tuning}
\label{sec:pivotalTuning}

Applying the latent code $w$ obtained in the inversion, produces an image that is similar to the original one $x$, but may yet exhibit significant distortion. Therefore, in the second step, we unfreeze the generator and tune it to reconstruct the input image $x$ given the latent code $w$ obtained in the first step, which we refer to as the \textit{pivot} code $w_p$. As we demonstrate in Section~\ref{sec:experiments}, it is crucial to use the pivot code, since using random or mean latent codes lead to unsuccessful convergence.
Let $x^{p} = G(w_p; \theta^*)$ be the generated image using $w_p$ and the tuned weights $\theta^*$. We fine tune the generator using the following loss term:

\begin{align}
\mathcal{L}_{pt} =  \mathcal{L}_{\text{LPIPS}}(x, x^{p}) + \lambda_{L2}\mathcal{L}_{L2}(x, x^{p}),
\end{align}
where the generator is initialized with the pretrained weights $\theta$. At this step, $w_p$ is constant. The pivotal tuning can trivially be extended to $N$ images $\{x_i\}_{i=0}^{N}$, given the $N$ inversion latent codes $\{w_i\}_{i=0}^{N}$:

\begin{align}
\mathcal{L}_{pt} = \frac{1}{N}\sum^{N}_{i=1} ( \mathcal{L}_{\text{LPIPS}}(x_i, x^{p}_i) + \lambda_{L2}\mathcal{L}_{L2}(x_i, x^{p}_i)), 
\end{align}

where $x^{p}_i = G(w_i; \theta^*)$.

Once the generator is tuned, we can edit the input image using any choice of latent-space editing techniques, such as those proposed by Shen et al. ~\cite{shen2020interpreting} or Harkonen et al. ~\cite{harkonen2020ganspace}. Numerous results are demonstrated in Section~\ref{sec:experiments}.

\subsection{Locality Regularization}

As we demonstrate in Section~\ref{sec:experiments}, applying pivotal tuning on a latent code indeed brings the generator to reconstruct the input image in high accuracy, and even enables successful edits around it. At the same time, as we demonstrate in Section~\ref{sec:reg}, Pivotal tuning induces a ripple effect --- the visual quality of images generated by non-local latent codes is compromised. This is especially true when tuning for a multitude of identities (see Figure.~\ref{fig:reg_visual}).   
To alleviate this side effect, we introduce a regularization term, that is designed to restrict the PTI changes to a local region in the latent space. In each iteration, we sample a normally distributed random vector $z$ and use StyleGAN's mapping network $f$ to produce a corresponding latent code $w_z = f(z)$. Then, we interpolate between $w_z$ and the pivotal latent code $w_p$ using the interpolation parameter $\alpha$, to obtain the interpolated code $w_r$:
\begin{align}
w_r = w_p + \alpha\frac{w_z - w_p}{\norm{w_z - w_p}_2}.
\end{align}

Finally, we minimize the distance between the image generated by feeding $w_r$ as input using the original weights $x_r = G(w_r; \theta)$ and the image generated using the currently tuned ones $x_r^* = G(w_r; \theta^*)$:
\begin{align}
\mathcal{L}_{R} =  \mathcal{L}_{\text{LPIPS}}(x_r,x_r^*) + \lambda^R_{{L2}}\mathcal{L}_{L2}(x_r, x_r^*).
\end{align}
This can be trivially extended to $N_r$ random latent codes:
\begin{align}
\mathcal{L}_{R} = \frac{1}{N_r}\sum^{N_r}_{i=1} (\mathcal{L_{\text{LPIPS}}}(x_{r,i},x_{r,i}^*) + \lambda^R_{L2}\mathcal{L}_{L2}(x_{r,i}, x_{r,i}^*)).
\end{align}
The new optimization is defined as:
\begin{align}
\theta^* = \underset{\theta^*}{\arg\min} \mathcal{L}_{pt} + \lambda_R \mathcal{L}_{R},
\end{align}
where $\lambda^R_{L2}$, $\lambda_R$, $N_r$ are constant positive hyperparameters. Additional discussion regarding the effects of different $\alpha$ values can be found in the Supplementary Materials.

\begin{figure}

\setlength{\tabcolsep}{1pt}
\begin{tabular}{l}

\hspace{0.1cm}  \textbf{Original}  \hspace{0.25cm} \textbf{SG2 $\mathcal{W+}$} \hspace{0.45cm} \textbf{e4e}  \hspace{0.9cm} \textbf{SG2} \hspace{0.85cm} \textbf{Ours} \\

\raisebox{-.5\totalheight}{\includegraphics[width=\columnwidth]{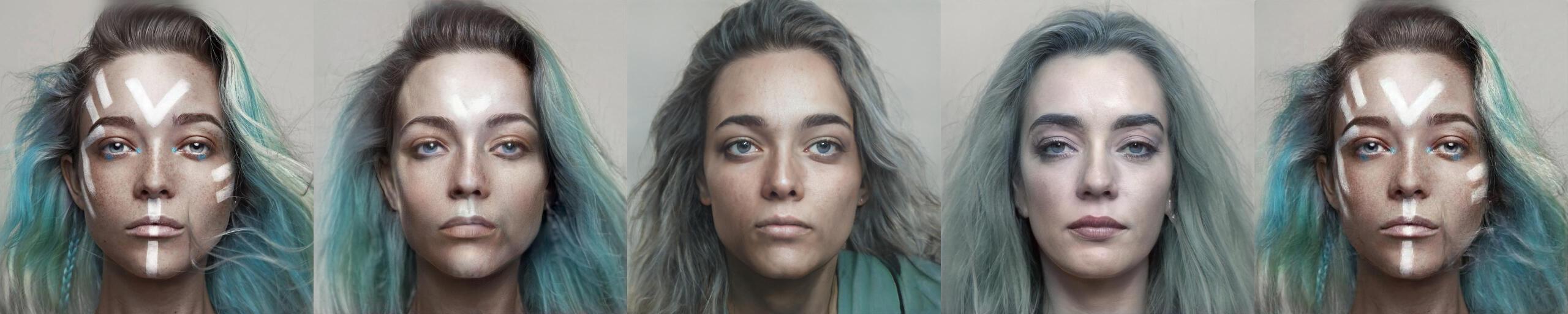}} \\
\noalign{\vskip 1mm}

\raisebox{-.5\totalheight}{\includegraphics[width=\columnwidth]{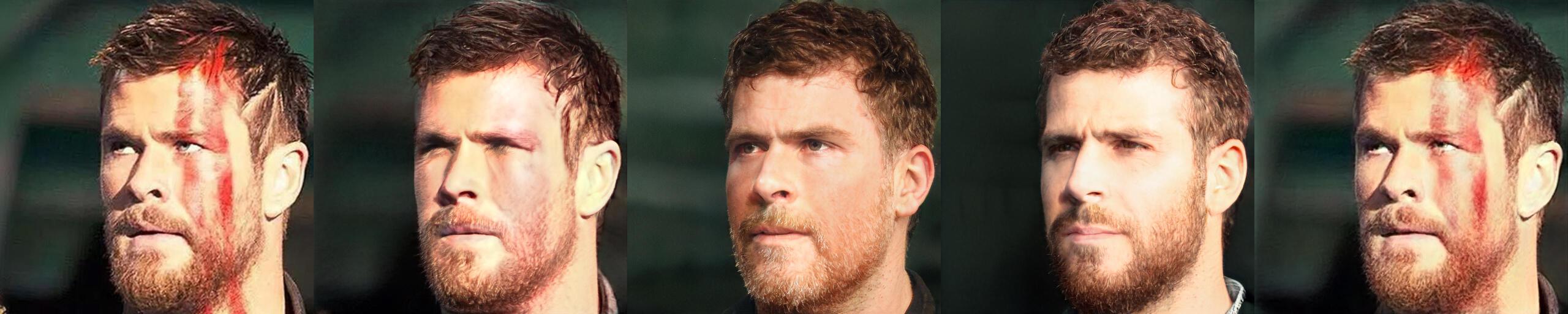}} \\
\noalign{\vskip 1mm}

\raisebox{-.5\totalheight}{\includegraphics[width=\columnwidth]{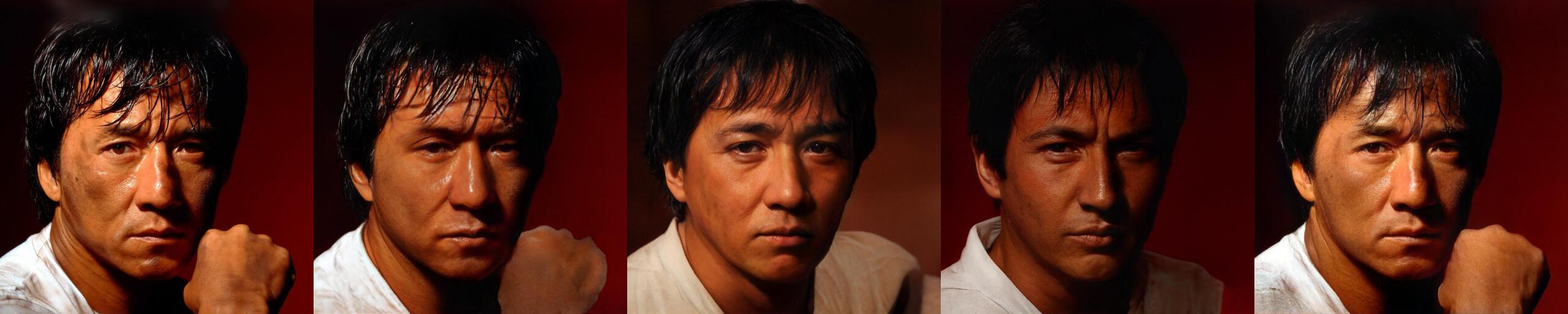}} \\

\end{tabular}
\vspace{0.1cm}
\caption{Reconstruction of out-of-domain samples. Our method (right) reconstructs out-of-domain visual details (left), such as face paintings or hands, significantly better than state-of-the-art methods (middle).}
\label{fig:rec_out}
\end{figure}

\begin{figure}

\setlength{\tabcolsep}{1pt}
\begin{tabular}{l}
\hspace{0.1cm}  \textbf{Original}  \hspace{0.25cm} \textbf{SG2 $\mathcal{W+}$} \hspace{0.45cm} \textbf{e4e}  \hspace{0.9cm} \textbf{SG2} \hspace{0.85cm} \textbf{Ours} \\
\raisebox{-.5\totalheight}{\includegraphics[width=\columnwidth]{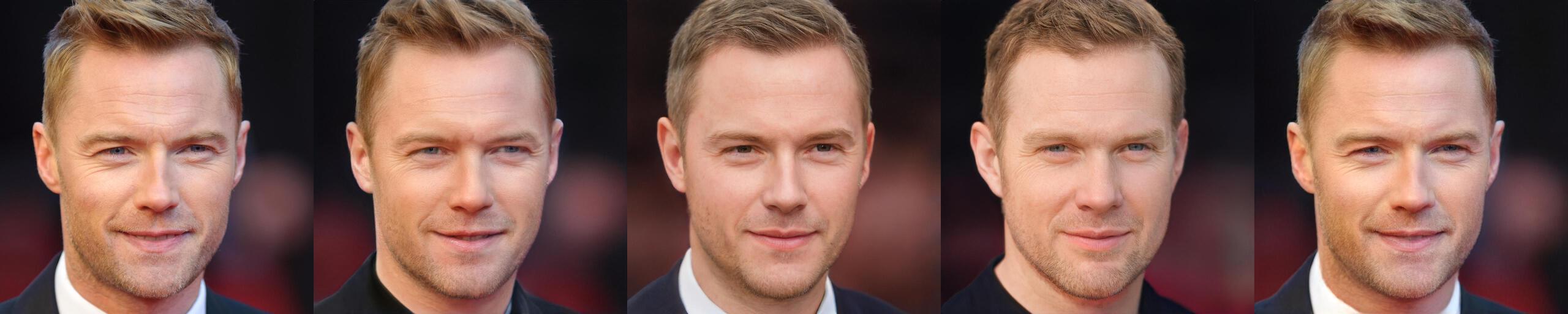}} \\
\noalign{\vskip 1mm}
\raisebox{-.5\totalheight}{\includegraphics[width=\columnwidth]{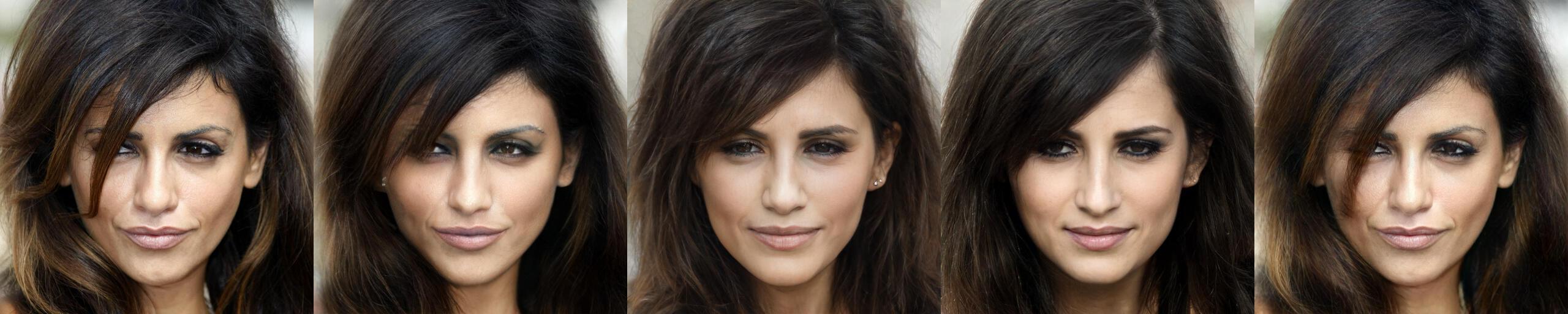}} \\
\end{tabular}
\vspace{0.1cm}
\caption{Reconstruction quality comparison using examples from the CelebA-HQ Dataset. As can be seen, even for less challenging inputs, our method offers higher level reconstruction for unseen identities compared to the state-of-the-art. Zoom-in recommended.}
\label{fig:rec_celebA}
\end{figure}

\section{Experiments}
\label{sec:experiments}

In this section, we justify the design choices made and evaluate our method. For all experiments we use the StyleGAN$2$ generator ~\cite{karras2020analyzing}. For facial images, we use a generator pre-trained over the FFHQ dataset~\cite{karras2019style}, and we use the CelebA-HQ dataset~\cite{liu2015faceattributes, karras2017progressive} for evaluation. In addition, we have also collected a handful of images of out-of-domain and famous figures, to highlight our identity preservation capabilities, and the unprecedented extent of images we can handle that could not be edited until now.

We start by qualitatively and quantitatively comparing our approach to current inversion methods, both in terms of reconstruction quality and the quality of downstream editing. We use the direct optimization scheme proposed by Karras et al.~\cite{karras2020analyzing} to invert real images to $\mathcal{W}$ space, which we denote by SG2. A similar optimization is used to invert to the extended $\mathcal{W+}$ space ~\cite{abdal2019image2stylegan}, denoted by SG2 $\mathcal{W+}$. We also compare to e4e, the encoder designed by Tov et al. \cite{tov2021designing}, which uses the $\mathcal{W+}$ space but seeks to remain relatively close to $\mathcal{W}$. Each baseline inverts to a different part of the latent space, demonstrating the different aspects of the distortion-editability trade-off. Note that we do not include Richardson et al.~\cite{richardson2020encoding} in our comparisons, since Tov et al. have convincingly shown editing superiority, rendering this comparison redundant.

\subsection{ Reconstruction Quality}
\label{sec:recon_quality}

\noindent{\bf Qualitative evaluation.} Figures~\ref{fig:rec_out} and~\ref{fig:rec_celebA} present a qualitative comparison of visual quality of inverted images. As can be seen, even before considering editability, our method achieves superior reconstruction results for all examples, especially for out-of-domain ones, as our method is the only one to successfully reconstruct challenging details such as face painting or hands (Figure~\ref{fig:rec_out}).
Our method is also capable of reconstructing fine-details which most people are sensitive to, such as the make-up, lighting, wrinkles, and more (Figure~\ref{fig:rec_celebA}). For more visual results, see the Supplementary Materials.

\noindent{\bf Quantitative evaluation.} For quantitative evaluation, we employ the following metrics: pixel-wise distance using $MSE$, perceptual similarity using $LPIPS$ \cite{zhang2018unreasonable},  structural similarity using $MS-SSIM$ \cite{wang2003multiscale}, and identity similarity by employing a pretrained face recognition network \cite{deng2019arcface}. The results are shown in Table~\ref{tab:rec}. As can be seen, the results align with our qualitative evaluation as we achieve the best score for each metric by a substantial margin.

\begin{table}
\begin{center}

\begin{tabular}{lccccc}
\toprule
Measure &\textbf{Ours}  & e4e & SG2 & SG2 $\mathcal{W+}$  \\  
\midrule
LPIPS $\downarrow$ & \textbf{0.09} & $0.4$ & $0.4$ & $0.34$ \\
MSE $\downarrow$ & \textbf{0.014} & $0.05$ & $0.08$ & $0.043$ \\
MS SSIM $\downarrow$ & \textbf{0.21} & $0.38$ & $0.38$ & $0.3$ \\
ID Similarity $\uparrow$ &  \textbf{0.9} & $0.75$ & $0.8$ & $0.85$ \\
\bottomrule
\end{tabular}

\end{center}
\caption{Quantitative reconstruction quality. Using a StyleGAN2 generator trained over the FFHQ dataset, we invert images from the CelebA-HQ test set and measure their reconstruction using four different metrics. All metrics indicate superior reconstruction for our method. 
}
\label{tab:rec} 
\end{table}

\begin{figure}
\centering
\begin{tabular}{l}
\hspace{0.1cm}  \textbf{Original}  \hspace{0.25cm} \textbf{SG2 $\mathcal{W+}$} \hspace{0.45cm} \textbf{e4e}  \hspace{0.9cm} \textbf{SG2} \hspace{0.85cm} \textbf{Ours} \\
\raisebox{-.5\totalheight} {\includegraphics[width=\columnwidth]{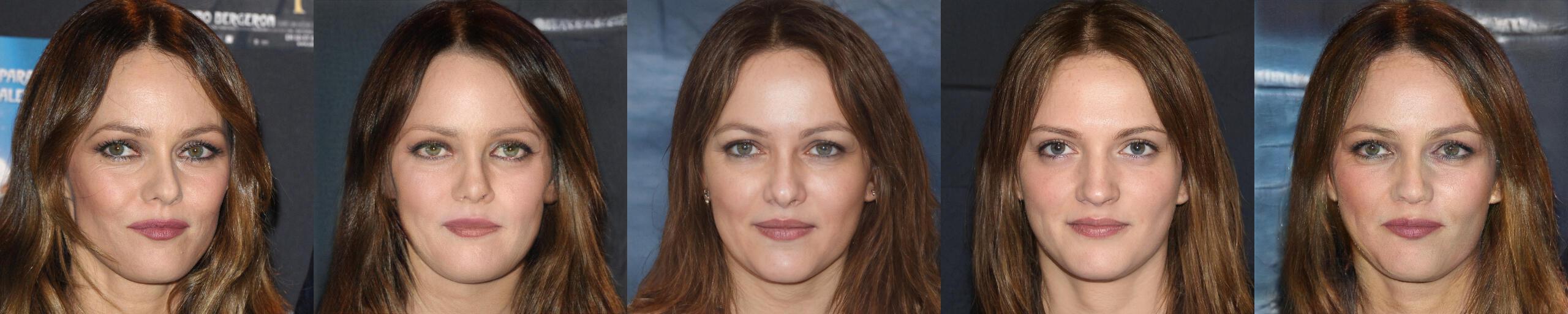}} \\
\noalign{\vskip 1mm}

\raisebox{-.5\totalheight}{\includegraphics[width=\columnwidth]{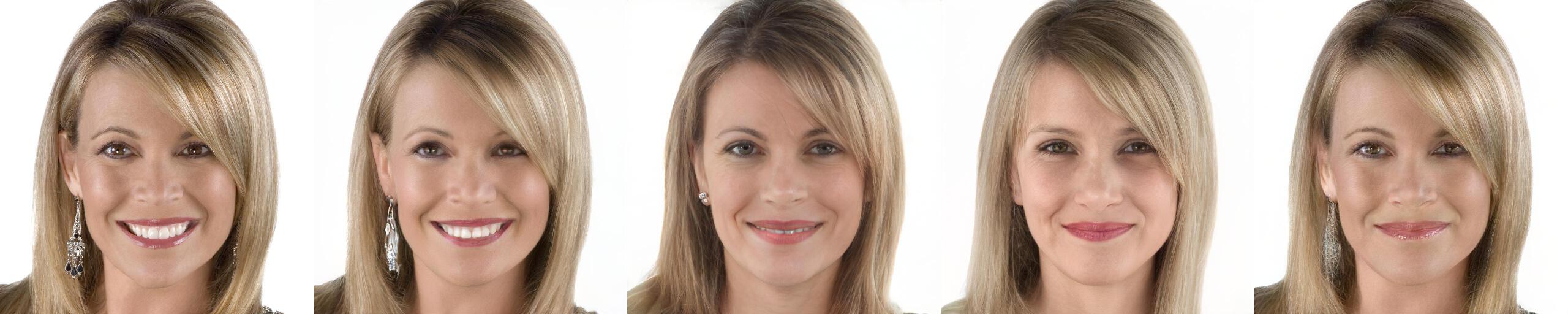}} \\
\end{tabular}
\vspace{0.1cm}
\caption{Editing comparison of images from the CelebA-HQ dataset. We demonstrate the pose (top) and smile removal (bottom) edits. The edits over SG2 $\mathcal{W+}$ do not create the desired effect, e.g., mouth is not closed in the bottom row. SG2 and e4e achieve better editing, but lose the original identity. PTI achieves high quality editing  while preserving the identity. For more uncurated examples, see the Supplementary Materials. Zoom-in recommended.}

\label{fig:edit_celeba}
\end{figure}

\subsection{Editing Quality}

\begin{figure}
\centering
\begin{tabular}{c}
\hspace{0.4cm}  \textbf{Original} \hspace{1.2cm} \textbf{StyleClip}  \hspace{0.8cm} \textbf{PTI+StyleClip} \\
 
 \raisebox{-.5\totalheight} {\includegraphics[width=\columnwidth]{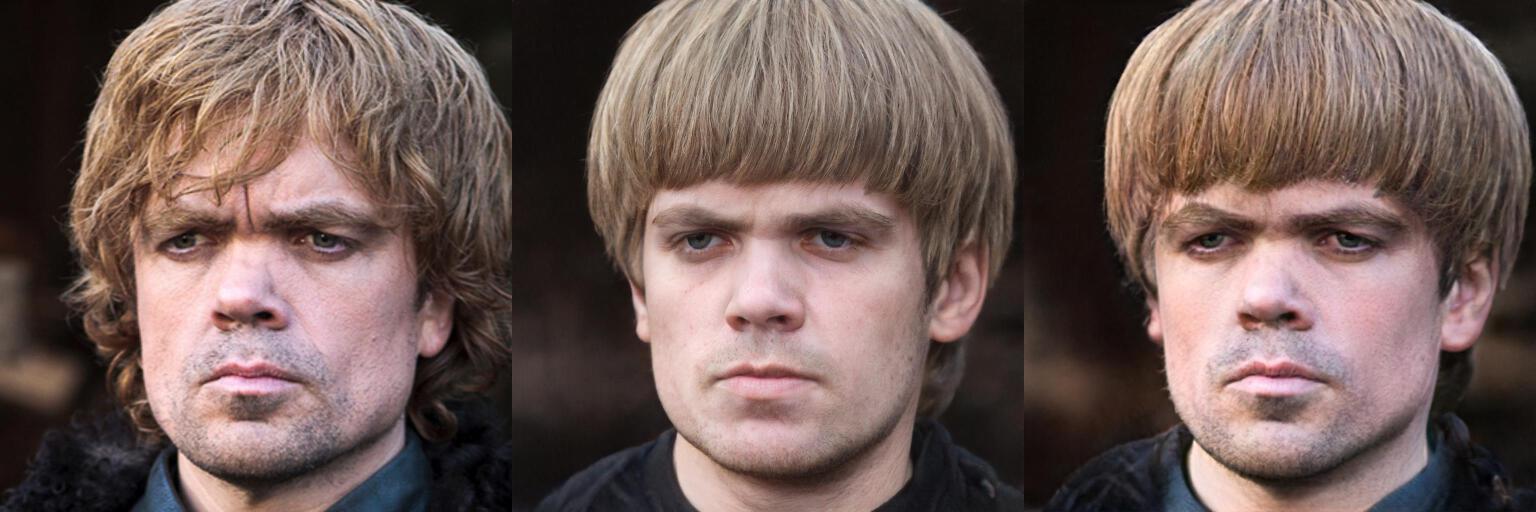}} \\
\noalign{\vskip 1mm}
\raisebox{-.5\totalheight} {\includegraphics[width=\columnwidth]{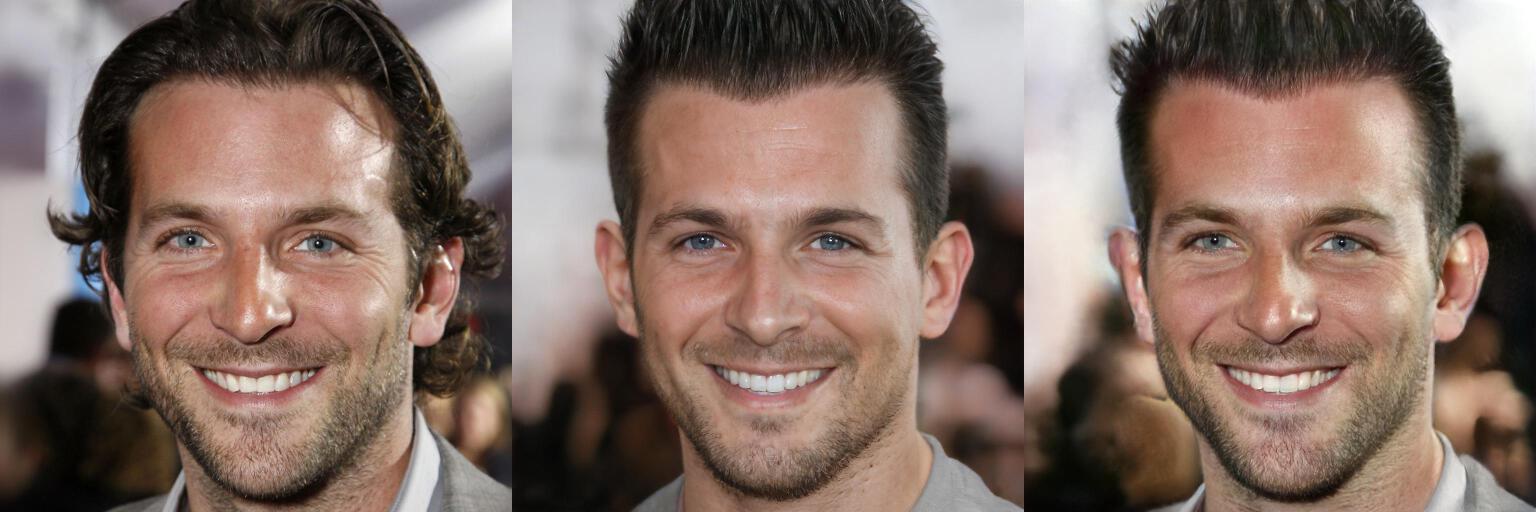}} \\
\noalign{\vskip 1mm}

\raisebox{-.5\totalheight} {\includegraphics[width=\columnwidth]{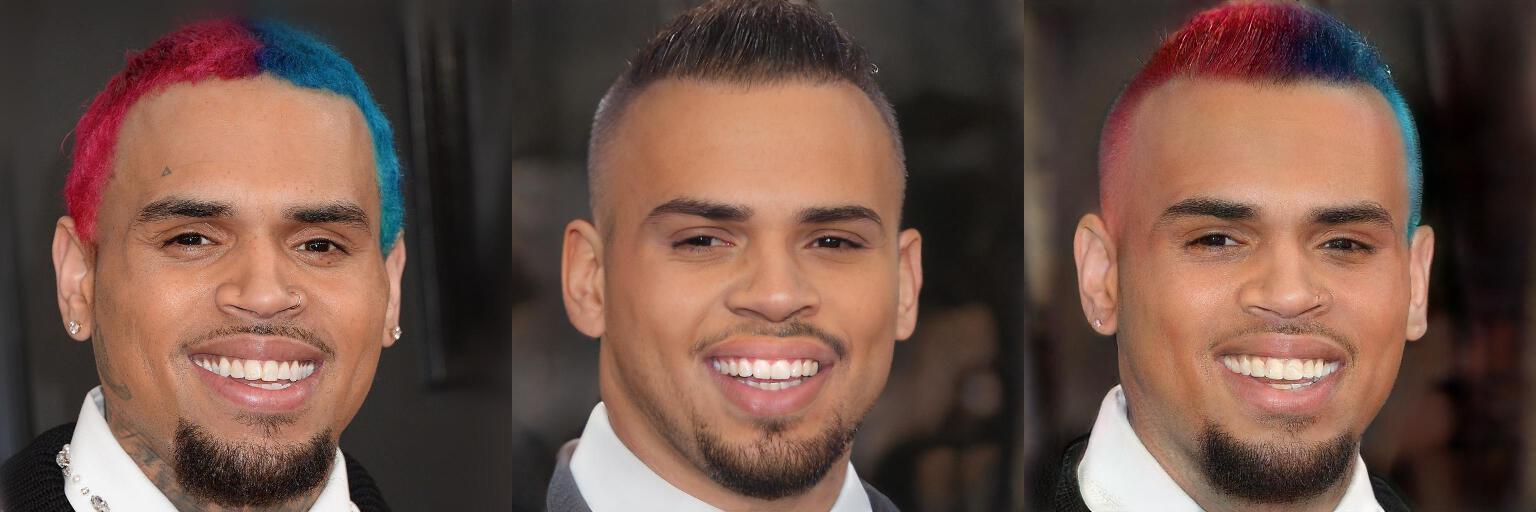}} \\
\end{tabular}
\vspace{0.1cm}
\caption{StyleClip editing demonstration. Using StyleClip~\cite{patashnik2021styleclip} to perform the "bowl cut" and "mohawk" edits (middle column), a clear improvement in identity preservation can be seen when first employing PTI (right).}
\label{fig:styleClip_inv}
\end{figure}

\begin{figure}
\centering
\begin{tabular}{c}
\hspace{0.4cm}  \textbf{Original} \hspace{1.2cm} \textbf{StyleClip}  \hspace{0.8cm} \textbf{PTI+StyleClip} \\
\raisebox{-.5\totalheight} {\includegraphics[width=\columnwidth]{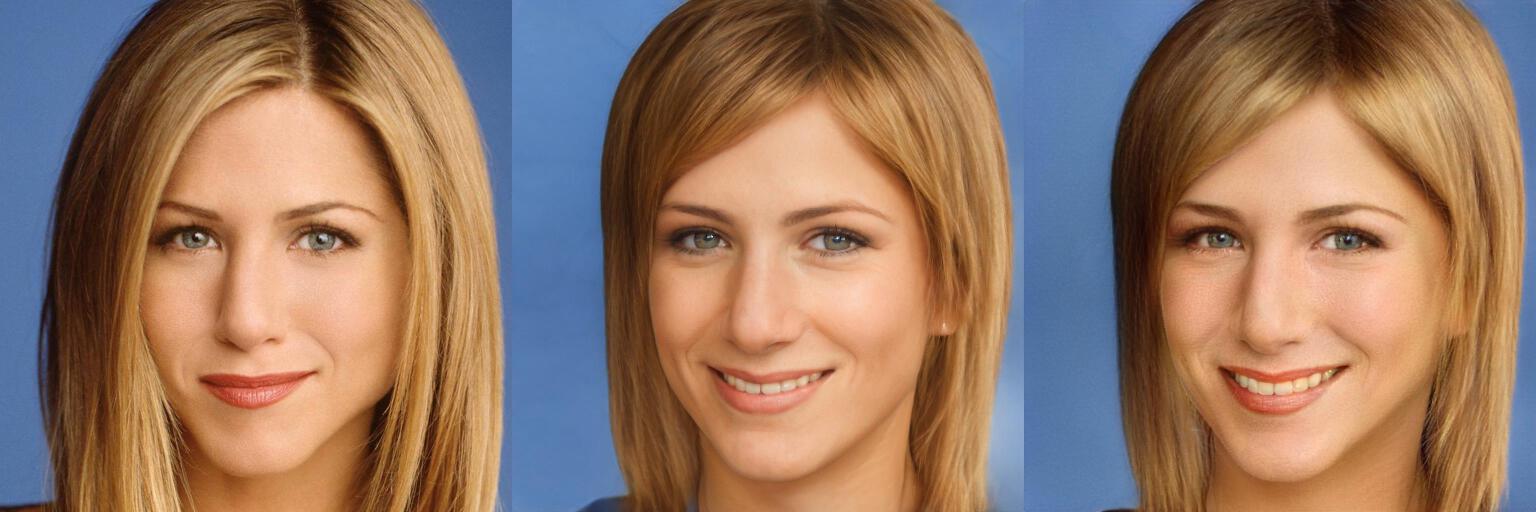}} \\
\noalign{\vskip 1mm}
\raisebox{-.5\totalheight} {\includegraphics[width=\columnwidth]{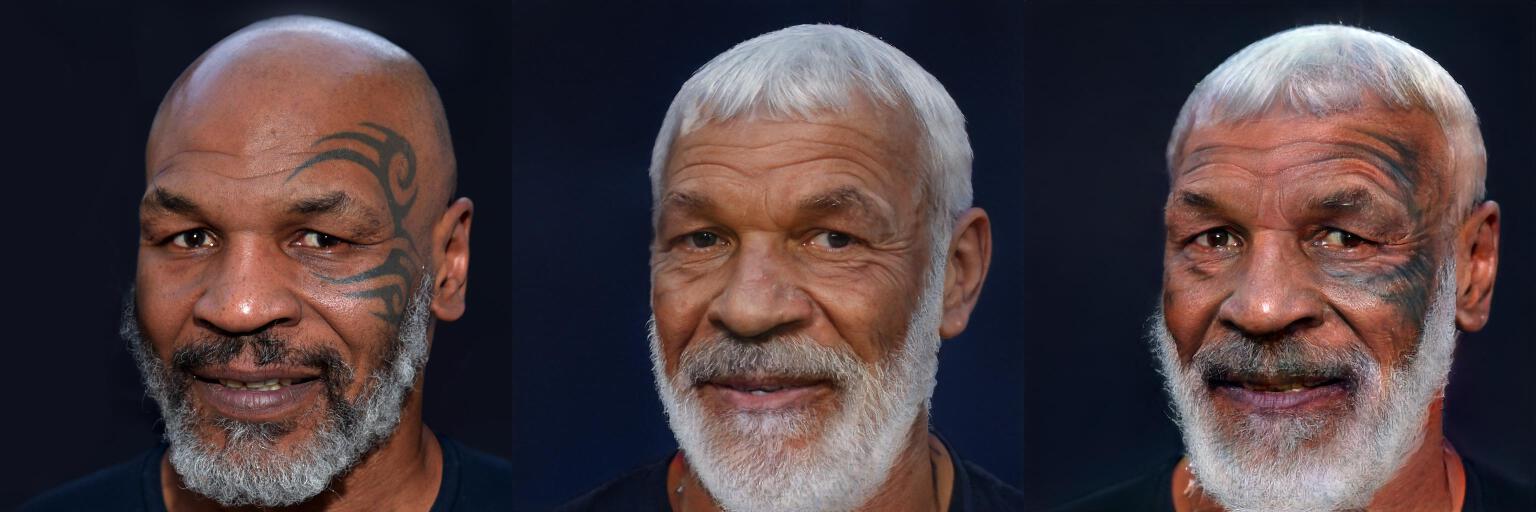}} \\
\noalign{\vskip 1mm}
\raisebox{-.5\totalheight} {\includegraphics[width=\columnwidth]{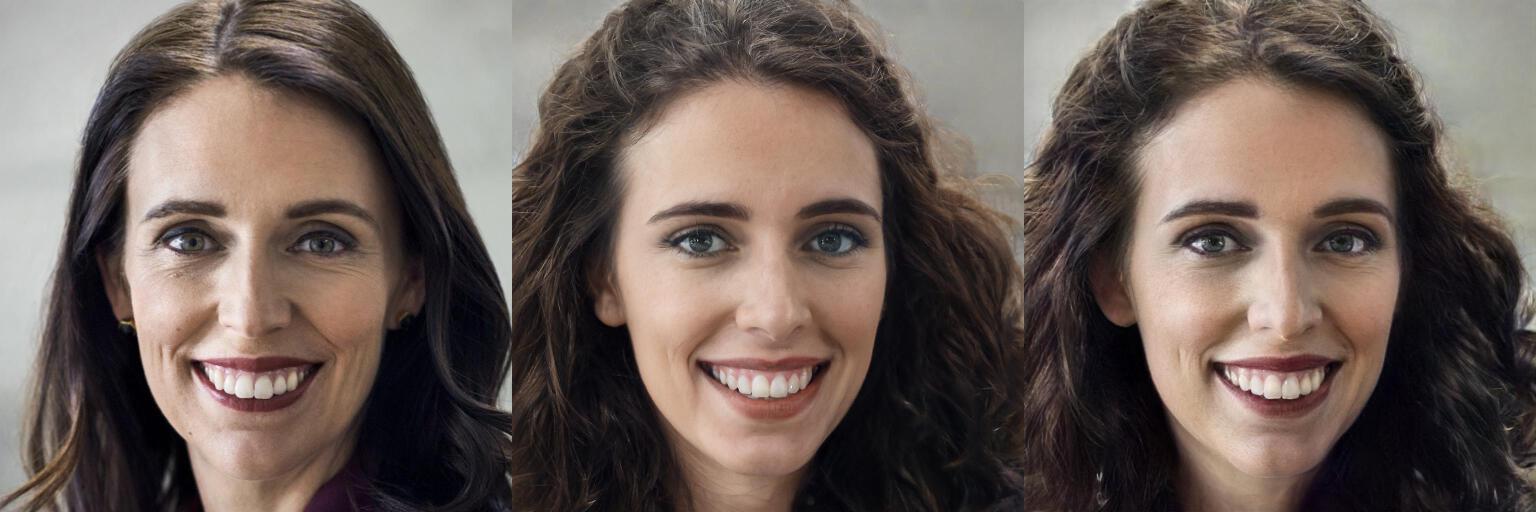}} \\
\end{tabular}
\vspace{0.1cm}
\caption{Sequential editing of StyleClip and InterfaceGAN edits with and without pivotal tuning inversion (PTI). Top row: "Bob cut hair", smile, and rotation. Middle row: "bowl cut hair" and older. Bottom row: "curly hair", younger and rotation.
}
\label{fig:styleClip}
\end{figure}

\begin{figure*}

\centering
\begin{tabular}{l}

\hspace{1.0cm}  \textbf{Original}  \hspace{2cm} \textbf{SG2 $\mathcal{W+}$} \hspace{2.3cm} \textbf{e4e}  \hspace{2.8cm} \textbf{SG2} \hspace{2.6cm} \textbf{Ours} \\

\raisebox{-.5\totalheight}{\includegraphics[width=1\textwidth]{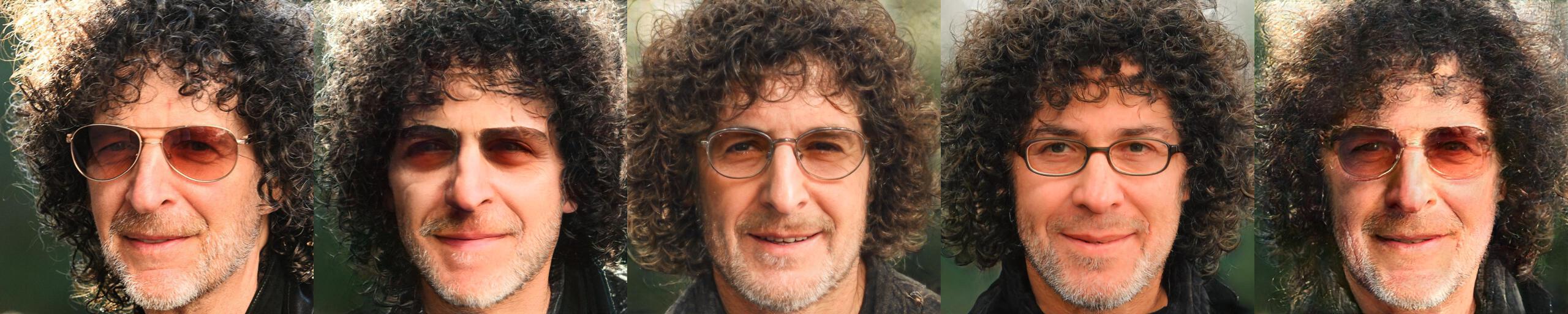}} \\
\noalign{\vskip 1mm}
\raisebox{-.5\totalheight}{\includegraphics[width=1\textwidth]{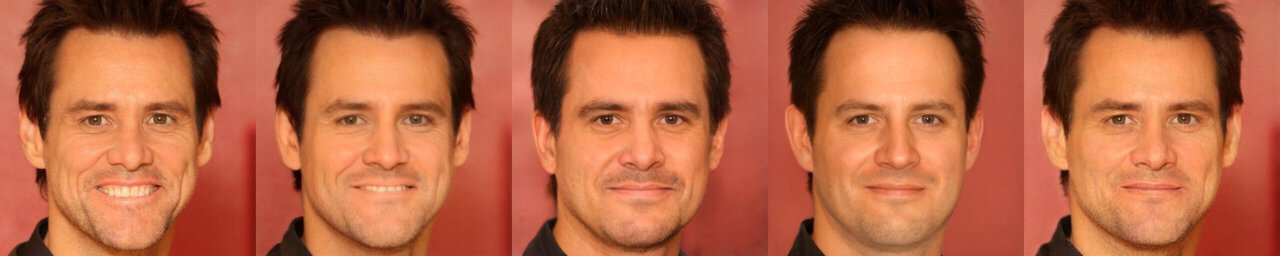}} \\
\noalign{\vskip 1mm}
\raisebox{-.5\totalheight}{\includegraphics[width=1\textwidth]{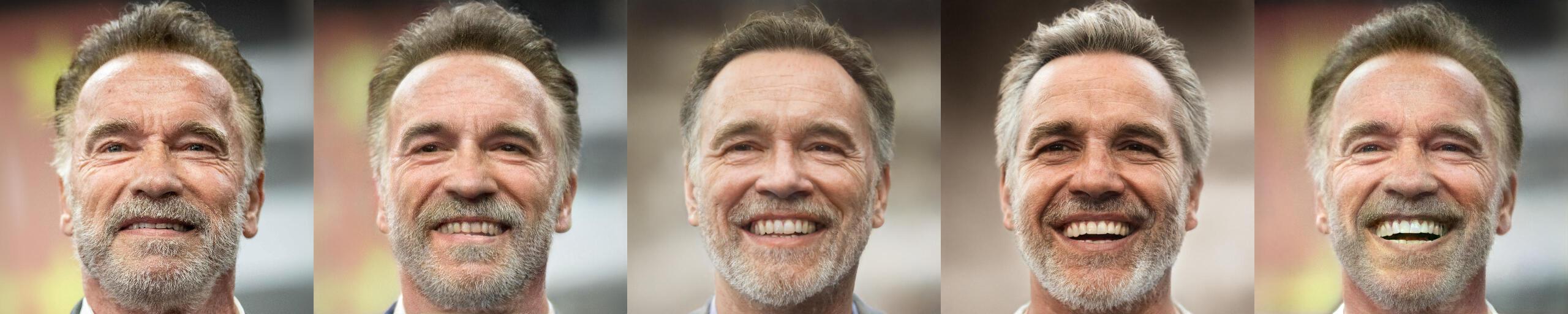}} \\
\noalign{\vskip 1mm}
\raisebox{-.5\totalheight}{\includegraphics[width=1\textwidth]{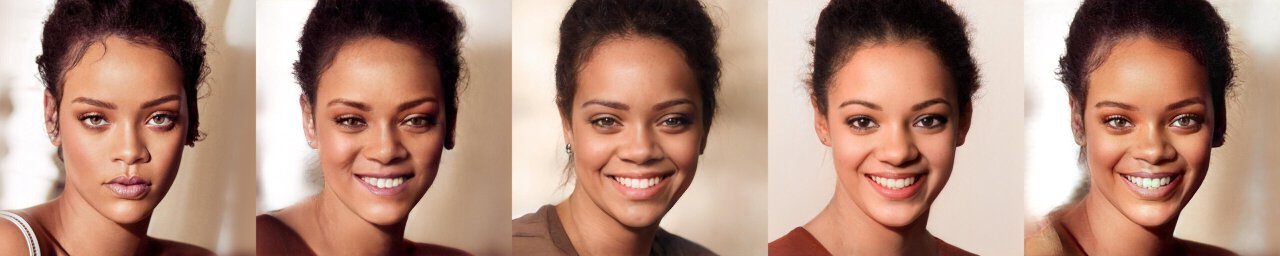}} \\
\end{tabular}
\vspace{0.1cm}
\caption{Editing comparison of famous figures collected from the web. We demonstrate the following edits (top to bottom): pose, mouth closing, and smile. Similar to Figure~\ref{fig:edit_celeba}, we again witness how SG2 $\mathcal{W+}$ do not induce significant edits, and the others do not preserve identity, in contrast to our approach, which achieves both.}
\label{fig:edit}
\end{figure*}

Editing a facial image should preserve the original identity while performing a meaningful and visually plausible modification. 
However, it has been shown \cite{zhu2020improved, tov2021designing} that using less editable embedding spaces, such as $\mathcal{W+}$, results in better reconstruction, but also in less meaningful editing compared to the native $\mathcal{W}$ space. For example, using the same latent edit, rotating a face in $\mathcal{W}$ space results in a higher rotation angle compared to $\mathcal{W+}$. Hence, in cases of minimal effective editing, the identity may seem to be preserved rather well. Therefore, we evaluate editing quality on two axes: identity preservation and editing magnitude.

\begin{figure*}

\centering
\begin{tabular}{l}
\hspace{1.0cm}  \textbf{Original}  \hspace{2cm} \textbf{SG2 $\mathcal{W+}$} \hspace{2.3cm} \textbf{e4e}  \hspace{2.8cm} \textbf{SG2} \hspace{2.6cm} \textbf{Ours} \\
\raisebox{-.5\totalheight}{\includegraphics[width=1\textwidth]{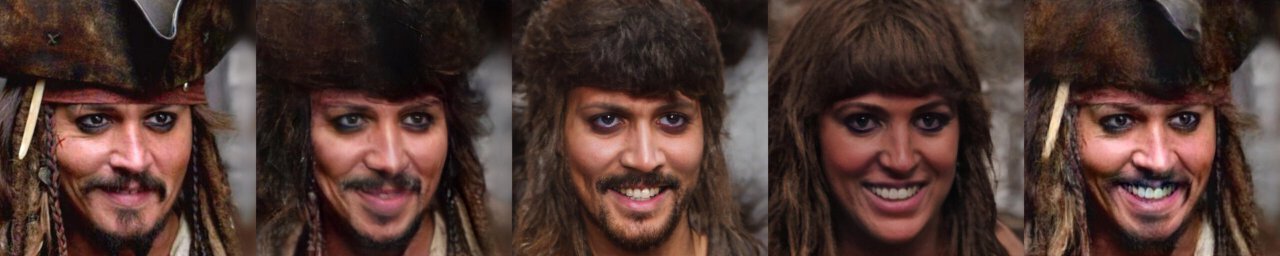}} \\
\noalign{\vskip 1mm}
\raisebox{-.5\totalheight}{\includegraphics[width=1\textwidth]{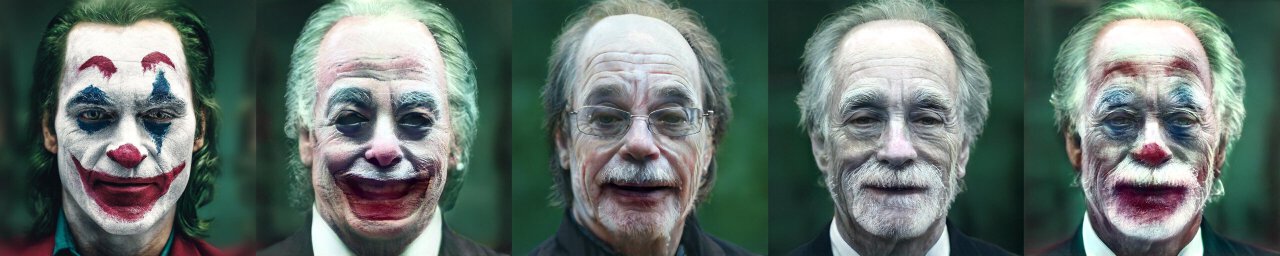}} \\
\noalign{\vskip 1mm}
\raisebox{-.5\totalheight}{\includegraphics[width=1\textwidth]{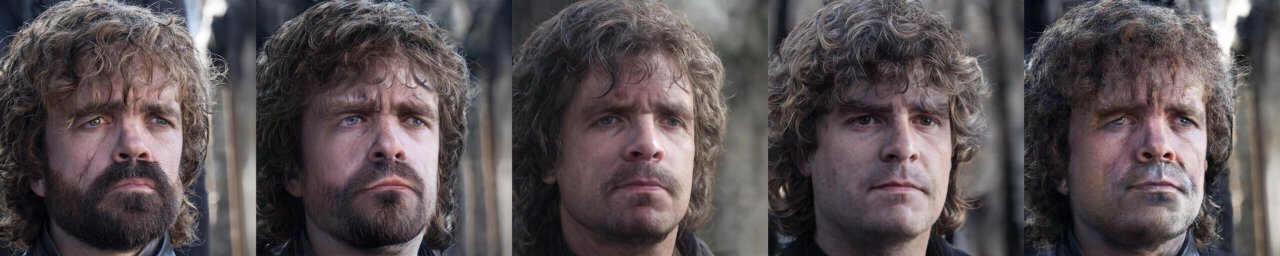}} \\

\end{tabular}
\vspace{0.1cm}
\caption{Editing of smile, age, and beard removal (top to bottom) comparison over out-of-domain images collected from the web. The collected images portray unique hairstyles, hair colors, and apparel, along with unique facial features, such as heavy make-up and scars. Even in these challenging cases, our results retain the original identity while enabling meaningful edits.}
\label{fig:edit_out}
\end{figure*}

\begin{figure}
\centering
\begin{tabular}{c c c c}
\rotatebox[origin=t]{90}{Original} & 
\raisebox{-.5\totalheight}{\includegraphics[width=0.28\columnwidth]{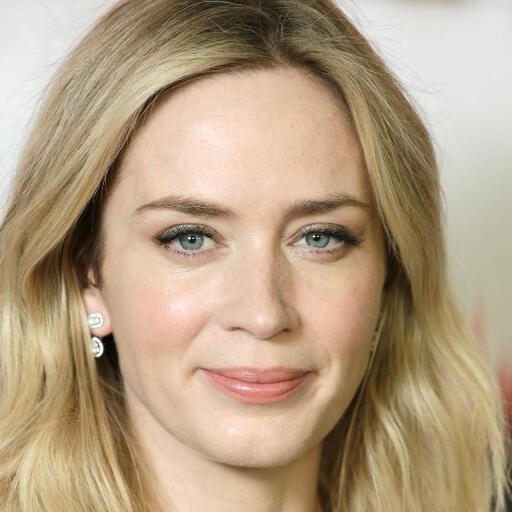}} & \raisebox{-.5\totalheight}{\includegraphics[width=0.28\columnwidth]{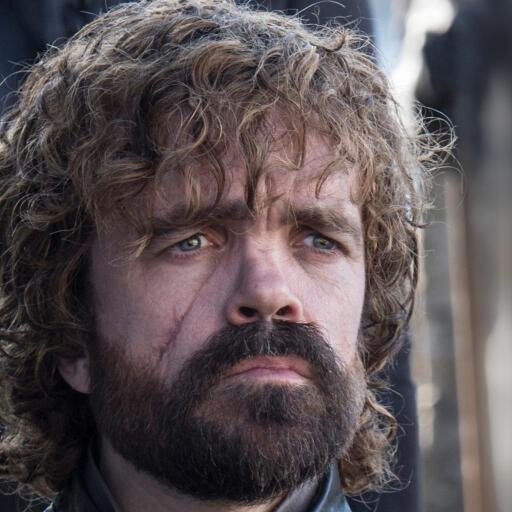}} &
\raisebox{-.5\totalheight}{\includegraphics[width=0.28\columnwidth]{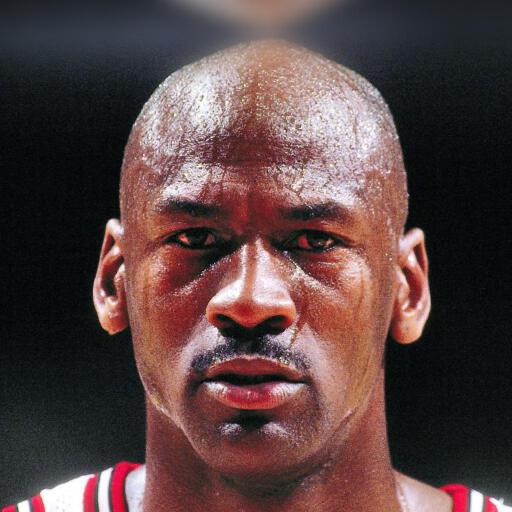}}\\
\noalign{\vskip 0.1cm}

\rotatebox[origin=t]{90}{Sequential} & 
\raisebox{-.5\totalheight}{\includegraphics[width=0.28\columnwidth]{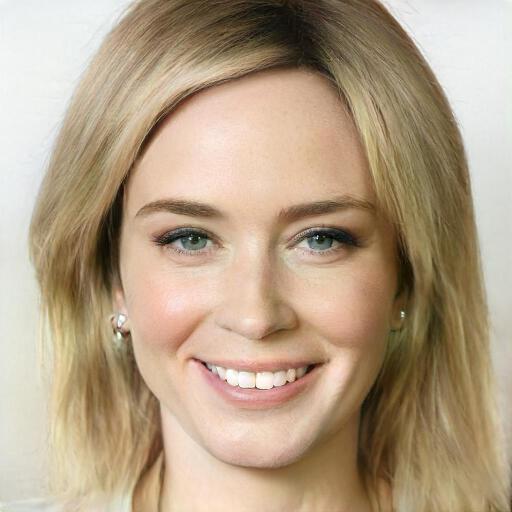}} &
\raisebox{-.5\totalheight}{\includegraphics[width=0.28\columnwidth]{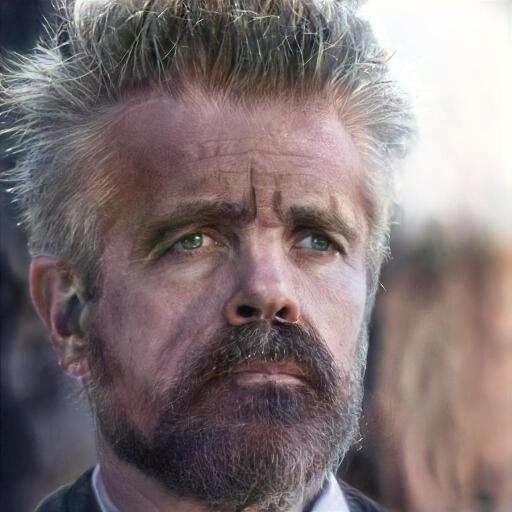}} & 
\raisebox{-.5\totalheight}{\includegraphics[width=0.28\columnwidth]{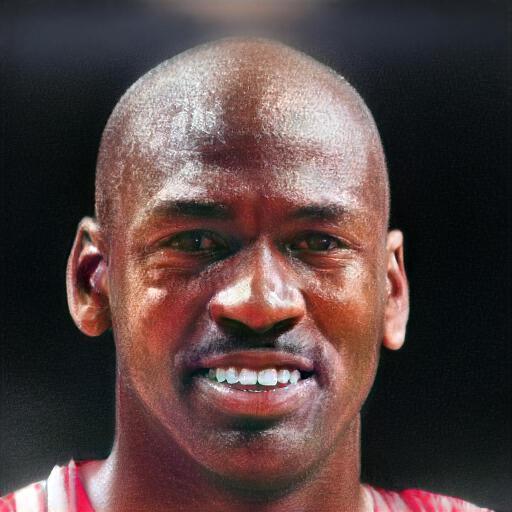}} \\

\end{tabular}
\vspace{0.1cm}
\caption{Additional sequential editing examples. Left: hair, pose and smile edits. Middle: hair and age. Right: pose and smile.}
\label{fig:sequential_editing4}
\end{figure}

\noindent{\bf Qualitative evaluation.} We use the popular GANSpace~\cite{harkonen2020ganspace} and InterfaceGAN~\cite{shen2020interpreting} methods for latent-based editing. These approaches are orthogonal to ours, as they require the use of an inversion algorithm to edit real images. As can be expected, the $\mathcal{W}+$-based method preserves the identity rather well, but fails to perform significant edits, the $\mathcal{W}$-based one is able to perform the edit, but loses the identity, and e4e provides a compromise between the two. In all cases, our method preserves identity the best and displays the same editing quality as for $\mathcal{W}$-based inversions. Figure~\ref{fig:edit_celeba} presents an editing comparison over the CelebA-HQ dataset. We also investigate our performance using images of other iconic characters (Figures~\ref{fig:teaser} and ~\ref{fig:edit}) and more challenging out-of-domain facial images (Figure~\ref{fig:edit_out}). The ability to perform sequential editing is presented in Figures~\ref{fig:sequential_editing3}, and ~\ref{fig:sequential_editing4}. In addition, we demonstrate our ability to invert multiple identities using the same generator in Figures~\ref{fig:leaders_multi_id1} and~\ref{fig:friends_multi_id}. 
For more visual and uncurated results, see the Supplementary Materials. As can be seen, our method successfully performs meaningful edits, while preserving the original identity successfully.

The recent work of StyleClip~\cite{patashnik2021styleclip} demonstrates unique edits, driven by natural language. In Figures~\ref{fig:styleClip_inv}, and \ref{fig:styleClip} we demonstrate editing results using this model, and demonstrate substantial identity preservation improvement, thus extending StyleClip's scope to more challenging images. We use the mapper-based variant proposed by the paper, where the edits are achieved by training a mapper network to edit input latent codes. Note that the StyleClip model is trained to handle codes returned by the e4e method. Hence, to employ this model, our PTI process uses e4e-based pivots instead of $\mathcal{W}$ ones. As can be expected, we observe that the editing capabilities of the e4e codes are preserved, while the inherent distortion caused by e4e is diminished using PTI.
More results for this experiment can be found in the supplementary materials.

\noindent{\bf Quantitative evaluation} results are summarized in Table~\ref{tab:id}. To measure the two aforementioned axes, we compare the effects of the same latent editing operation between the various aforementioned baselines, and the effects of editing operations that yield the same editing magnitude. 
To evaluate editing magnitude, we apply a single pose editing operation and measure the rotation angle using Microsoft Face API ~\cite{azure}, as proposed by Zhu et al. ~\cite{zhu2020improved}. As the editability increases, the magnitude of the editing effect increases as well. As expected, $\mathcal{W}$-based inversion induces a more significant edit compared to $\mathcal{W+}$ inversion for the same editing operation. As can be seen, our approach yields a magnitude that is almost identical to $\mathcal{W}$'s, surpassing e4e and $\mathcal{W+}$ inversions, which indicates we achieve high editability (first row). 

In addition, we report the identity preservation for several edits. We evaluate the identity change using a pretrained facial recognition network \cite{deng2019arcface}, 
and the edits we report for are smile, pose, and age. We report both the mean identity preservation induced by each of these edits (second row), and the one induced by performing them sequentially one after the other (third row). Results indeed validate that our method obtains better identity similarity compared to the baselines. 
 
Since the lack of editability might increase identity similarity, as previously mentioned, we also measure the identity similarity while performing rotation of the same magnitude. Expectedly, the identity similarity for $\mathcal{W+}$ inversion decreases significantly when using a fixed rotation angle editing, demonstrating it is less editable compared to other inversion methods.
Overall, the quantitative results demonstrate the distortion-editability tradeoff, as $\mathcal{W+}$ inversion achieves better ID similarity but lower edit magnitude, and $\mathcal{W}$ inversion achieves inferior ID similarity but higher edit magnitude. In contrast, our method preserves the identity well and provides highly editable embeddings, or in other words, we alleviate the distortion-editability trade-off.

\begin{figure*}

\centering
\begin{tabular}{ccccccc}
\rotatebox[origin=t]{90}{Input} & \raisebox{-.5\totalheight}{\includegraphics[width=0.13\textwidth]{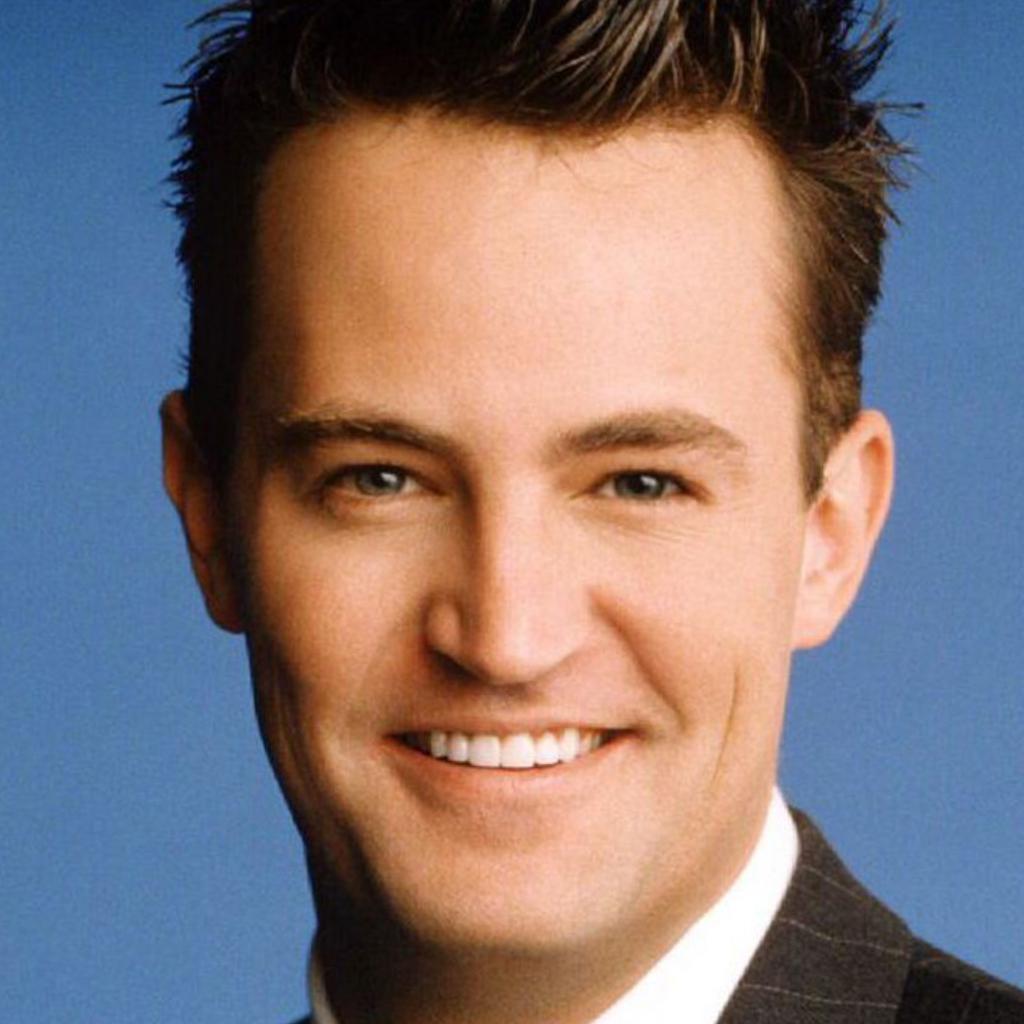}} & \raisebox{-.5\totalheight}{\includegraphics[width=0.13\textwidth]{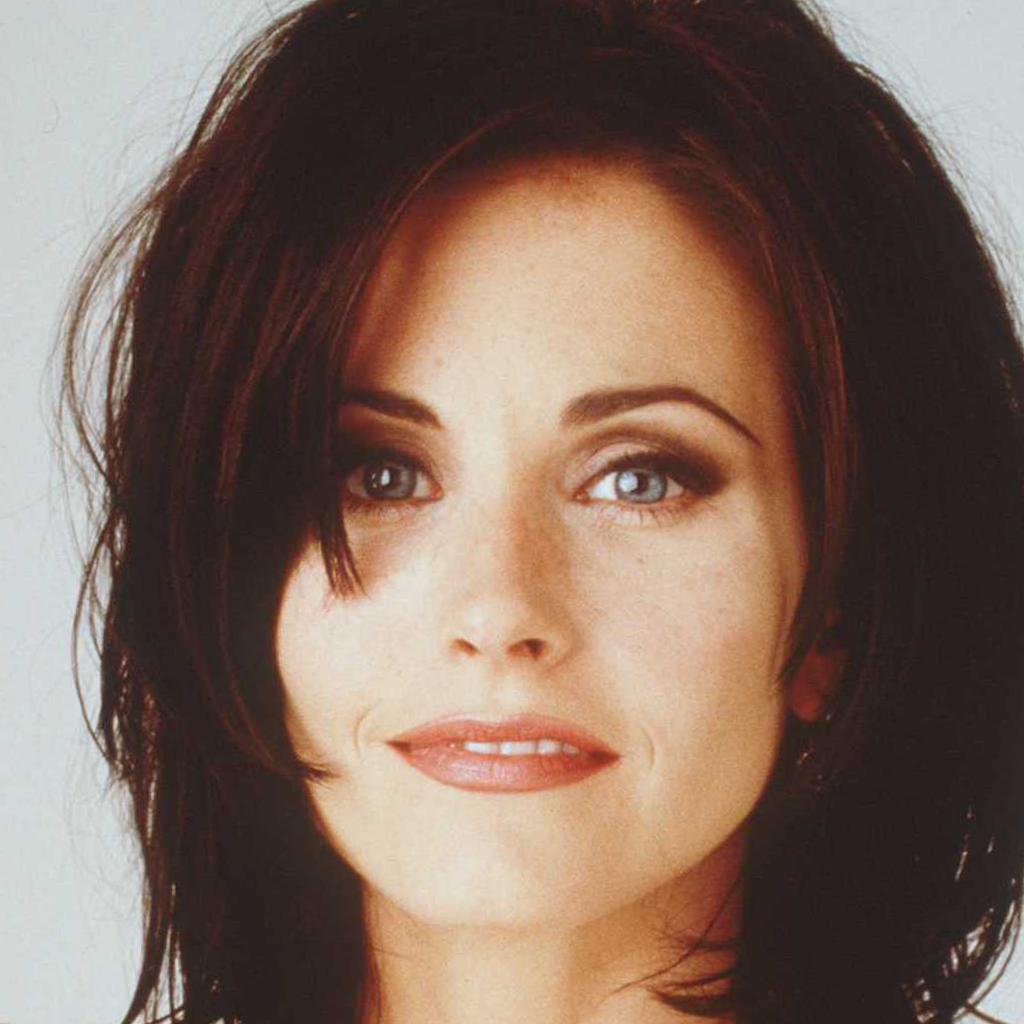}} & 
\raisebox{-.5\totalheight}{\includegraphics[width=0.13\textwidth]{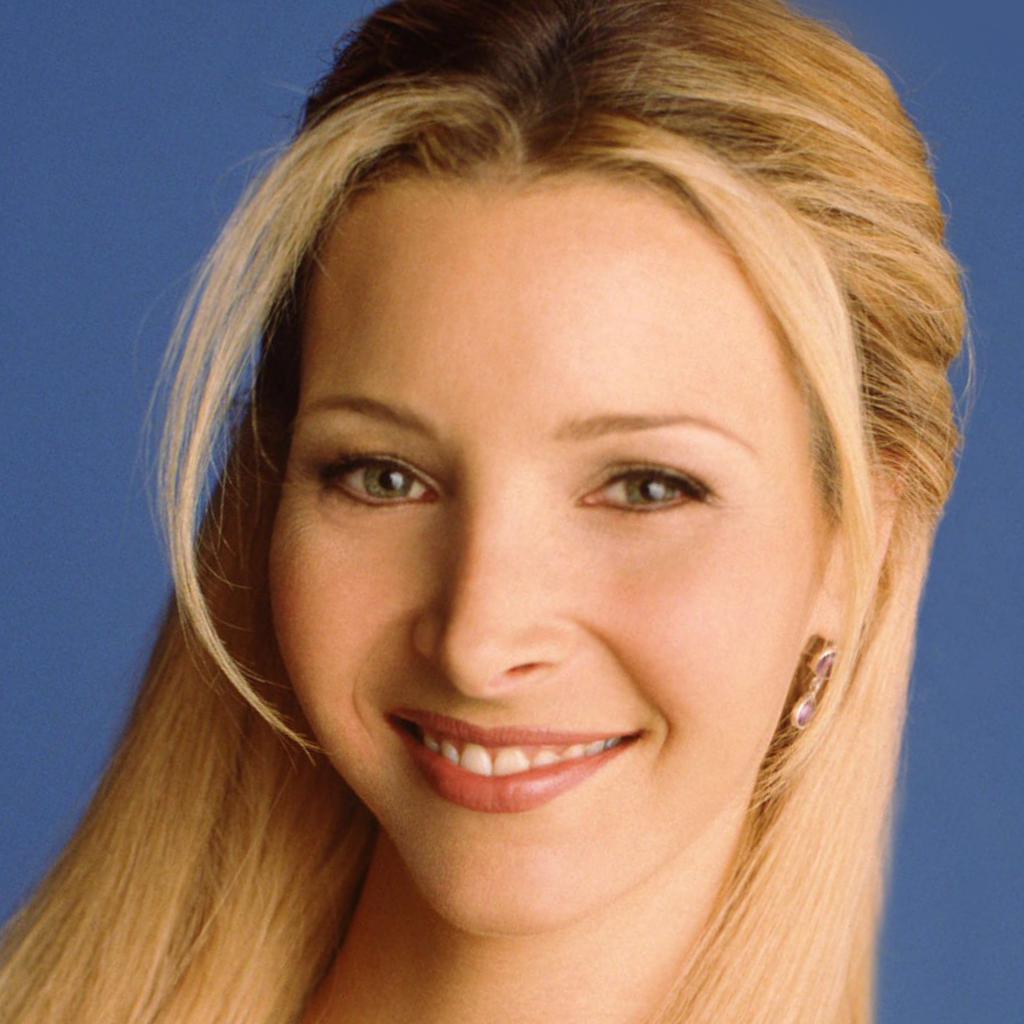}} & 
\raisebox{-.5\totalheight}{\includegraphics[width=0.13\textwidth]{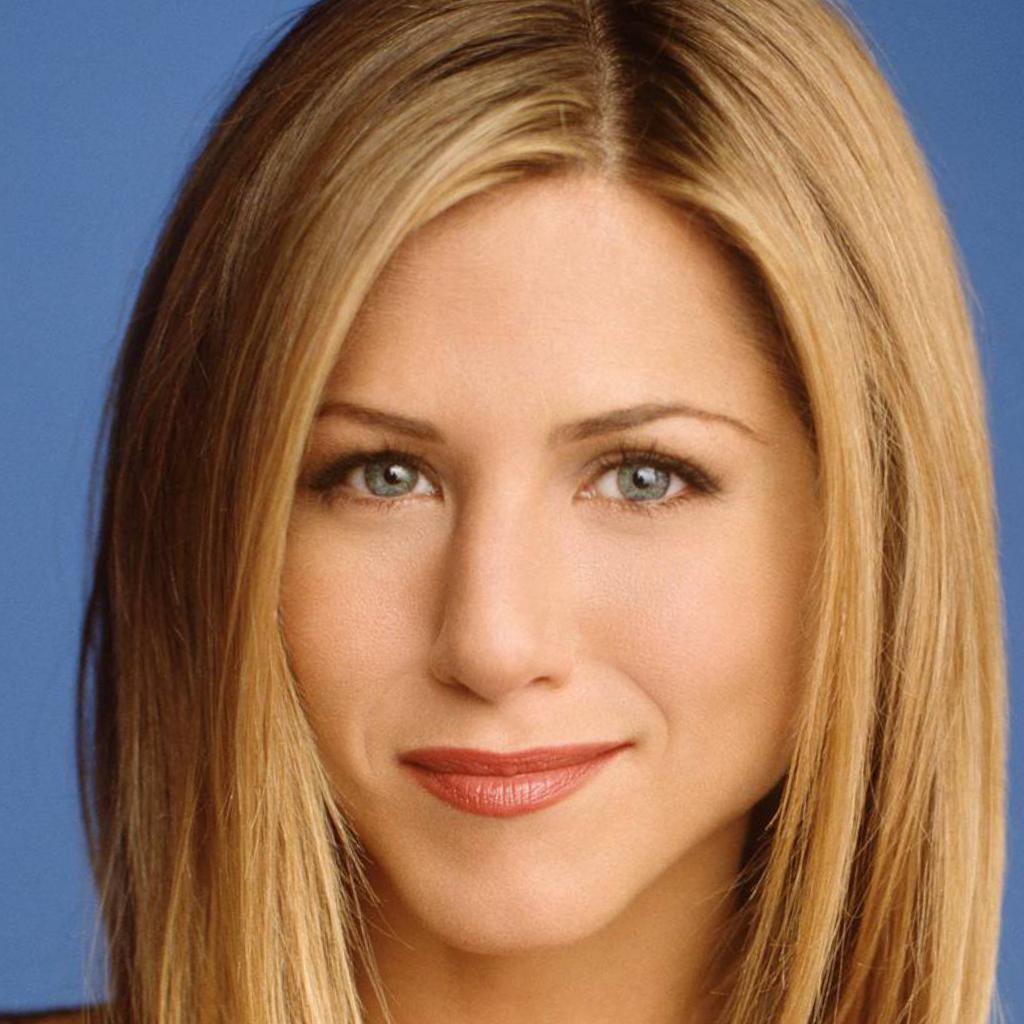}} & 
\raisebox{-.5\totalheight}{\includegraphics[width=0.13\textwidth]{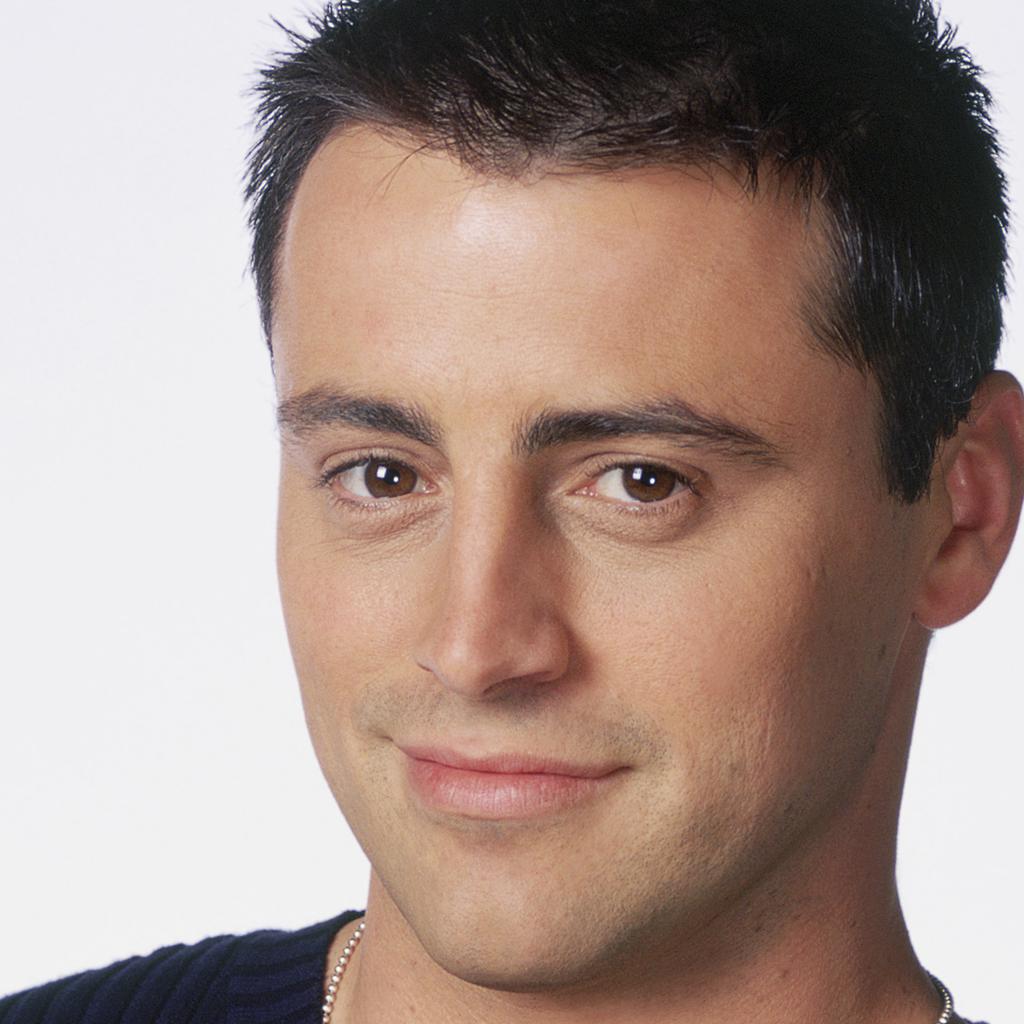}} &
\raisebox{-.5\totalheight}{\includegraphics[width=0.13\textwidth]{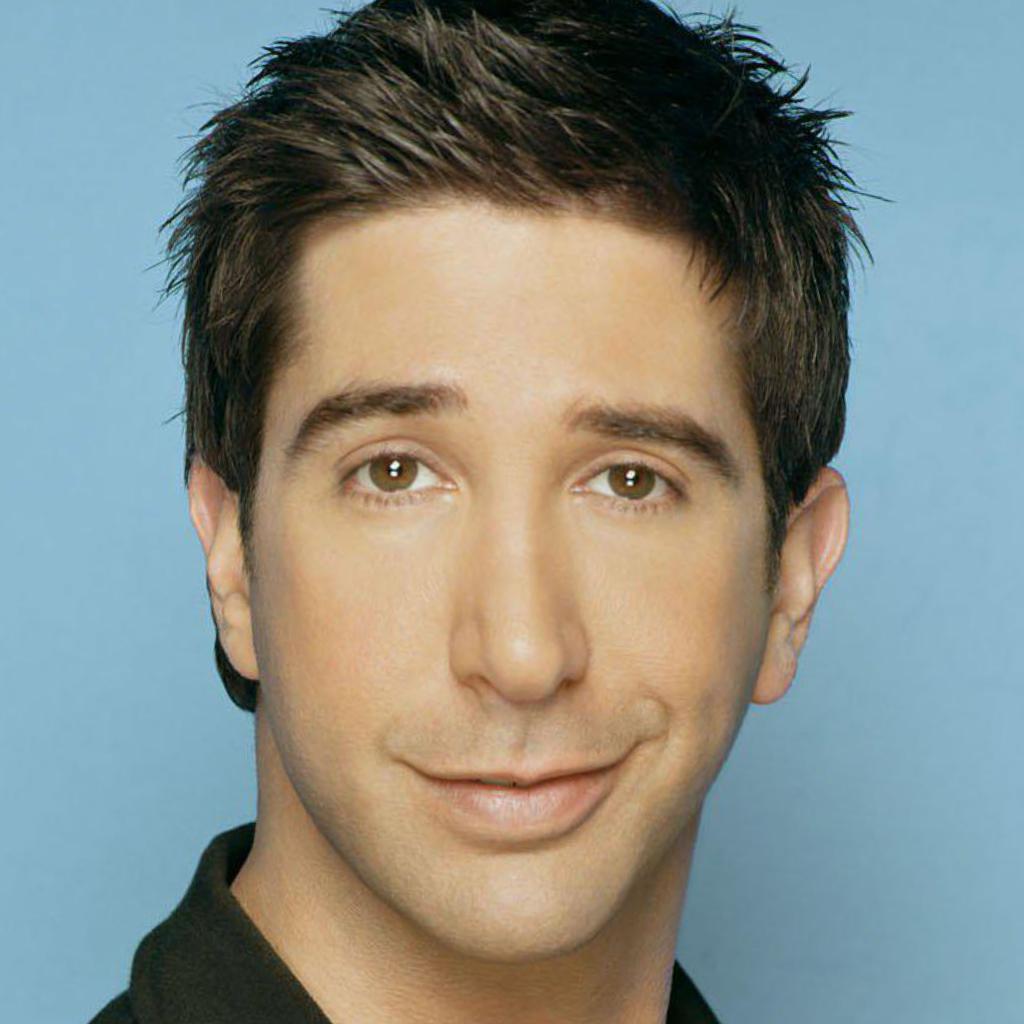}} \\
\noalign{\vskip 1mm}
\rotatebox[origin=t]{90}{Inversion} &
\raisebox{-.5\totalheight}{\includegraphics[width=0.13\textwidth]{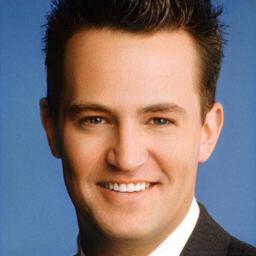}} &
\raisebox{-.5\totalheight}{\includegraphics[width=0.13\textwidth]{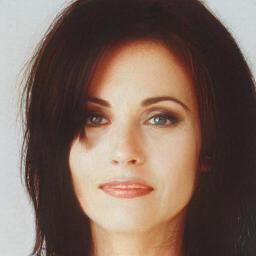}}&
\raisebox{-.5\totalheight}{\includegraphics[width=0.13\textwidth]{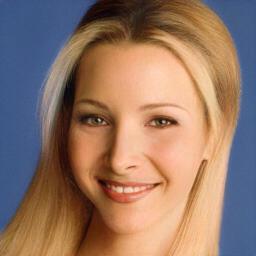}}&
\raisebox{-.5\totalheight}{\includegraphics[width=0.13\textwidth]{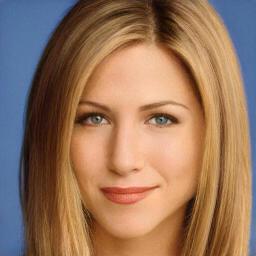}}&
\raisebox{-.5\totalheight}{\includegraphics[width=0.13\textwidth]{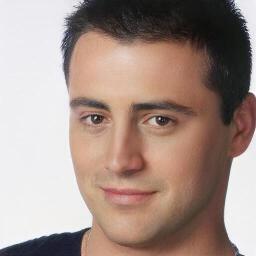}} &
\raisebox{-.5\totalheight}{\includegraphics[width=0.13\textwidth]{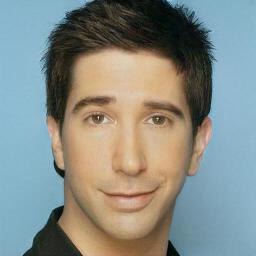}} \\ 
\noalign{\vskip 0.1cm}
\rotatebox[origin=t]{90}{+Age} &
\raisebox{-.5\totalheight}{\includegraphics[width=0.13\textwidth]{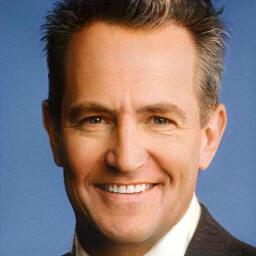}} &
\raisebox{-.5\totalheight}{\includegraphics[width=0.13\textwidth]{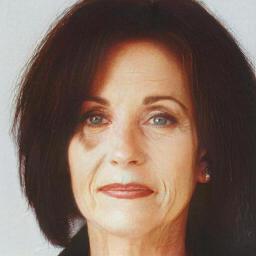}}&
\raisebox{-.5\totalheight}{\includegraphics[width=0.13\textwidth]{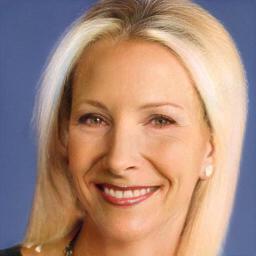}}&
\raisebox{-.5\totalheight}{\includegraphics[width=0.13\textwidth]{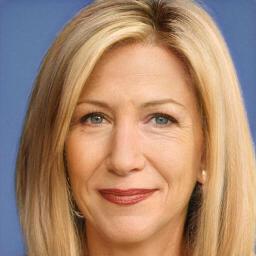}}&
\raisebox{-.5\totalheight}{\includegraphics[width=0.13\textwidth]{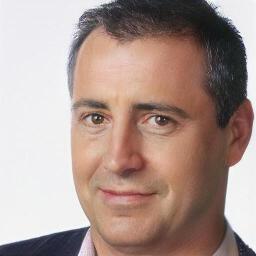}} &
\raisebox{-.5\totalheight}{\includegraphics[width=0.13\textwidth]{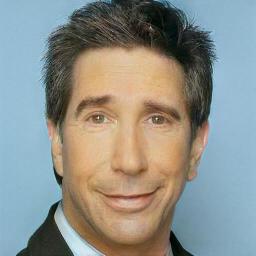}} \\
\noalign{\vskip 0.1cm}
\rotatebox[origin=t]{90}{$\pm$Smile} &
\raisebox{-.5\totalheight}{\includegraphics[width=0.13\textwidth]{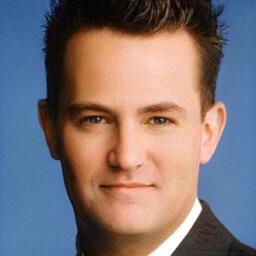}} &
\raisebox{-.5\totalheight}{\includegraphics[width=0.13\textwidth]{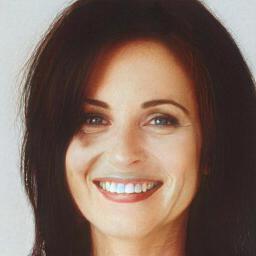}}&
\raisebox{-.5\totalheight}{\includegraphics[width=0.13\textwidth]{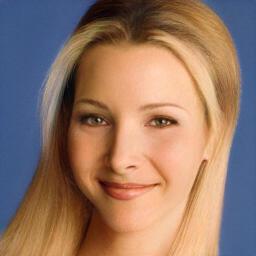}}&
\raisebox{-.5\totalheight}{\includegraphics[width=0.13\textwidth]{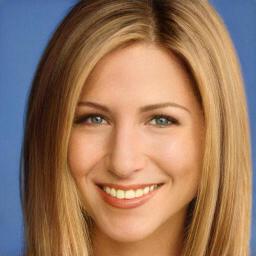}}&
\raisebox{-.5\totalheight}{\includegraphics[width=0.13\textwidth]{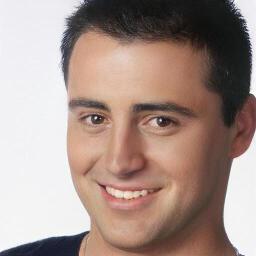}}  &
\raisebox{-.5\totalheight}{\includegraphics[width=0.13\textwidth]{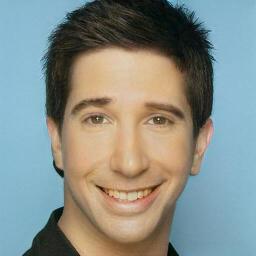}}\\
\end{tabular}
\vspace{0.1cm}
\caption{"Friends" StyleGAN. We simultaneously invert multiple identities into StyleGAN latent space, while retaining high editability and identity similarity.}
\label{fig:friends_multi_id}
\end{figure*}

\newcommand{\seqSize}{0.13}

\begin{figure*}

\centering
\begin{tabular}{c c c c c c c}
\rotatebox[origin=t]{90}{Original} &
\raisebox{-.5\totalheight}{\includegraphics[width=\seqSize\textwidth]{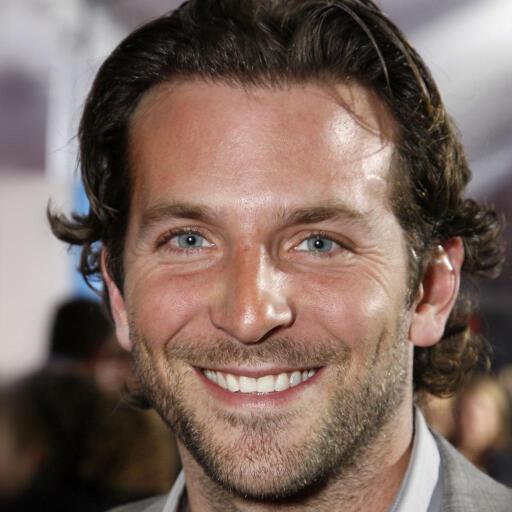}} & \raisebox{-.5\totalheight}{\includegraphics[width=\seqSize\textwidth]{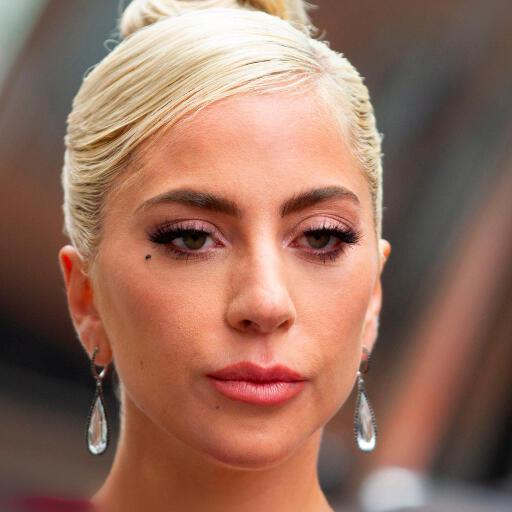}} & \raisebox{-.5\totalheight}{\includegraphics[width=\seqSize\textwidth]{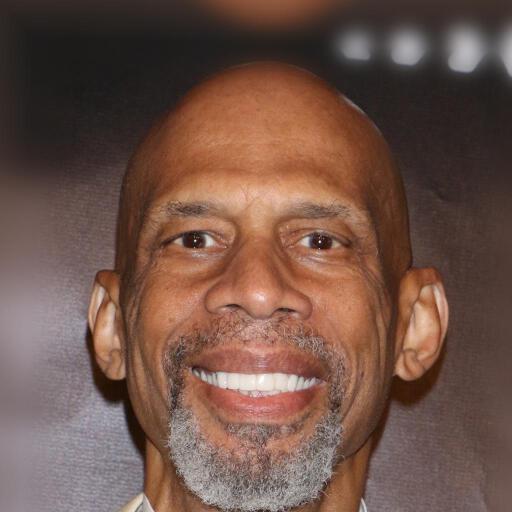}}  & \raisebox{-.5\totalheight}{\includegraphics[width=\seqSize\textwidth]{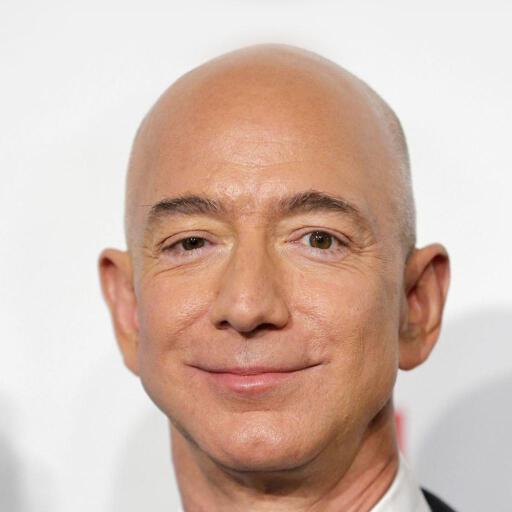}}   & \raisebox{-.5\totalheight}{\includegraphics[width=\seqSize\textwidth]{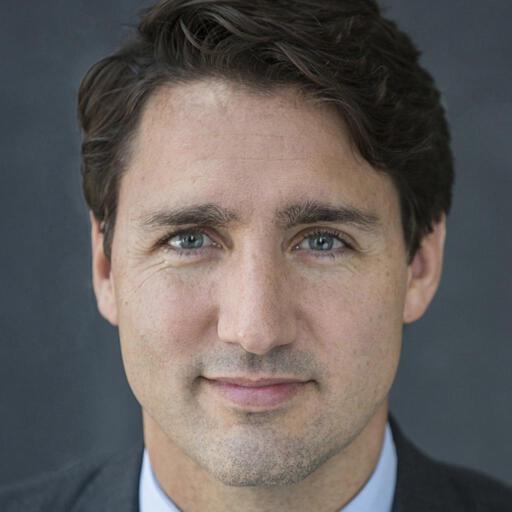}}   & \raisebox{-.5\totalheight}{\includegraphics[width=\seqSize\textwidth]{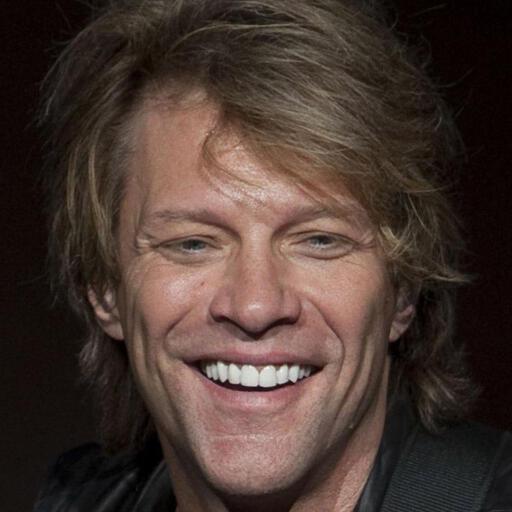}}\\
\noalign{\vskip 2mm}
\rotatebox[origin=t]{90}{$\pm$ Smile, Pose} &
\raisebox{-.5\totalheight}{\includegraphics[width=\seqSize\textwidth]{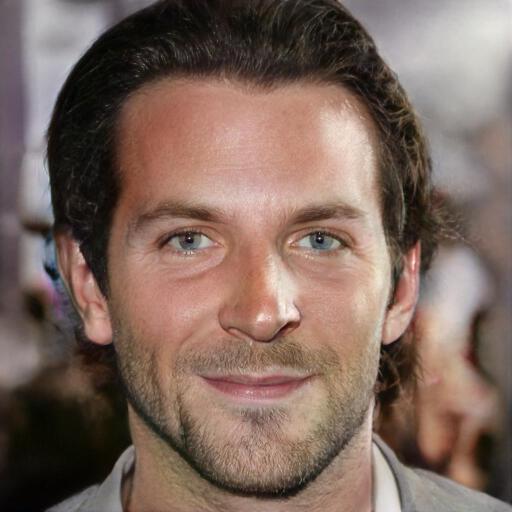}} & \raisebox{-.5\totalheight}{\includegraphics[width=\seqSize\textwidth]{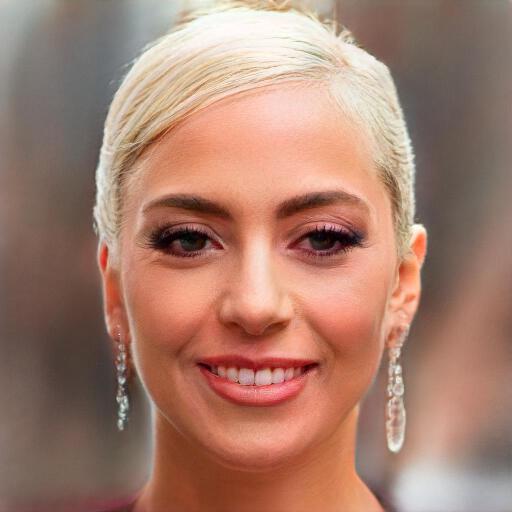}} & \raisebox{-.5\totalheight}{\includegraphics[width=\seqSize\textwidth]{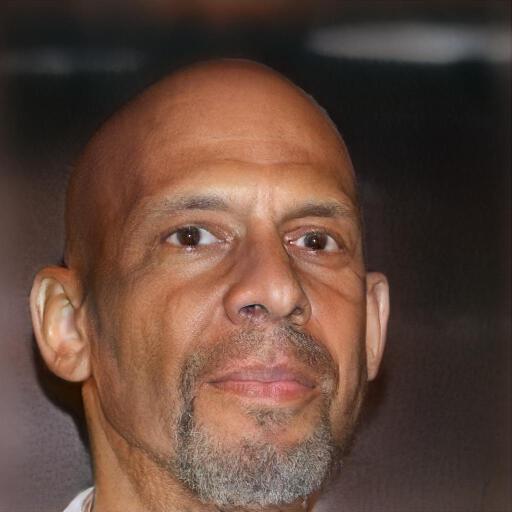}}
& \raisebox{-.5\totalheight}{\includegraphics[width=\seqSize\textwidth]{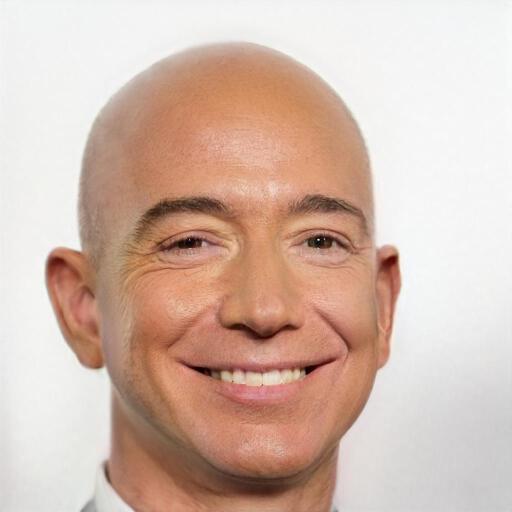}}
& \raisebox{-.5\totalheight}{\includegraphics[width=\seqSize\textwidth]{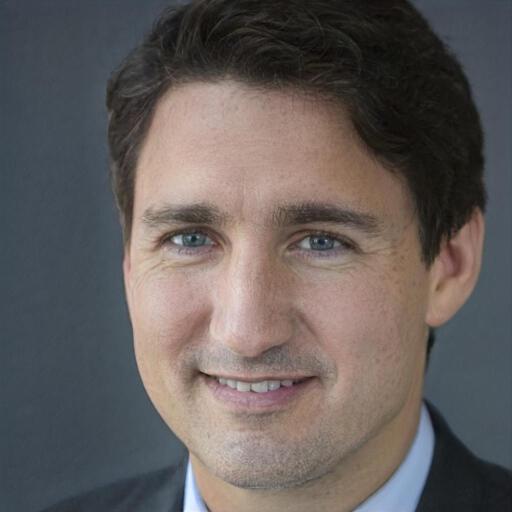}}
& \raisebox{-.5\totalheight}{\includegraphics[width=\seqSize\textwidth]{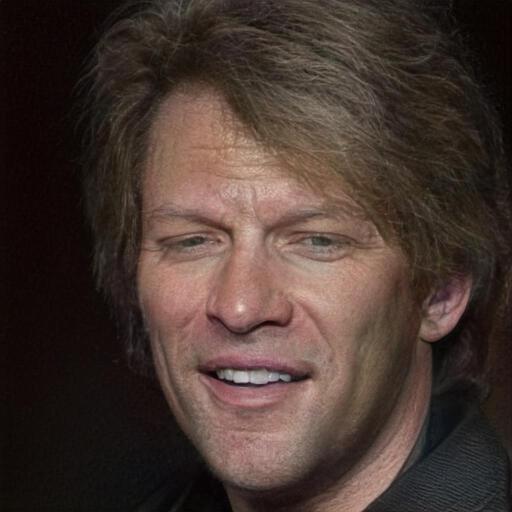}} \\

\end{tabular}
\vspace{0.1cm}
\caption{Sequential editing. We perform pivotal tuning inversion followed by two edits sequentially: rotation and smile.}
\label{fig:sequential_editing3}
\end{figure*}

\begin{table}
\begin{center}
\begin{tabular}{lcccc}
\toprule
Edit Magnitude &  ours & e4e & SG2 &  SG2 $\mathcal{W}+$ \\ 
\bottomrule
Pose & 14.86 & 14.6 & \textbf{15} & 11.15 \\
\toprule
\multicolumn{5}{l}{ID similarity, same edit} \\
\bottomrule  
single edits & \textbf{0.9} & $0.79$ & $0.82$ & $0.85$  \\

\midrule
sequential edits & \textbf{0.82} & $0.73$ & $0.78$ & $0.81$   \\
\toprule
\multicolumn{5}{l}{ID similarity, same magnitude} \\
\bottomrule  
 $\pm 5$ rotation & \textbf{0.84} & 0.77 & 0.79 & 0.82 \\
 \midrule
$\pm 10$ rotation & \textbf{0.78} & 0.72 & 0.75 & 0.77 \\
\end{tabular}
\end{center}
\vspace{-0.1cm}
\caption{Editing evaluation. Top: we compare the edit magnitude for the same latent edit over the different baselines, as proposed by Zhu et al.~\cite{zhu2020improved}. The conjecture is that more editable codes yield more significant change from the same latent edit. Middle rows: ID preservation is measured after editing. We have used rotation, smile, and age. We report the mean ID correlation for the different edits (single edits), as well as the ID preservation after applying all three edits sequentially (sequential edits). Bottom: Id preservation when applying an edit that yields the same effect for all baselines. The yaw angle change is measured by Microsoft Face API~\cite{azure}. As can be seen, our editability is similar to $\mathcal{W}$-based inversion, while our identity preservation is better even than $\mathcal{W}+$-based inversions.}
\label{tab:id} 
\end{table}

\subsection{Regularization}
\label{sec:reg}

\begin{figure}
    
\centering
\begin{tabular}{c c c}
Original & w/o Regularization & w/ regularization \\
\raisebox{-.5\totalheight}{\includegraphics[width=0.3\columnwidth]{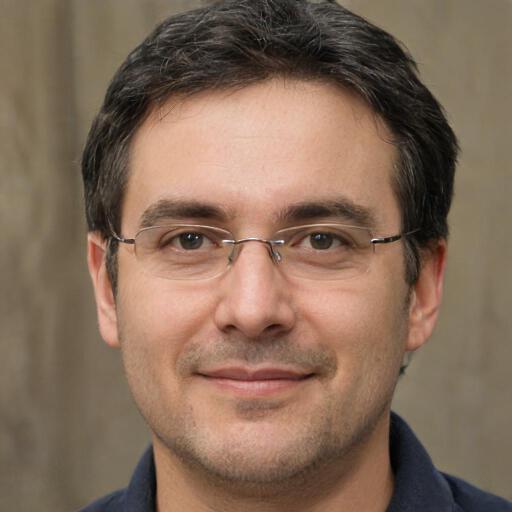}} & \raisebox{-.5\totalheight}{\includegraphics[width=0.3\columnwidth]{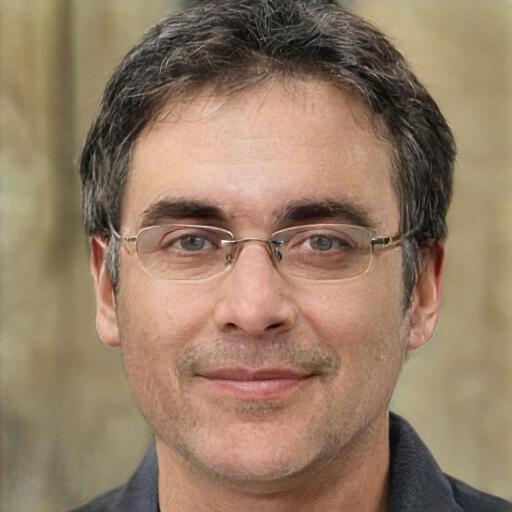}} &
\raisebox{-.5\totalheight}{\includegraphics[width=0.3\columnwidth]{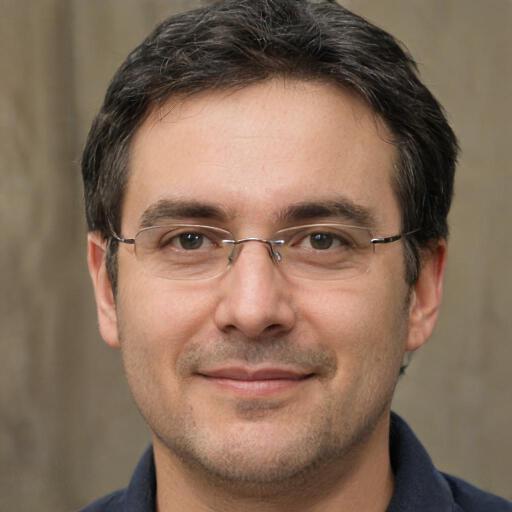}} \\
\noalign{\vskip 0.1cm}
\raisebox{-.5\totalheight}{\includegraphics[width=0.3\columnwidth]{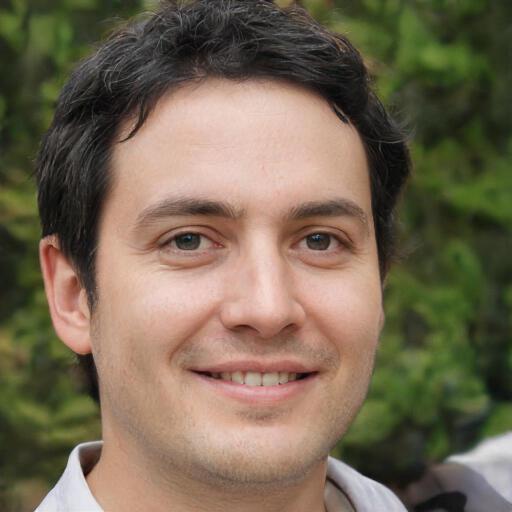}} & \raisebox{-.5\totalheight}{\includegraphics[width=0.3\columnwidth]{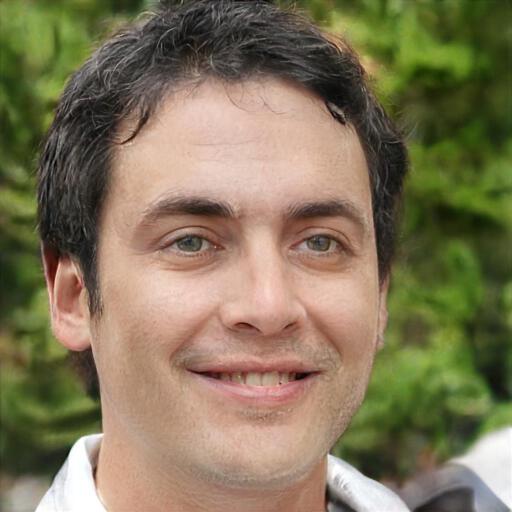}} &
\raisebox{-.5\totalheight}{\includegraphics[width=0.3\columnwidth]{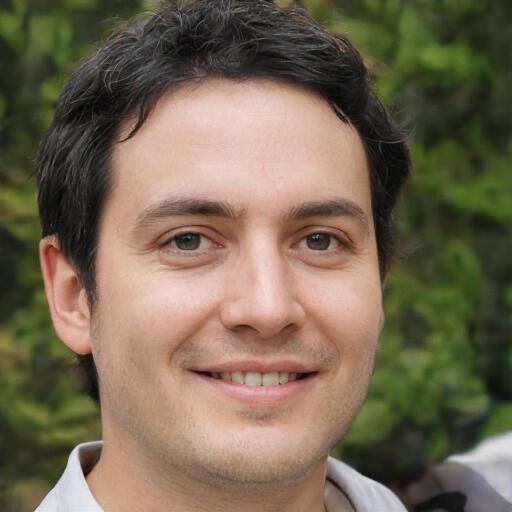}} \\
\noalign{\vskip 0.1cm}
\raisebox{-.5\totalheight}{\includegraphics[width=0.3\columnwidth]{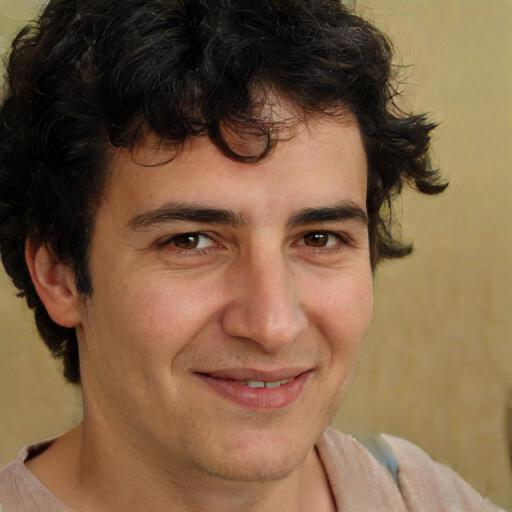}} & \raisebox{-.5\totalheight}{\includegraphics[width=0.3\columnwidth]{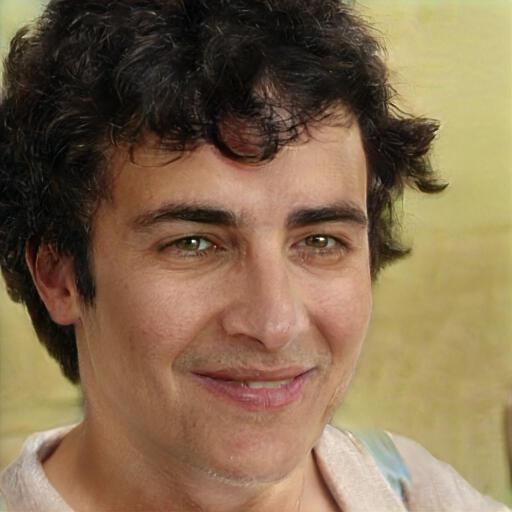}} &
\raisebox{-.5\totalheight}{\includegraphics[width=0.3\columnwidth]{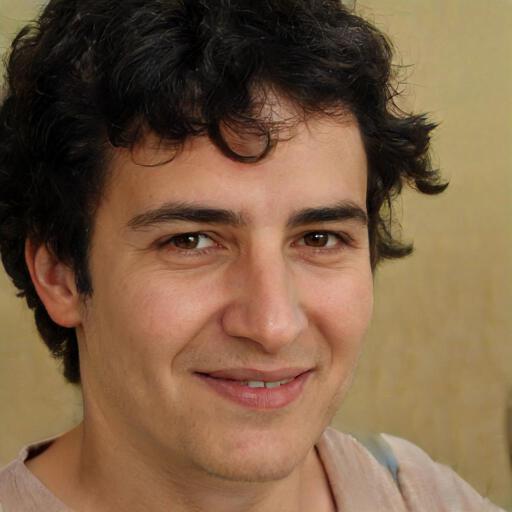}} \\
\end{tabular}
\vspace{0.1cm}
\caption{Ablation of the Locality Regularization over random latent codes for pivotal tuning applied on multiple identities. As can be seen, without regularization the generated images suffer from artifacts and ID shifts. These are almost completely removed by employing our regularization.}
\label{fig:reg_visual}
\end{figure}

\begin{figure}
    
\centering
\begin{tabular}{c}
\raisebox{-.5\totalheight}{\includegraphics[width=0.5\textwidth]{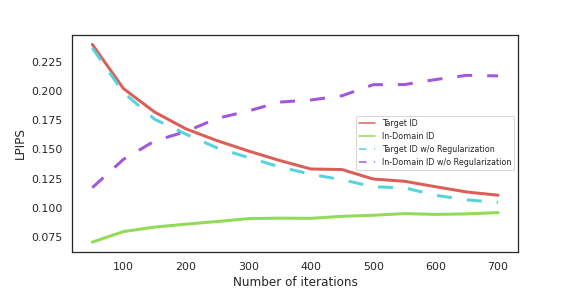}} \\
\raisebox{-.5\totalheight}{\includegraphics[width=0.5\textwidth]{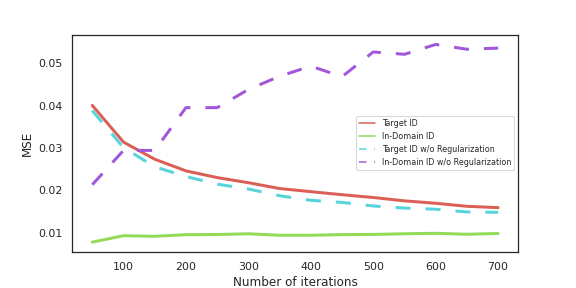}}
\end{tabular}
\vspace{0.1cm}
\caption{Quantitative evaluation of the locality regularization. To measure the magnitude of the side effect caused by pivotal tuning, we sample random latent codes, denoted In-Domain, and evaluate the reconstruction, measured by MSE and LPIPS, compared to the original pretrained generator. Our regularization reaches similar target reconstruction scores while reducing the artifacts caused to the entire domain substantially.}
\label{fig:Iter_Graph}
\end{figure}

Our locality regularization restricts the pivotal tuning side effects, causing diminishing disturbance to distant latent codes. We evaluate this effect by sampling random latent codes and comparing their generated images between the original and tuned generators. Visual results, presented in Figure~\ref{fig:reg_visual}, demonstrate that the regularization significantly minimizes the change. The images generated without regularization suffer from artifacts and ID shifting, while the images generated while employing the regularization are almost identical to the original ones. We perform the regularization evaluation using a model tuned to invert $12$ identities, as the side effects are more substantial in the multiple identities case. In addition, Figure~\ref{fig:Iter_Graph} presents quantitative results. We measure the reconstruction of random latent codes with and without the regularization compared to using the original pretrained generator. To demonstrate that our regularization does not decrease the pivotal tuning results, we also measure the reconstruction of the target image. As can be seen, our regularization reduces the side effects significantly while obtaining similar reconstruction for the target image.

\subsection{Ablation study}
\label{sec:abl}

An ablation analysis is presented in Figure~\ref{fig:ablation}. First, we show that using a pivot latent code from $\mathcal{W}+$ space rather of $\mathcal{W}$ ($(B)$), results in less editability, as the editing is less meaningful for both smile and pose. Skipping the initial inversion step and using the mean latent code ($(C)$) or a random latent code ($(E)$), results in substantially more distortion compared to ours. Similar results were obtained by optimizing the pivot latent code in addition to the generator, initialized to mean latent code ($(D)$) or random latent code ($(F)$) similar to Pan et al.~\cite{pan2020exploiting}. In addition, we demonstrate that optimizing the pivot latent code is not necessary even when starting from an inverted pivot code. To do this, we start from an inverted code $w_p$ and perform PTI while allowing $w_p$ to change. We then feed the resulting code $\Tilde{w_p}$ back to the original StyleGAN. Inspecting the two images, produced by $w_p$ and $\Tilde{w_p}$ over the same generator, we see negligible change: $0.015 \pm 5e^{-6}$ for LPIPS and $0.0012 \pm 1e^{-6}$ for MSE. We conclude that our choice for pivot code is almost optimal and hence can lighten the optimization process by keeping the code fixed.

\newcommand{\ablationSize}{0.12}

\begin{figure*}

\centering
\begin{tabular}{cccccccc}
Original & $(A)$ & $(B)$ & $(C)$ & $(D)$ & $(E)$ & $(F)$\\
  \raisebox{-.5\totalheight}{\includegraphics[width=\ablationSize\textwidth]{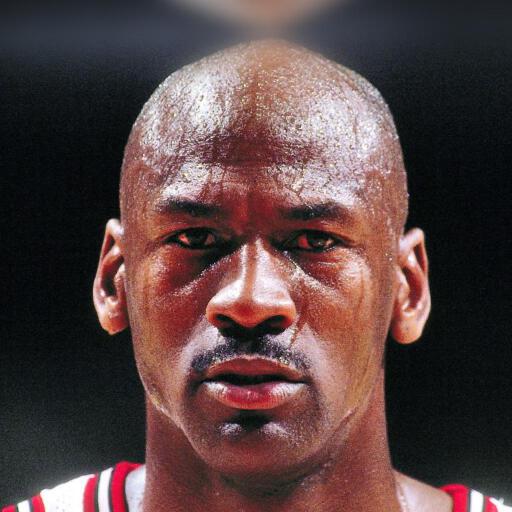}} & 
  \raisebox{-.5\totalheight}{\includegraphics[width=\ablationSize\textwidth]{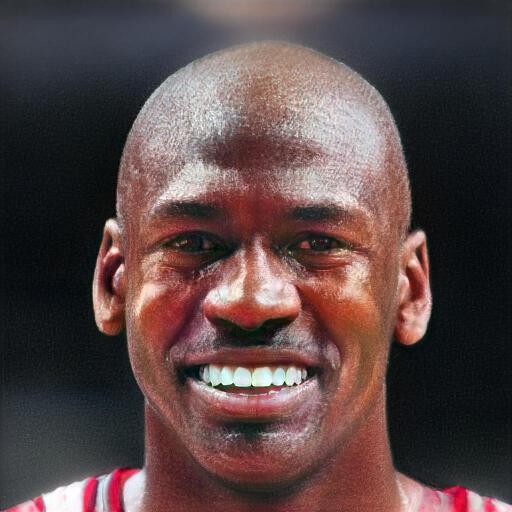}} &
 \raisebox{-.5\totalheight}{\includegraphics[width=\ablationSize\textwidth]{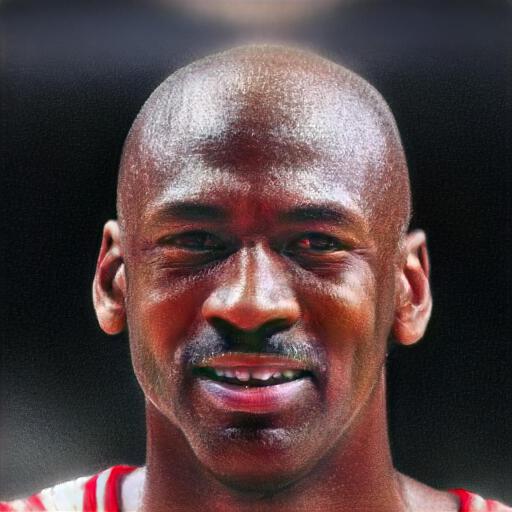}} &
  \raisebox{-.5\totalheight}{\includegraphics[width=\ablationSize\textwidth]{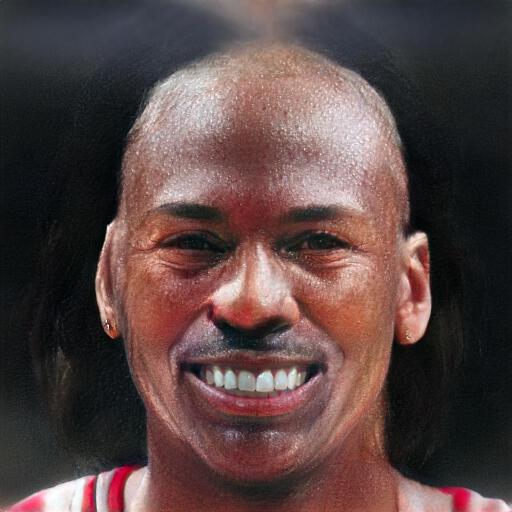}} & 
 \raisebox{-.5\totalheight}{\includegraphics[width=\ablationSize\textwidth]{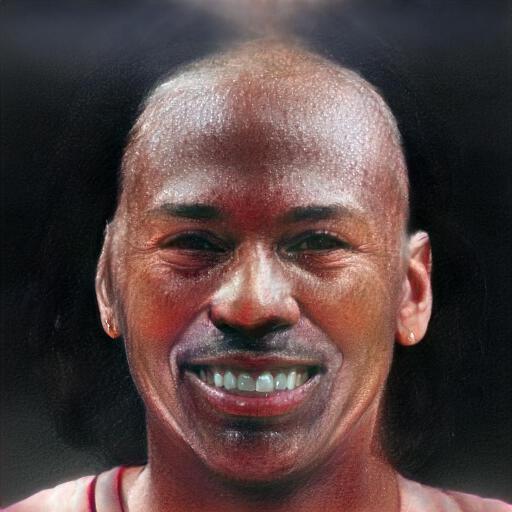}} &
 \raisebox{-.5\totalheight}{\includegraphics[width=\ablationSize\textwidth]{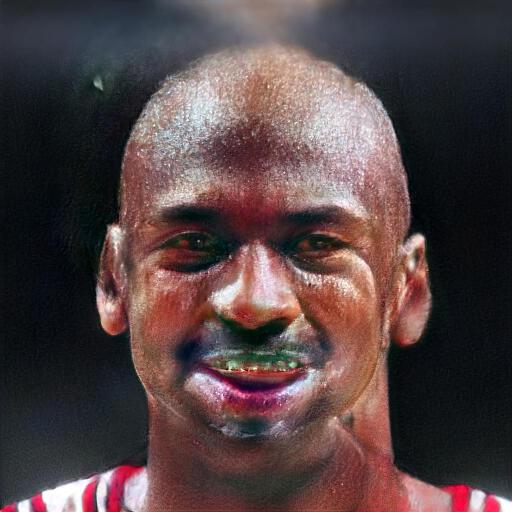}} &
 \raisebox{-.5\totalheight}{\includegraphics[width=\ablationSize\textwidth]{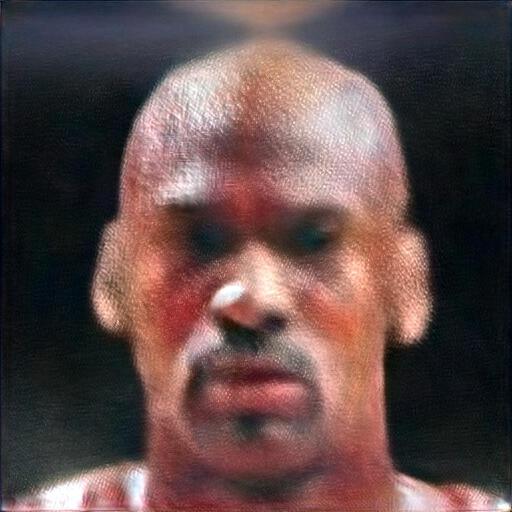}}  \\
 \noalign{\vskip 1mm}
  \raisebox{-.5\totalheight}{\includegraphics[width=\ablationSize\textwidth]{interface_results/jordan/jordan.jpg}} & 
  \raisebox{-.5\totalheight}{\includegraphics[width=\ablationSize\textwidth]{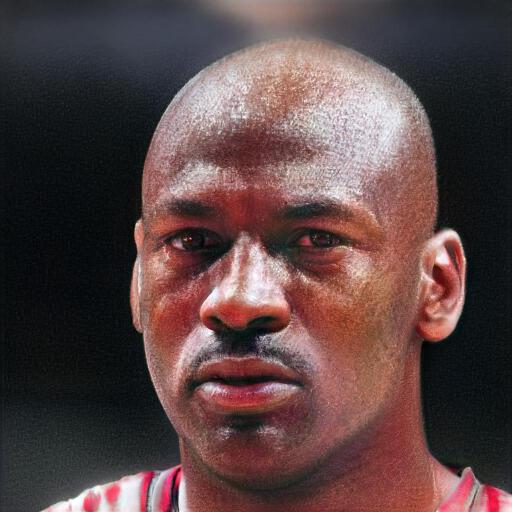}} &
   \raisebox{-.5\totalheight}{\includegraphics[width=\ablationSize\textwidth]{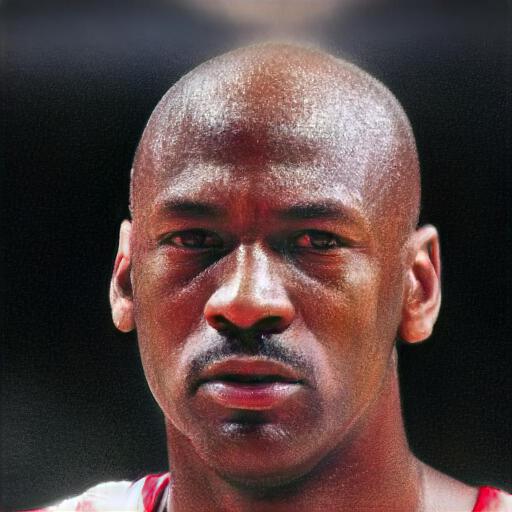}} &
  \raisebox{-.5\totalheight}{\includegraphics[width=\ablationSize\textwidth]{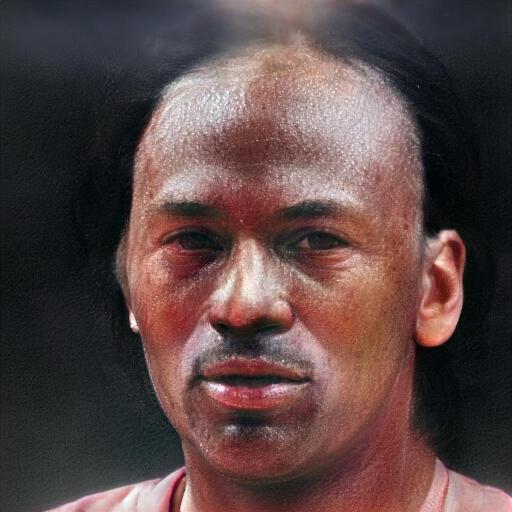}} &
 \raisebox{-.5\totalheight}{\includegraphics[width=\ablationSize\textwidth]{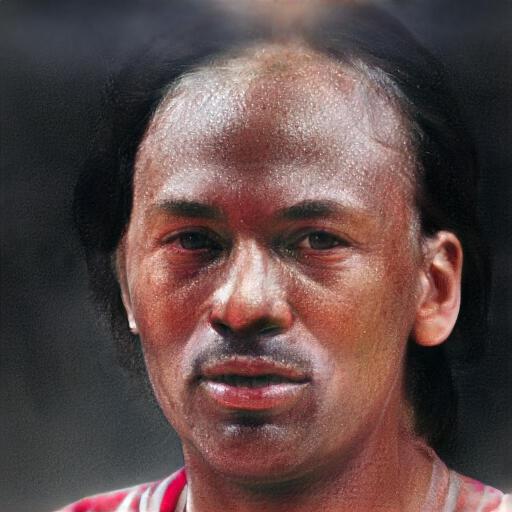}} & 
  \raisebox{-.5\totalheight}{\includegraphics[width=\ablationSize\textwidth]{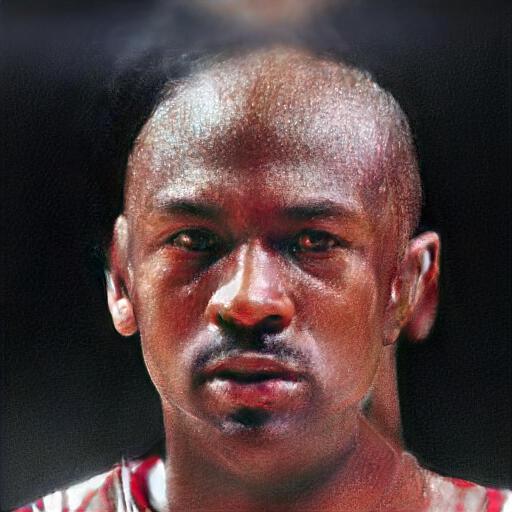}} &
 \raisebox{-.5\totalheight}{\includegraphics[width=\ablationSize\textwidth]{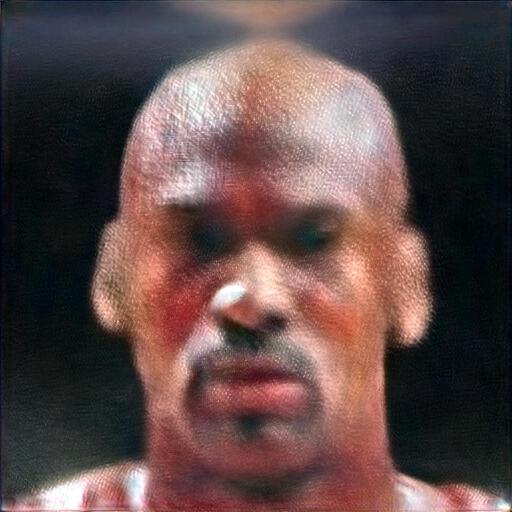}}   \\

\end{tabular}
\vspace{0.1cm}
\caption{Ablation study. We apply the same edits of smile (top) and pose (bottom).  $(A)$ full approach. $(B)$ We invert to $\mathcal{W}+$ space at the first step instead of $\mathcal{W}$, which yields inferior editability. $(C)$ the pivot latent code $w_p$ is replaced with the mean latent code $\mu_w$. $(D)$ the pivot $w_p$ is replaced with a random latent code. $(E)$ We optimize the pivot latent code $w_p$ along with the generator, initialized to the mean latent code $\mu_w$.  $(F)$ We optimize the pivot latent code $w_p$ along with the generator, initialized to a random latent code. As can be seen, $(C) - (F)$ results in significantly more distortion.} 
\label{fig:ablation}
\end{figure*}

\subsection{Implementation details}
For the initial inversion step, we use the same hyperparameters as described by Karras et al.~\cite{karras2020analyzing}, except for the learning rate which is changed to $5e^{-3}$. We run the inversion for $450$ iterations. Then, for pivotal tuning, we further optimize for $350$ iterations with a learning rate of $3e^{-4}$ using the Adam \cite{Kingma2015AdamAM} optimizer. For reconstruction, we use $\lambda_{L_2}=1$ and $\lambda_{LPIPS}=1$ and for the regularization we use $\alpha=30$, $\lambda_{L_2}^{R}=1$, $\lambda_{R}=0.1$, and $N_r = 1$.

All quantitative experiments were performed on the first 1000 samples from the CelebA-HQ test set.

Our two-step inversion takes less than 3 minutes on a single Nvidia GeForce RTX 2080. The initial $\mathcal{W}$-space inversion step takes approximately one minute, just like the SG2 inversion does. The pivotal tuning takes less than a minute without regularization, and less than two with it. This training time grows linearly with the number of inverted identities. The SG2 $\mathcal{W+}$ inversion takes $4.5$ minutes for $2100$ iterations. The inversion time of e4e is less than a second, as it is encoder-based and does not require optimization at inference.

\section{Conclusions}

We have presented Pivotal Tuning Inversion --- an inversion method that allows using latent-based editing techniques on practical, real-life facial images. In a sense, we break the notorious trade-off between reconstruction and editability through personalization, or in other words through surgical adjustments to the generator that address the desired image specifically well. This is achieved by leveraging the disentanglement between appearance and geometry that naturally emerges from StyleGAN's behavior.

In other words, we have demonstrated increased quality at the cost of additional computation. As it turns out, this deal is quite lucrative: Our PTI optimization boosts performance considerably, while entailing a computation cost of around three minutes to incorporate a new identity --- similar to what some of the current optimization-based inversion methods require. 
Furthermore, we have shown that PTI can be successfully applied to several individuals. We envision this mode of editing sessions to apply, for example, to a casting team of a movie.

Nevertheless, it is still desirable to develop a trainable mapper that approximates the PTI in a short forward pass. This would diminish the current low computational cost that entails real image editing, situating StyleGAN as a practical and accessible facial editing tool for the masses. In addition to a single-pass PTI process, in the future we plan also to consider using a set of photographs of the individual for PTI. This would extend and stabilize the notion of personalization of the target individual, compared to seeing just a single example. Another research direction is to take PTI beyond the architecture of StyleGAN, for example to BigGAN~\cite{brock2018large} or other novel generative models.

In general, we believe the presented approach of ad-hoc fine tuning a pretrained generator potentially bears merits for many other applications in editing and manipulations of specific images or other generation-based tasks in Machine Learning.

\section*{Acknowledgements}
We thank Or Patashnik, Rinon Gal and Dani Lischinski for their help and useful suggestions.

\newpage

{\small
\bibliography{egbib}

\begin{thebibliography}{46}
\providecommand{\natexlab}[1]{#1}
\providecommand{\url}[1]{\texttt{#1}}
\expandafter\ifx\csname urlstyle\endcsname\relax
  \providecommand{\doi}[1]{doi: #1}\else
  \providecommand{\doi}{doi: \begingroup \urlstyle{rm}\Url}\fi

\bibitem[Abdal et~al.(2019)Abdal, Qin, and Wonka]{abdal2019image2stylegan}
Rameen Abdal, Yipeng Qin, and Peter Wonka.
\newblock Image2stylegan: How to embed images into the stylegan latent space?
\newblock In \emph{Proceedings of the IEEE international conference on computer
  vision}, pages 4432--4441, 2019.

\bibitem[Abdal et~al.(2020{\natexlab{a}})Abdal, Qin, and
  Wonka]{abdal2020image2stylegan++}
Rameen Abdal, Yipeng Qin, and Peter Wonka.
\newblock Image2stylegan++: How to edit the embedded images?
\newblock In \emph{Proceedings of the IEEE/CVF Conference on Computer Vision
  and Pattern Recognition}, pages 8296--8305, 2020{\natexlab{a}}.

\bibitem[Abdal et~al.(2020{\natexlab{b}})Abdal, Zhu, Mitra, and
  Wonka]{abdal2020styleflow}
Rameen Abdal, Peihao Zhu, Niloy Mitra, and Peter Wonka.
\newblock Styleflow: Attribute-conditioned exploration of stylegan-generated
  images using conditional continuous normalizing flows, 2020{\natexlab{b}}.

\bibitem[Alaluf et~al.(2021)Alaluf, Patashnik, and Cohen{-}Or]{alaluf-restyle}
Yuval Alaluf, Or~Patashnik, and Daniel Cohen{-}Or.
\newblock Restyle: {A} residual-based stylegan encoder via iterative
  refinement.
\newblock \emph{CoRR}, abs/2104.02699, 2021.
\newblock URL \url{https://arxiv.org/abs/2104.02699}.

\bibitem[Bau et~al.(2020)Bau, Strobelt, Peebles, Wulff, Zhou, Zhu, and
  Torralba]{bau2020semantic}
David Bau, Hendrik Strobelt, William Peebles, Jonas Wulff, Bolei Zhou, Jun-Yan
  Zhu, and Antonio Torralba.
\newblock Semantic photo manipulation with a generative image prior.
\newblock \emph{arXiv preprint arXiv:2005.07727}, 2020.

\bibitem[Brock et~al.(2018)Brock, Donahue, and Simonyan]{brock2018large}
Andrew Brock, Jeff Donahue, and Karen Simonyan.
\newblock Large scale gan training for high fidelity natural image synthesis.
\newblock In \emph{International Conference on Learning Representations}, 2018.

\bibitem[Collins et~al.(2020)Collins, Bala, Price, and
  Susstrunk]{collins2020editing}
Edo Collins, Raja Bala, Bob Price, and Sabine Susstrunk.
\newblock Editing in style: Uncovering the local semantics of gans.
\newblock In \emph{Proceedings of the IEEE/CVF Conference on Computer Vision
  and Pattern Recognition}, pages 5771--5780, 2020.

\bibitem[Creswell and Bharath(2018)]{creswell2018inverting}
Antonia Creswell and Anil~Anthony Bharath.
\newblock Inverting the generator of a generative adversarial network.
\newblock \emph{IEEE transactions on neural networks and learning systems},
  30\penalty0 (7):\penalty0 1967--1974, 2018.

\bibitem[Deng et~al.(2019)Deng, Guo, Xue, and Zafeiriou]{deng2019arcface}
Jiankang Deng, Jia Guo, Niannan Xue, and Stefanos Zafeiriou.
\newblock Arcface: Additive angular margin loss for deep face recognition.
\newblock In \emph{Proceedings of the IEEE Conference on Computer Vision and
  Pattern Recognition}, pages 4690--4699, 2019.

\bibitem[Denton et~al.(2019)Denton, Hutchinson, Mitchell, and
  Gebru]{denton2019detecting}
Emily Denton, Ben Hutchinson, Margaret Mitchell, and Timnit Gebru.
\newblock Detecting bias with generative counterfactual face attribute
  augmentation.
\newblock \emph{arXiv preprint arXiv:1906.06439}, 2019.

\bibitem[Goetschalckx et~al.(2019)Goetschalckx, Andonian, Oliva, and
  Isola]{goetschalckx2019ganalyze}
Lore Goetschalckx, Alex Andonian, Aude Oliva, and Phillip Isola.
\newblock Ganalyze: Toward visual definitions of cognitive image properties,
  2019.

\bibitem[Goodfellow et~al.(2014)Goodfellow, Pouget-Abadie, Mirza, Xu,
  Warde-Farley, Ozair, Courville, and Bengio]{Goodfellow2014GenerativeAN}
Ian~J. Goodfellow, Jean Pouget-Abadie, Mehdi Mirza, Bing Xu, David
  Warde-Farley, Sherjil Ozair, Aaron Courville, and Yoshua Bengio.
\newblock Generative adversarial nets.
\newblock In \emph{Proceedings of the 27th International Conference on Neural
  Information Processing Systems - Volume 2}, NIPS’14, page 2672–2680,
  Cambridge, MA, USA, 2014. MIT Press.

\bibitem[Guan et~al.(2020)Guan, Tai, Ni, Zhu, Huang, and
  Yang]{guan2020collaborative}
Shanyan Guan, Ying Tai, Bingbing Ni, Feida Zhu, Feiyue Huang, and Xiaokang
  Yang.
\newblock Collaborative learning for faster stylegan embedding.
\newblock \emph{arXiv preprint arXiv:2007.01758}, 2020.

\bibitem[H{\"a}rk{\"o}nen et~al.(2020)H{\"a}rk{\"o}nen, Hertzmann, Lehtinen,
  and Paris]{harkonen2020ganspace}
Erik H{\"a}rk{\"o}nen, Aaron Hertzmann, Jaakko Lehtinen, and Sylvain Paris.
\newblock Ganspace: Discovering interpretable gan controls.
\newblock \emph{arXiv preprint arXiv:2004.02546}, 2020.

\bibitem[Jahanian et~al.(2019)Jahanian, Chai, and
  Isola]{jahanian2019steerability}
Ali Jahanian, Lucy Chai, and Phillip Isola.
\newblock On the''steerability" of generative adversarial networks.
\newblock \emph{arXiv preprint arXiv:1907.07171}, 2019.

\bibitem[Karras et~al.(2017)Karras, Aila, Laine, and
  Lehtinen]{karras2017progressive}
Tero Karras, Timo Aila, Samuli Laine, and Jaakko Lehtinen.
\newblock Progressive growing of gans for improved quality, stability, and
  variation.
\newblock \emph{arXiv preprint arXiv:1710.10196}, 2017.

\bibitem[Karras et~al.(2019)Karras, Laine, and Aila]{karras2019style}
Tero Karras, Samuli Laine, and Timo Aila.
\newblock A style-based generator architecture for generative adversarial
  networks.
\newblock In \emph{Proceedings of the IEEE conference on computer vision and
  pattern recognition}, pages 4401--4410, 2019.

\bibitem[Karras et~al.(2020{\natexlab{a}})Karras, Aittala, Hellsten, Laine,
  Lehtinen, and Aila]{Karras2020ada}
Tero Karras, Miika Aittala, Janne Hellsten, Samuli Laine, Jaakko Lehtinen, and
  Timo Aila.
\newblock Training generative adversarial networks with limited data.
\newblock In \emph{Proc. NeurIPS}, 2020{\natexlab{a}}.

\bibitem[Karras et~al.(2020{\natexlab{b}})Karras, Laine, Aittala, Hellsten,
  Lehtinen, and Aila]{karras2020analyzing}
Tero Karras, Samuli Laine, Miika Aittala, Janne Hellsten, Jaakko Lehtinen, and
  Timo Aila.
\newblock Analyzing and improving the image quality of stylegan.
\newblock In \emph{Proceedings of the IEEE/CVF Conference on Computer Vision
  and Pattern Recognition}, pages 8110--8119, 2020{\natexlab{b}}.

\bibitem[Kingma and Ba(2015)]{Kingma2015AdamAM}
Diederik~P. Kingma and Jimmy Ba.
\newblock Adam: A method for stochastic optimization.
\newblock \emph{CoRR}, abs/1412.6980, 2015.

\bibitem[Lipton and Tripathi(2017)]{lipton2017precise}
Zachary~C Lipton and Subarna Tripathi.
\newblock Precise recovery of latent vectors from generative adversarial
  networks.
\newblock \emph{arXiv preprint arXiv:1702.04782}, 2017.

\bibitem[Liu et~al.(2015)Liu, Luo, Wang, and Tang]{liu2015faceattributes}
Ziwei Liu, Ping Luo, Xiaogang Wang, and Xiaoou Tang.
\newblock Deep learning face attributes in the wild.
\newblock In \emph{Proceedings of International Conference on Computer Vision
  (ICCV)}, December 2015.

\bibitem[Luo et~al.(2017)Luo, Xu, Tang, and Lv]{luo2017learning}
Junyu Luo, Yong Xu, Chenwei Tang, and Jiancheng Lv.
\newblock Learning inverse mapping by autoencoder based generative adversarial
  nets.
\newblock In \emph{International Conference on Neural Information Processing},
  pages 207--216. Springer, 2017.

\bibitem[Menon et~al.(2020)Menon, Damian, Hu, Ravi, and Rudin]{menon2020pulse}
Sachit Menon, Alexandru Damian, Shijia Hu, Nikhil Ravi, and Cynthia Rudin.
\newblock Pulse: Self-supervised photo upsampling via latent space exploration
  of generative models.
\newblock In \emph{Proceedings of the IEEE/CVF Conference on Computer Vision
  and Pattern Recognition}, pages 2437--2445, 2020.

\bibitem[Microsoft(2020)]{azure}
Microsoft.
\newblock Azure face, 2020.

\bibitem[Pan et~al.(2020)Pan, Zhan, Dai, Lin, Loy, and Luo]{pan2020exploiting}
Xingang Pan, Xiaohang Zhan, Bo~Dai, Dahua Lin, Chen~Change Loy, and Ping Luo.
\newblock Exploiting deep generative prior for versatile image restoration and
  manipulation.
\newblock In \emph{European Conference on Computer Vision}, pages 262--277.
  Springer, 2020.

\bibitem[Patashnik et~al.(2021)Patashnik, Wu, Shechtman, Cohen-Or, and
  Lischinski]{patashnik2021styleclip}
Or~Patashnik, Zongze Wu, Eli Shechtman, Daniel Cohen-Or, and Dani Lischinski.
\newblock Styleclip: Text-driven manipulation of stylegan imagery.
\newblock \emph{arXiv preprint arXiv:2103.17249}, 2021.

\bibitem[Perarnau et~al.(2016)Perarnau, Van De~Weijer, Raducanu, and
  {\'A}lvarez]{perarnau2016invertible}
Guim Perarnau, Joost Van De~Weijer, Bogdan Raducanu, and Jose~M {\'A}lvarez.
\newblock Invertible conditional gans for image editing.
\newblock \emph{arXiv preprint arXiv:1611.06355}, 2016.

\bibitem[Pidhorskyi et~al.(2020)Pidhorskyi, Adjeroh, and
  Doretto]{pidhorskyi2020adversarial}
Stanislav Pidhorskyi, Donald~A Adjeroh, and Gianfranco Doretto.
\newblock Adversarial latent autoencoders.
\newblock In \emph{Proceedings of the IEEE/CVF Conference on Computer Vision
  and Pattern Recognition}, pages 14104--14113, 2020.

\bibitem[Plumerault et~al.(2020)Plumerault, Borgne, and
  Hudelot]{plumerault2020controlling}
Antoine Plumerault, Herv{\'e}~Le Borgne, and C{\'e}line Hudelot.
\newblock Controlling generative models with continuous factors of variations.
\newblock \emph{arXiv preprint arXiv:2001.10238}, 2020.

\bibitem[Radford et~al.(2021)Radford, Kim, Hallacy, Ramesh, Goh, Agarwal,
  Sastry, Askell, Mishkin, Clark, et~al.]{radford2021learning}
Alec Radford, Jong~Wook Kim, Chris Hallacy, Aditya Ramesh, Gabriel Goh,
  Sandhini Agarwal, Girish Sastry, Amanda Askell, Pamela Mishkin, Jack Clark,
  et~al.
\newblock Learning transferable visual models from natural language
  supervision.
\newblock \emph{arXiv preprint arXiv:2103.00020}, 2021.

\bibitem[Richardson et~al.(2020)Richardson, Alaluf, Patashnik, Nitzan, Azar,
  Shapiro, and Cohen-Or]{richardson2020encoding}
Elad Richardson, Yuval Alaluf, Or~Patashnik, Yotam Nitzan, Yaniv Azar, Stav
  Shapiro, and Daniel Cohen-Or.
\newblock Encoding in style: a stylegan encoder for image-to-image translation.
\newblock \emph{arXiv preprint arXiv:2008.00951}, 2020.

\bibitem[Shen and Zhou(2020)]{shen2020closedform}
Yujun Shen and Bolei Zhou.
\newblock Closed-form factorization of latent semantics in gans.
\newblock \emph{arXiv preprint arXiv:2007.06600}, 2020.

\bibitem[Shen et~al.(2020)Shen, Gu, Tang, and Zhou]{shen2020interpreting}
Yujun Shen, Jinjin Gu, Xiaoou Tang, and Bolei Zhou.
\newblock Interpreting the latent space of gans for semantic face editing.
\newblock In \emph{Proceedings of the IEEE/CVF Conference on Computer Vision
  and Pattern Recognition}, pages 9243--9252, 2020.

\bibitem[Spingarn-Eliezer et~al.(2020)Spingarn-Eliezer, Banner, and
  Michaeli]{spingarn2020gan}
Nurit Spingarn-Eliezer, Ron Banner, and Tomer Michaeli.
\newblock Gan steerability without optimization.
\newblock \emph{arXiv preprint arXiv:2012.05328}, 2020.

\bibitem[Tewari et~al.(2020{\natexlab{a}})Tewari, Elgharib, Bharaj, Bernard,
  Seidel, P{\'e}rez, Zollh{\"o}fer, and Theobalt]{tewari2020stylerig}
Ayush Tewari, Mohamed Elgharib, Gaurav Bharaj, Florian Bernard, Hans-Peter
  Seidel, Patrick P{\'e}rez, Michael Zollh{\"o}fer, and Christian Theobalt.
\newblock Stylerig: Rigging stylegan for 3d control over portrait images.
\newblock \emph{arXiv preprint arXiv:2004.00121}, 2020{\natexlab{a}}.

\bibitem[Tewari et~al.(2020{\natexlab{b}})Tewari, Elgharib, R., Bernard,
  Seidel, Pérez, Zollhöfer, and Theobalt]{tewari2020pie}
Ayush Tewari, Mohamed Elgharib, Mallikarjun~B R., Florian Bernard, Hans-Peter
  Seidel, Patrick Pérez, Michael Zollhöfer, and Christian Theobalt.
\newblock Pie: Portrait image embedding for semantic control,
  2020{\natexlab{b}}.

\bibitem[Tov et~al.(2021)Tov, Alaluf, Nitzan, Patashnik, and
  Cohen-Or]{tov2021designing}
Omer Tov, Yuval Alaluf, Yotam Nitzan, Or~Patashnik, and Daniel Cohen-Or.
\newblock Designing an encoder for stylegan image manipulation.
\newblock \emph{arXiv preprint arXiv:2102.02766}, 2021.

\bibitem[Voynov and Babenko(2020)]{voynov2020unsupervised}
Andrey Voynov and Artem Babenko.
\newblock Unsupervised discovery of interpretable directions in the gan latent
  space.
\newblock \emph{arXiv preprint arXiv:2002.03754}, 2020.

\bibitem[Wang and Ponce(2021)]{wang2021a}
Binxu Wang and Carlos~R Ponce.
\newblock A geometric analysis of deep generative image models and its
  applications.
\newblock In \emph{International Conference on Learning Representations}, 2021.
\newblock URL \url{https://openreview.net/forum?id=GH7QRzUDdXG}.

\bibitem[Wang et~al.(2003)Wang, Simoncelli, and Bovik]{wang2003multiscale}
Zhou Wang, Eero~P Simoncelli, and Alan~C Bovik.
\newblock Multiscale structural similarity for image quality assessment.
\newblock In \emph{The Thrity-Seventh Asilomar Conference on Signals, Systems
  \& Computers, 2003}, volume~2, pages 1398--1402. Ieee, 2003.

\bibitem[Wu et~al.(2020)Wu, Lischinski, and Shechtman]{wu2020stylespace}
Zongze Wu, Dani Lischinski, and Eli Shechtman.
\newblock Stylespace analysis: Disentangled controls for stylegan image
  generation, 2020.

\bibitem[Zhang et~al.(2018)Zhang, Isola, Efros, Shechtman, and
  Wang]{zhang2018unreasonable}
Richard Zhang, Phillip Isola, Alexei~A. Efros, Eli Shechtman, and Oliver Wang.
\newblock The unreasonable effectiveness of deep features as a perceptual
  metric, 2018.

\bibitem[Zhu et~al.(2020{\natexlab{a}})Zhu, Shen, Zhao, and
  Zhou]{zhu2020domain}
Jiapeng Zhu, Yujun Shen, Deli Zhao, and Bolei Zhou.
\newblock In-domain gan inversion for real image editing.
\newblock \emph{arXiv preprint arXiv:2004.00049}, 2020{\natexlab{a}}.

\bibitem[Zhu et~al.(2016)Zhu, Kr{\"a}henb{\"u}hl, Shechtman, and
  Efros]{zhu2016generative}
Jun-Yan Zhu, Philipp Kr{\"a}henb{\"u}hl, Eli Shechtman, and Alexei~A Efros.
\newblock Generative visual manipulation on the natural image manifold.
\newblock In \emph{European conference on computer vision}, pages 597--613.
  Springer, 2016.

\bibitem[Zhu et~al.(2020{\natexlab{b}})Zhu, Abdal, Qin, and
  Wonka]{zhu2020improved}
Peihao Zhu, Rameen Abdal, Yipeng Qin, and Peter Wonka.
\newblock Improved stylegan embedding: Where are the good latents?,
  2020{\natexlab{b}}.

\end{thebibliography}
}

\clearpage
\appendix
\appendixpage



\section{Locality Regularization}
\begin{figure}[h]
    
\centering
\begin{tabular}{c}
\raisebox{-.5\totalheight}{\includegraphics[width=0.5\textwidth]{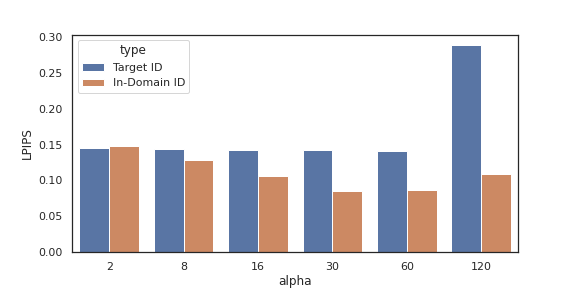}} \\ 
\raisebox{-.5\totalheight}{\includegraphics[width=0.5\textwidth]{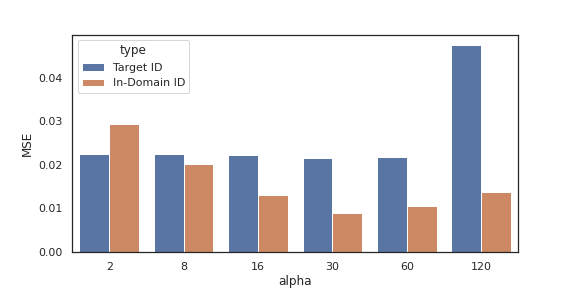}}
\end{tabular}
\vspace{0.1cm}
\caption{Quantitative evaluation of different $\alpha$ values for the locality regularization. We sample random latent codes, denoted In-Domain, and evaluate the reconstruction, measured by MSE and LPIPS, compared to the original pretrained generator. Similarly, we measure the target image reconstruction.} 

\label{fig:alpha_graph}
\end{figure}

We show the effect of different $\alpha$ values when using the locality regularization. Figure~\ref{fig:alpha_graph} presents quantitative evaluation and Figure~\ref{fig:alpha} visually demonstrates the interpolated code $w_r$. We measure both the reconstruction of the target image and the reconstruction of sampled random latent codes, denoted as \textit{in-domain}, before and after the tuning. For small $\alpha$ values (e.g., $\alpha=8$), the image generated by the interpolated code $w_r$ is very similar to the image generated by the pivot code $w_p$. Therefore, the regularization is limited and less effective. For high $\alpha$ values (e.g., $\alpha=60$) the interpolated code image is more or less equivalent to simply using a random latent $w_z$. Hence, the interpolated code is less affected by the pivotal tuning, which decreases the regularization constraint. Extremely high $\alpha$ values (e.g., $\alpha=120$) result in extremely unrealistic images, as can be seen in Figure~\ref{fig:alpha}, which cause deterioration of the target image reconstruction. Overall, we get the most effective regularization using an interpolated image which is highly similar to both pivot and the random images, e.g., $\alpha = 30$ in Figure~\ref{fig:alpha}.

\newcommand{\alphaSize}{0.1}

\begin{figure*}[h]

\begin{tabular}{cccccccc}
Pivot & Random & $\alpha = 2$ & $\alpha = 8$ & $\alpha = 16$ & $\alpha = 30$ & $\alpha = 60$ & $\alpha = 120$ \\
\raisebox{-.5\totalheight}{\includegraphics[width=\alphaSize\textwidth]{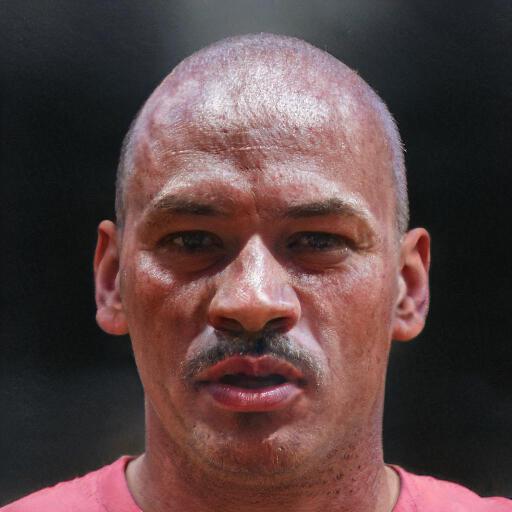}} &  \raisebox{-.5\totalheight}{\includegraphics[width=\alphaSize\textwidth]{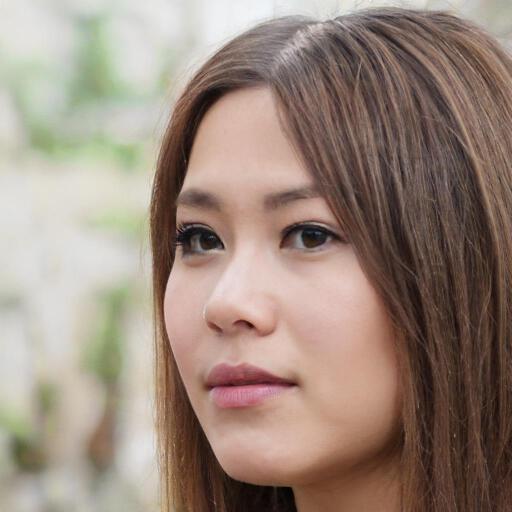}} &
\raisebox{-.5\totalheight}{\includegraphics[width=\alphaSize\textwidth]{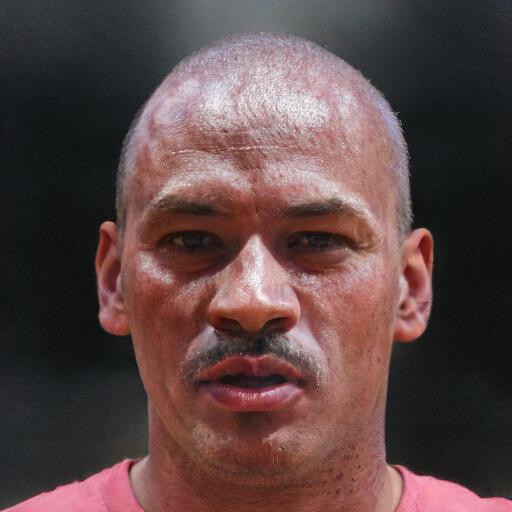}} & \raisebox{-.5\totalheight}{\includegraphics[width=\alphaSize\textwidth]{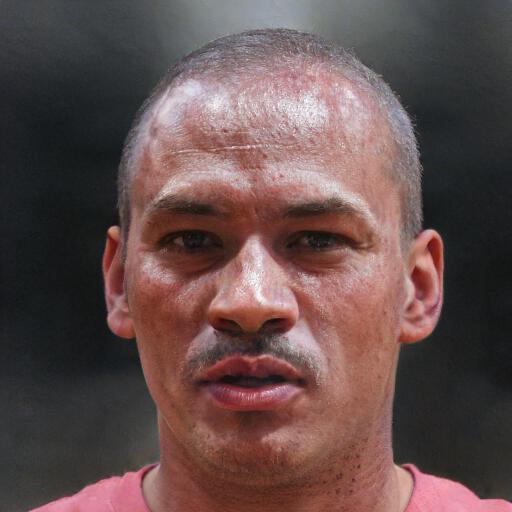}} & \raisebox{-.5\totalheight}{\includegraphics[width=\alphaSize\textwidth]{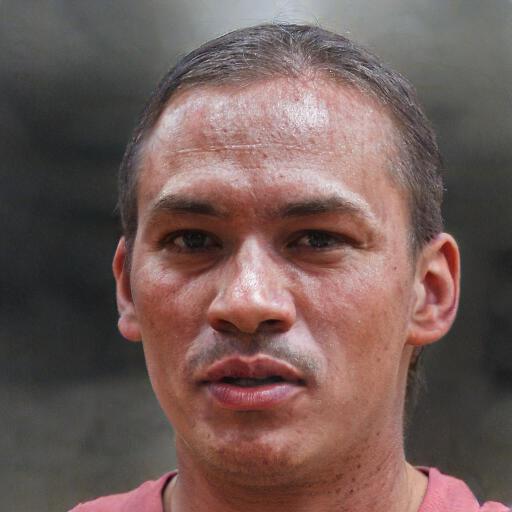}} & \raisebox{-.5\totalheight}{\includegraphics[width=\alphaSize\textwidth]{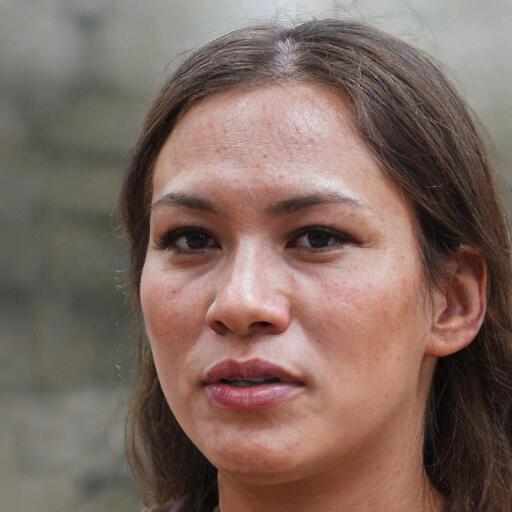}} & \raisebox{-.5\totalheight}{\includegraphics[width=\alphaSize\textwidth]{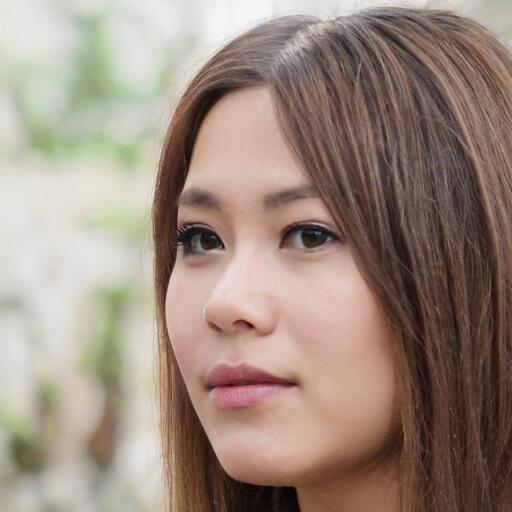}} & \raisebox{-.5\totalheight}{\includegraphics[width=\alphaSize\textwidth]{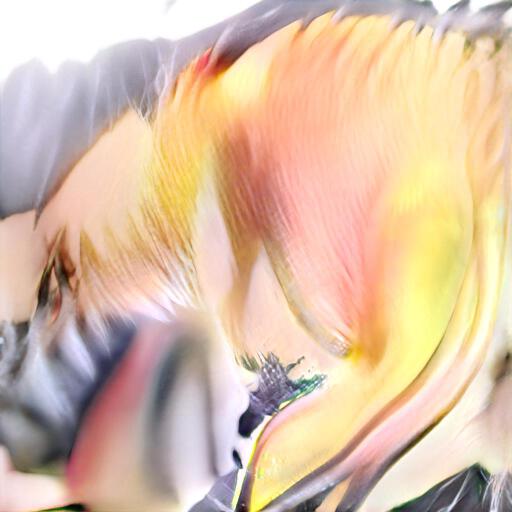}} \\

\end{tabular}
\vspace{0.15cm}
\caption{Visual demonstration of the image generated by interpolating the pivot image and a random image for different $\alpha$ values. As $\alpha$ increased, the image generated by the interpolated code is less similar to the pivot image and more to the random image, until reaching extremely high $\alpha$, where the generated image is no longer realistic. }
\label{fig:alpha}
\end{figure*}

\section{Visual Results}

\begin{figure*}[h]

\centering
\begin{tabular}{ccccccc}
\noalign{\vskip 2cm}

\rotatebox[origin=t]{90}{Real Image} & \raisebox{-.5\totalheight}{\includegraphics[width=0.13\textwidth]{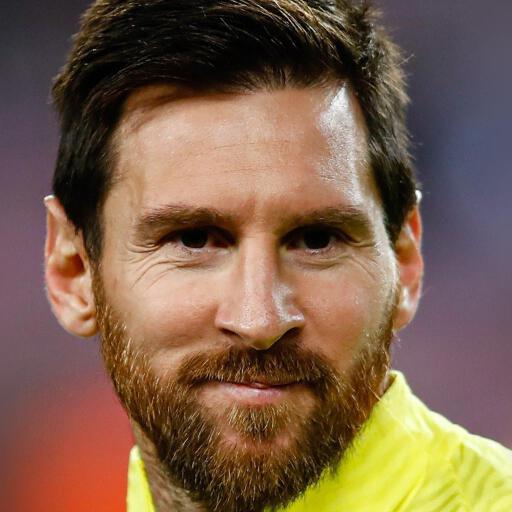}} &
\raisebox{-.5\totalheight}{\includegraphics[width=0.13\textwidth]{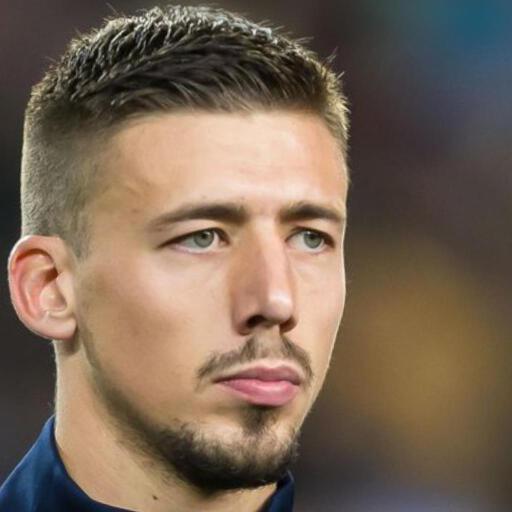}} & 
\raisebox{-.5\totalheight}{\includegraphics[width=0.13\textwidth]{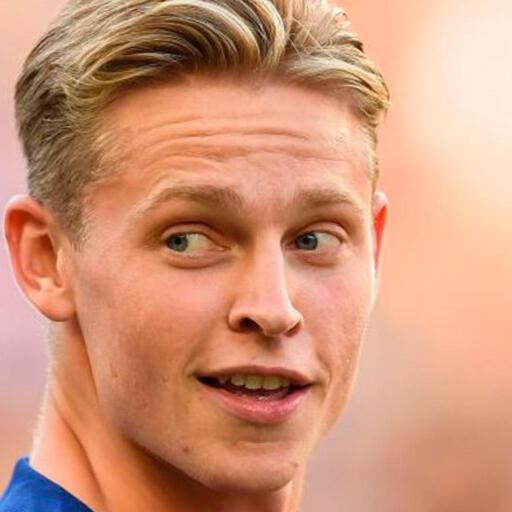}} & 
\raisebox{-.5\totalheight}{\includegraphics[width=0.13\textwidth]{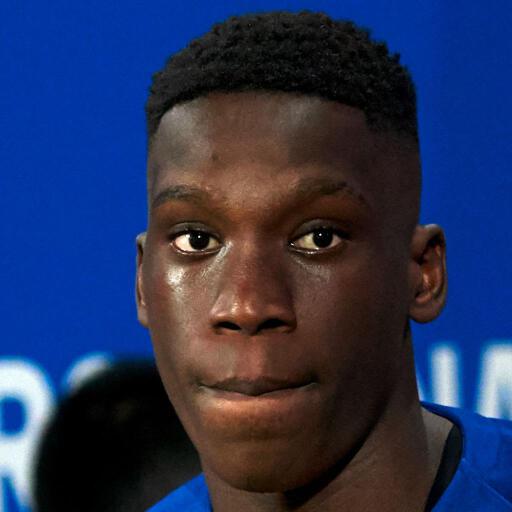}} & 
\raisebox{-.5\totalheight}{\includegraphics[width=0.13\textwidth]{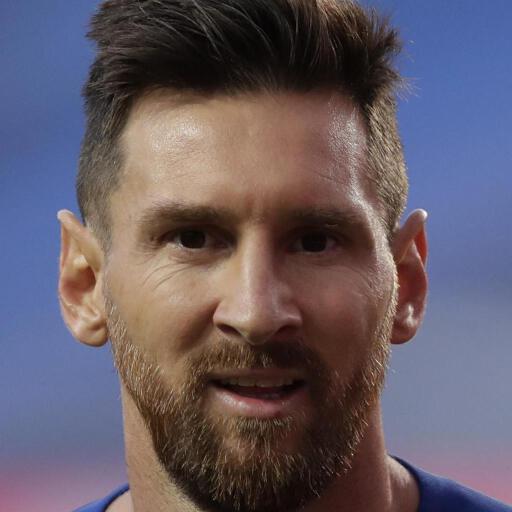}} &
\raisebox{-.5\totalheight}{\includegraphics[width=0.13\textwidth]{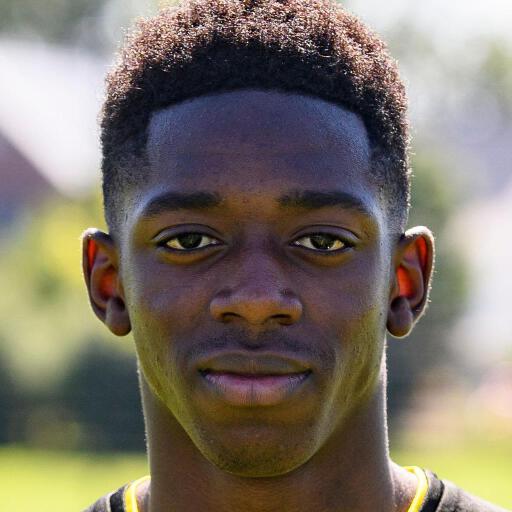}} \\
\noalign{\vskip 1mm}
\rotatebox[origin=t]{90}{Inversion} &
\raisebox{-.5\totalheight}{\includegraphics[width=0.13\textwidth]{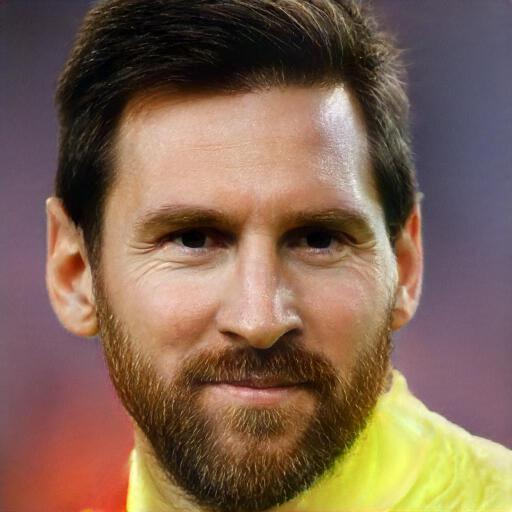}} &
\raisebox{-.5\totalheight}{\includegraphics[width=0.13\textwidth]{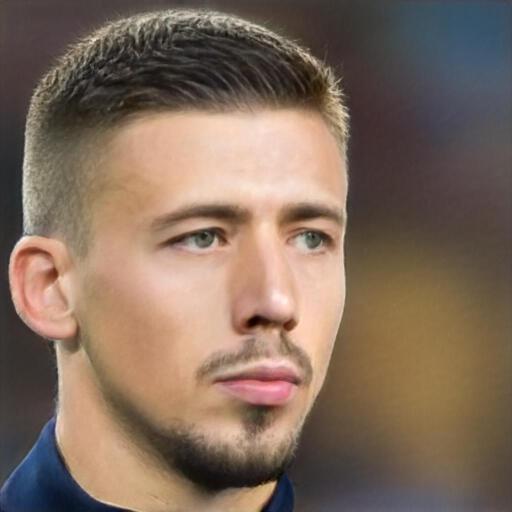}}&
\raisebox{-.5\totalheight}{\includegraphics[width=0.13\textwidth]{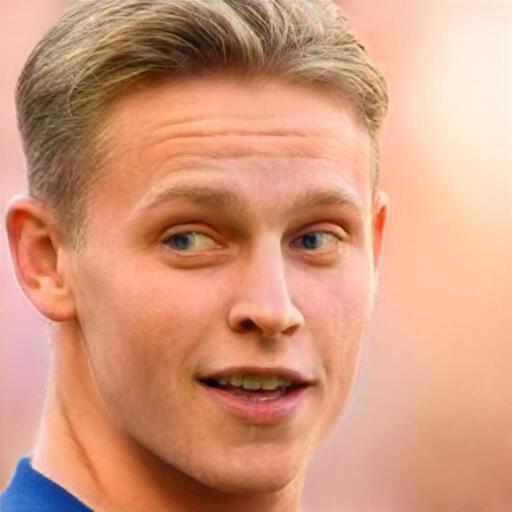}}&
\raisebox{-.5\totalheight}{\includegraphics[width=0.13\textwidth]{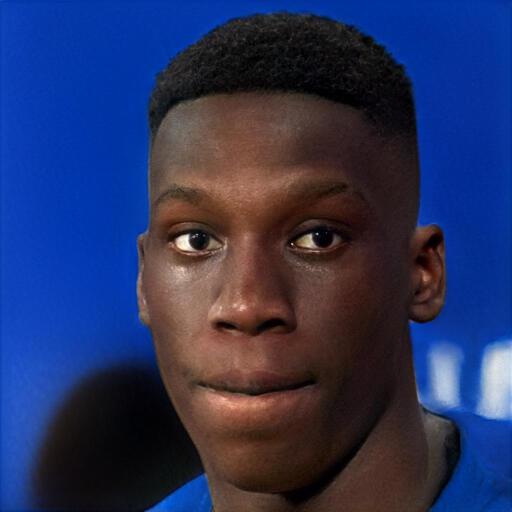}}&
\raisebox{-.5\totalheight}{\includegraphics[width=0.13\textwidth]{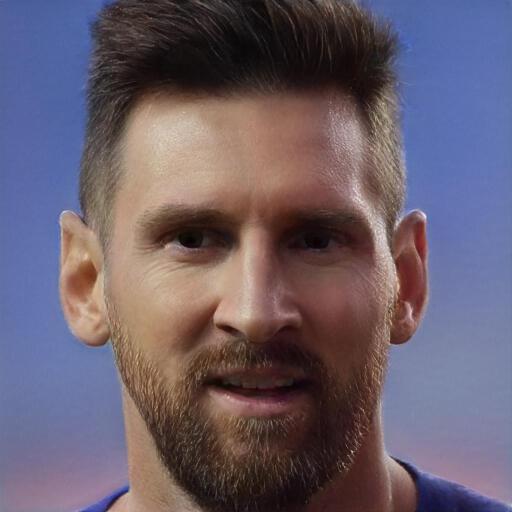}} &
\raisebox{-.5\totalheight}{\includegraphics[width=0.13\textwidth]{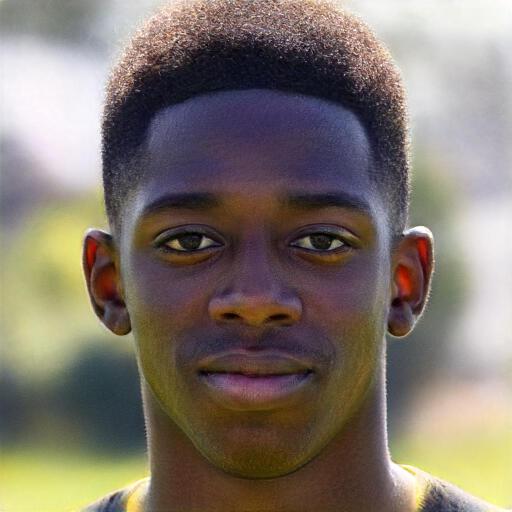}} \\ 
\noalign{\vskip 0.1cm}
\rotatebox[origin=t]{90}{Age} &
\raisebox{-.5\totalheight}{\includegraphics[width=0.13\textwidth]{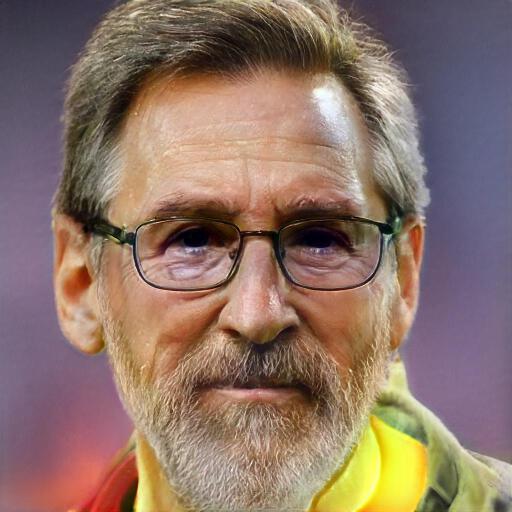}} &
\raisebox{-.5\totalheight}{\includegraphics[width=0.13\textwidth]{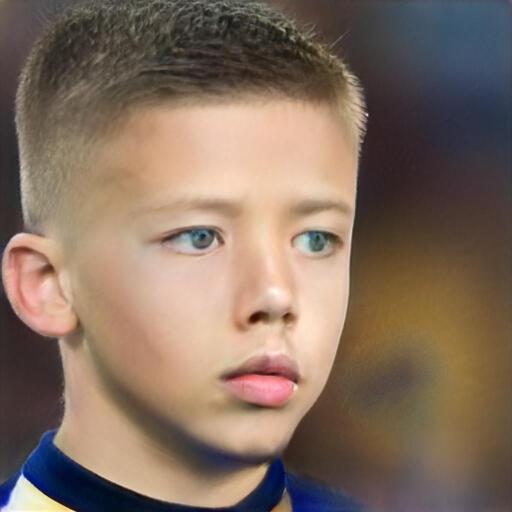}}&
\raisebox{-.5\totalheight}{\includegraphics[width=0.13\textwidth]{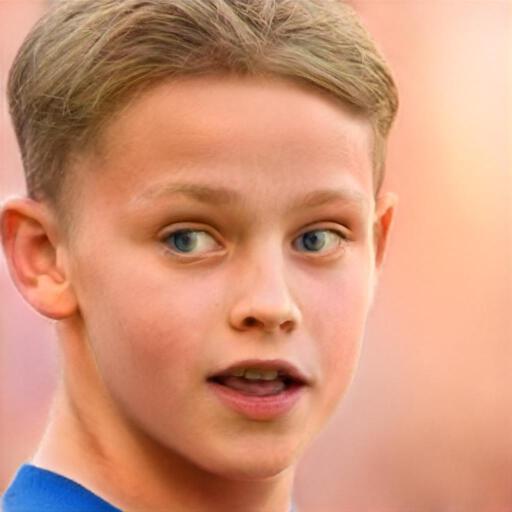}}&
\raisebox{-.5\totalheight}{\includegraphics[width=0.13\textwidth]{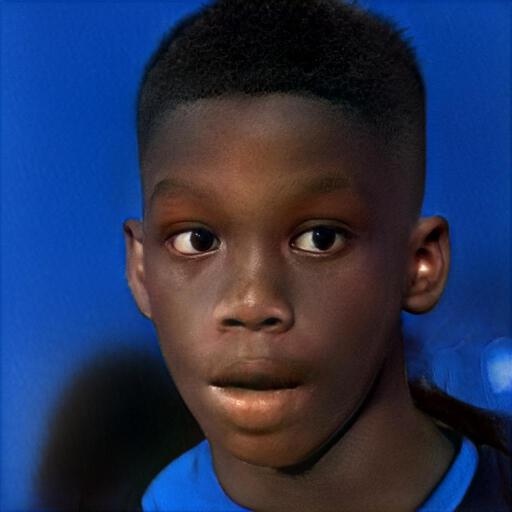}}&
\raisebox{-.5\totalheight}{\includegraphics[width=0.13\textwidth]{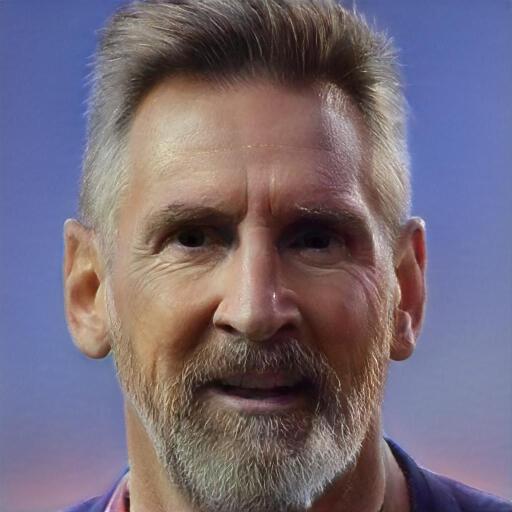}} &
\raisebox{-.5\totalheight}{\includegraphics[width=0.13\textwidth]{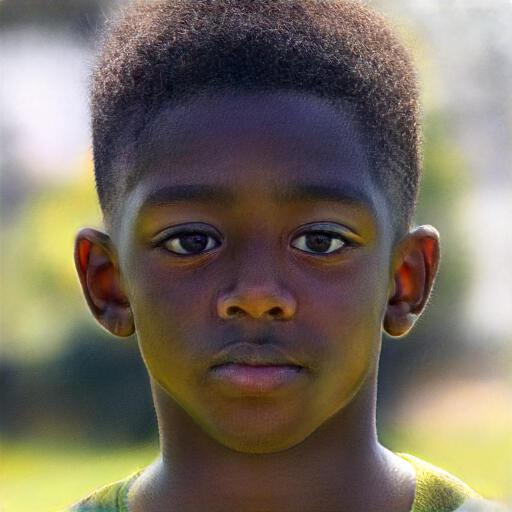}} \\
\noalign{\vskip 0.1cm}
\rotatebox[origin=t]{90}{Smile} &
\raisebox{-.5\totalheight}{\includegraphics[width=0.13\textwidth]{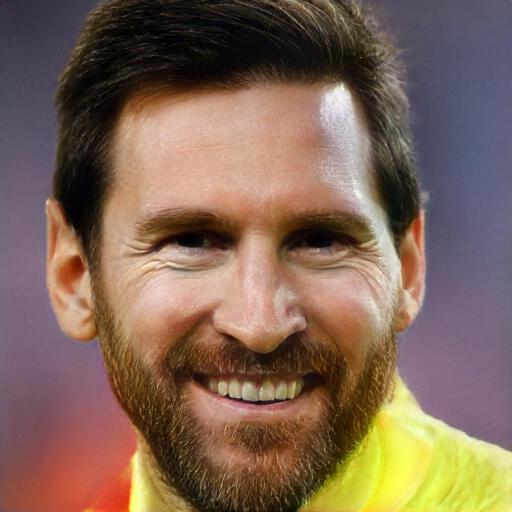}} &
\raisebox{-.5\totalheight}{\includegraphics[width=0.13\textwidth]{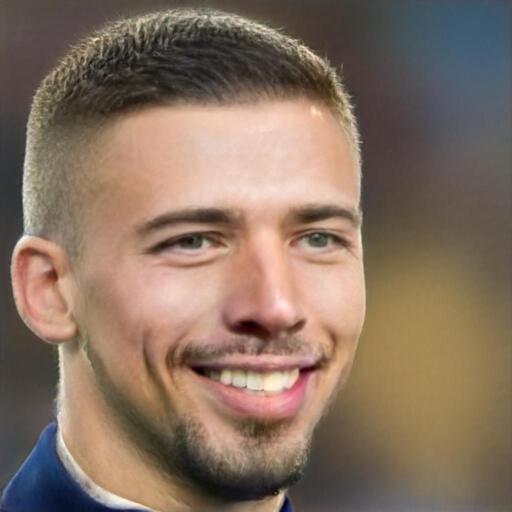}}&
\raisebox{-.5\totalheight}{\includegraphics[width=0.13\textwidth]{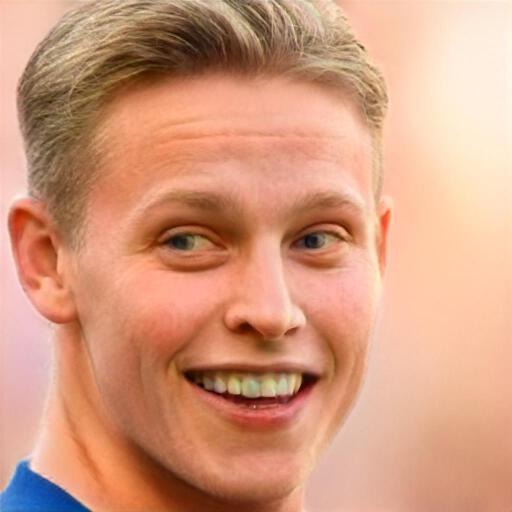}}&
\raisebox{-.5\totalheight}{\includegraphics[width=0.13\textwidth]{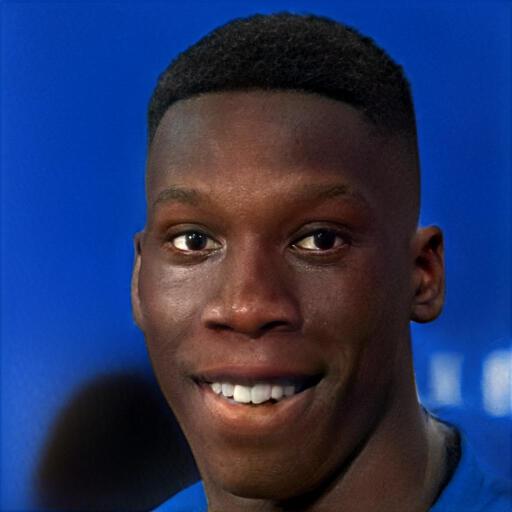}}&
\raisebox{-.5\totalheight}{\includegraphics[width=0.13\textwidth]{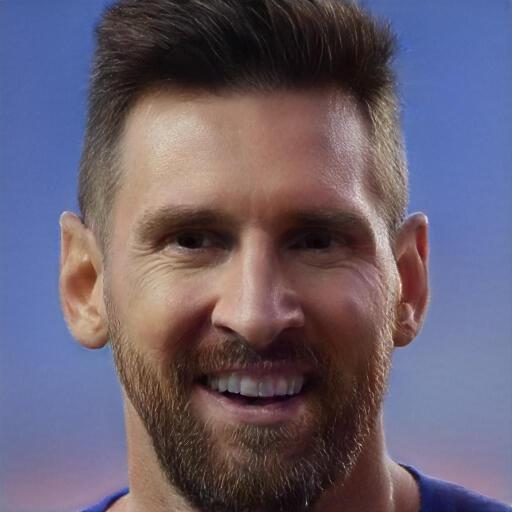}}  &
\raisebox{-.5\totalheight}{\includegraphics[width=0.13\textwidth]{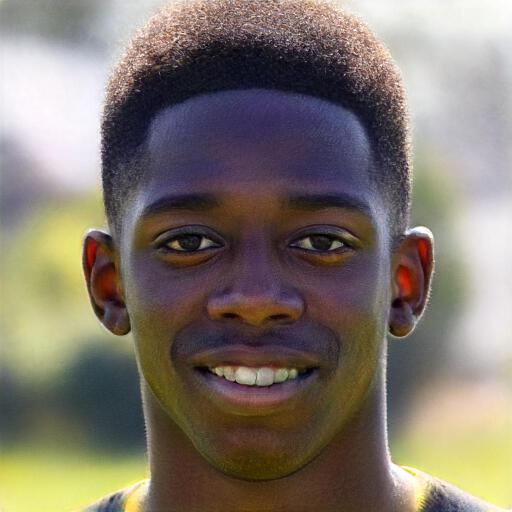}}\\
\noalign{\vskip 0.1cm}
\rotatebox[origin=t]{90}{Rotation} &
\raisebox{-.5\totalheight}{\includegraphics[width=0.13\textwidth]{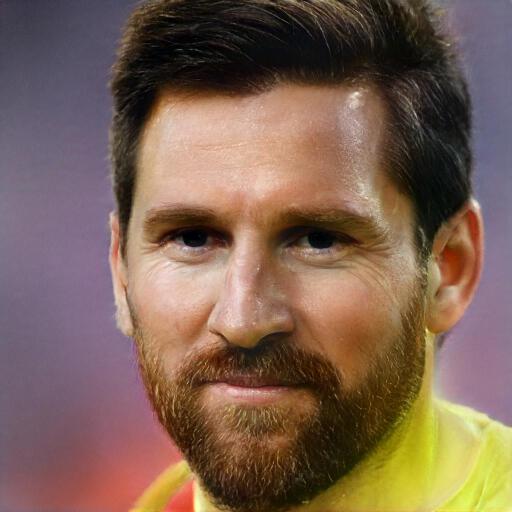}} &
\raisebox{-.5\totalheight}{\includegraphics[width=0.13\textwidth]{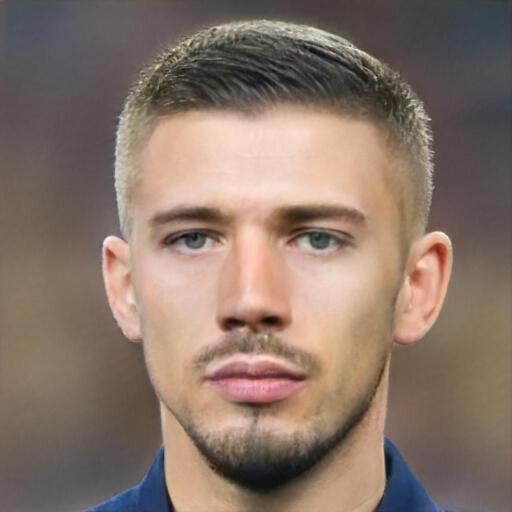}}&
\raisebox{-.5\totalheight}{\includegraphics[width=0.13\textwidth]{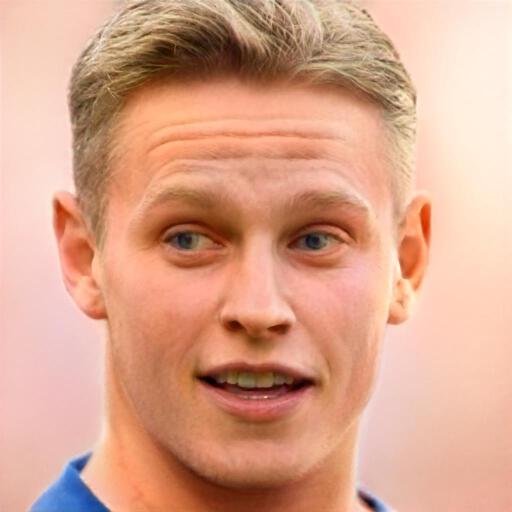}}&
\raisebox{-.5\totalheight}{\includegraphics[width=0.13\textwidth]{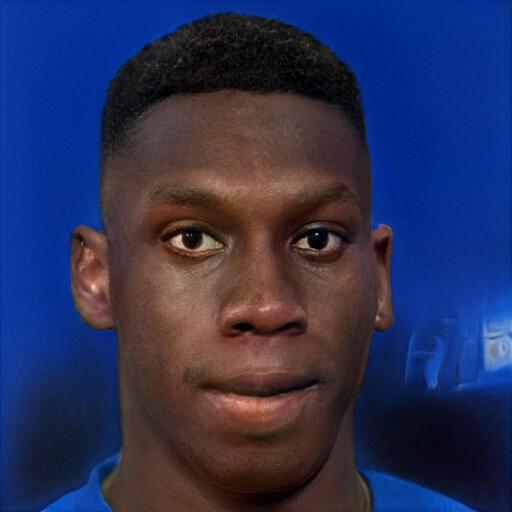}}&
\raisebox{-.5\totalheight}{\includegraphics[width=0.13\textwidth]{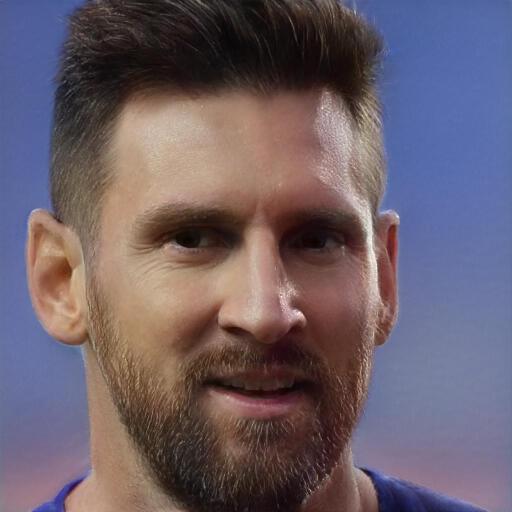}}  &
\raisebox{-.5\totalheight}{\includegraphics[width=0.13\textwidth]{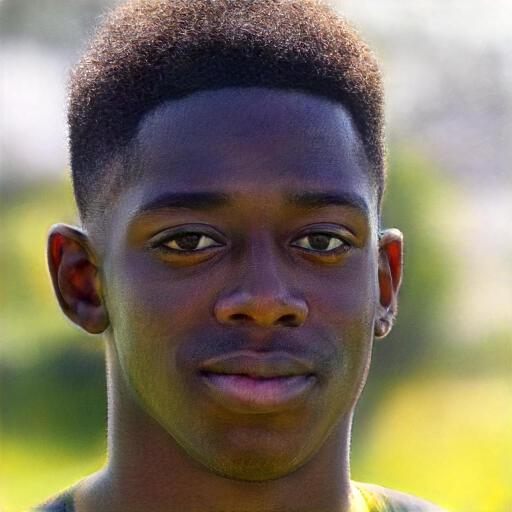}}\\
\end{tabular}
\vspace{0.15cm}
\caption{FCB-StyleGAN. We invert multiple images of Barcelona Football Club players into a single StyleGAN latent space, and demonstrate both high reconstruction and editing quality for the inverted identities. }
\label{fig:barcelona_multi_id}
\end{figure*}

\begin{figure*}[h]

\centering
\begin{tabular}{c c}
 &    \hspace{-0.6cm} \textbf{Original}  \hspace{1.7cm} \textbf{SG2 $\mathcal{W+}$} \hspace{2cm} \textbf{e4e}  \hspace{2.3cm} \textbf{SG2} \hspace{2.2cm} \textbf{Ours} \\
\rotatebox[origin=t]{90}{Inversion} &
\raisebox{-.5\totalheight}{\includegraphics[width=0.9\textwidth]{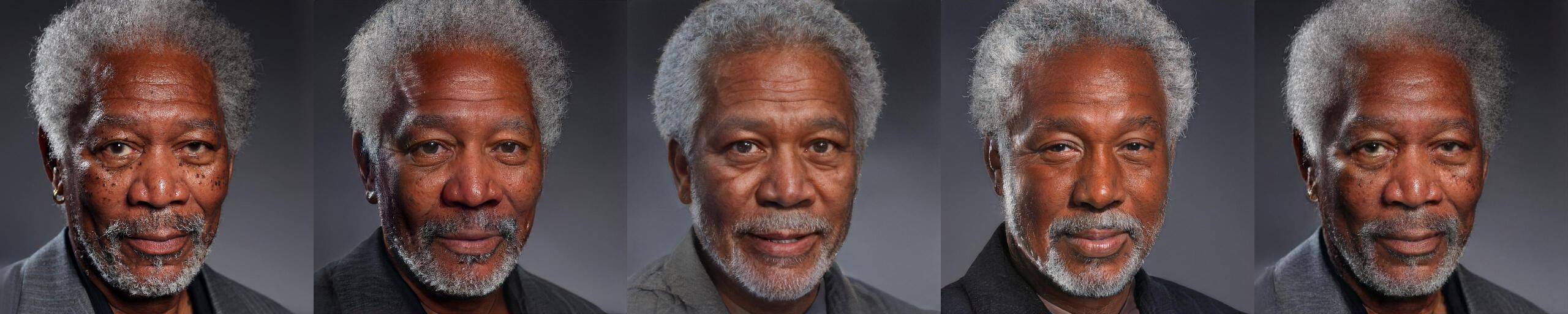}}  \\
\noalign{\vskip 1mm}
\rotatebox[origin=t]{90}{Old} &
\raisebox{-.5\totalheight}{\includegraphics[width=0.9\textwidth]{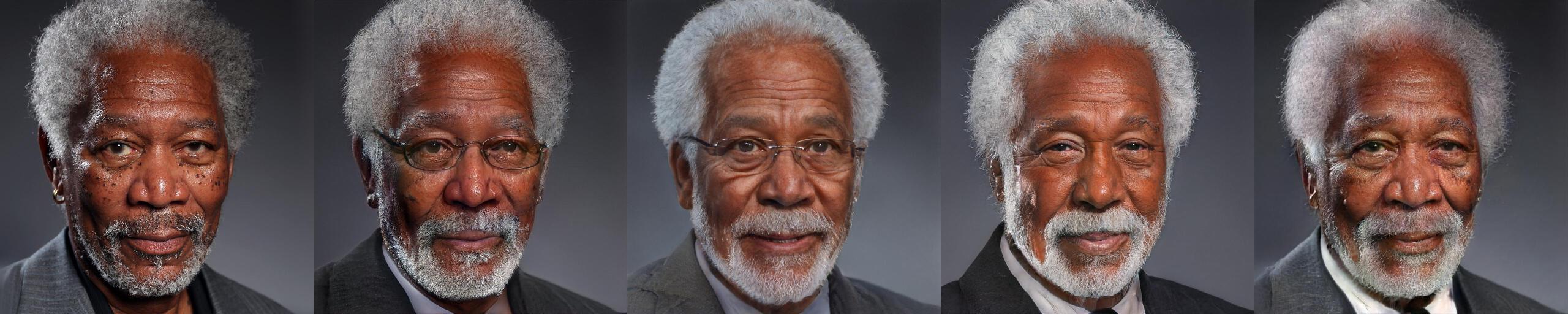}} \\ 
\noalign{\vskip 0.1cm}
\rotatebox[origin=t]{90}{Young} &
\raisebox{-.5\totalheight}{\includegraphics[width=0.9\textwidth]{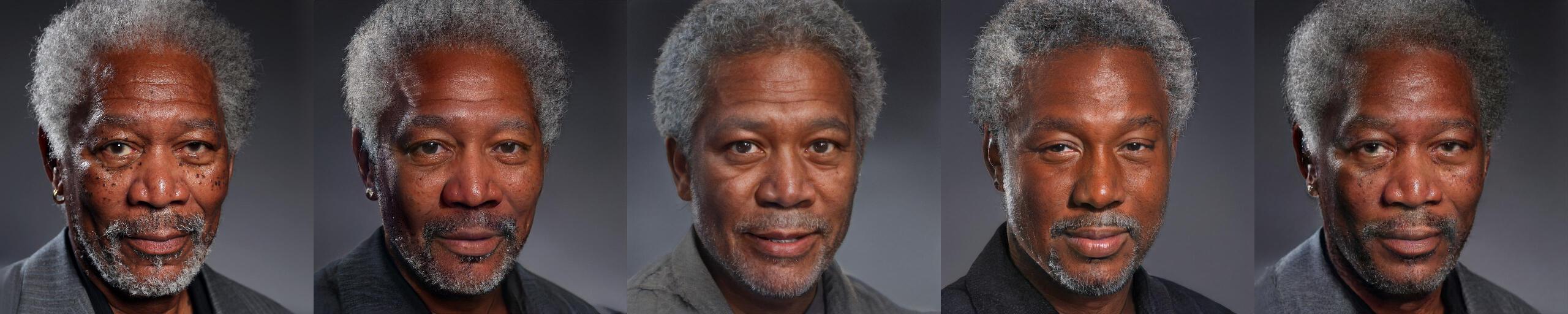}} \\
\noalign{\vskip 0.1cm}
\rotatebox[origin=t]{90}{Eyes Closed} &
\raisebox{-.5\totalheight}{\includegraphics[width=0.9\textwidth]{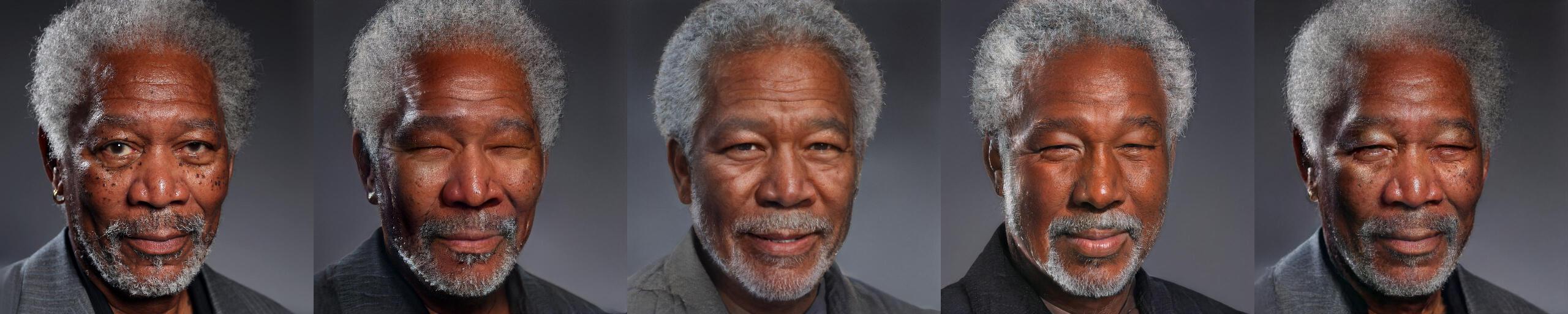}}\\
\noalign{\vskip 0.1cm}
\rotatebox[origin=t]{90}{No beard} &
\raisebox{-.5\totalheight}{\includegraphics[width=0.9\textwidth]{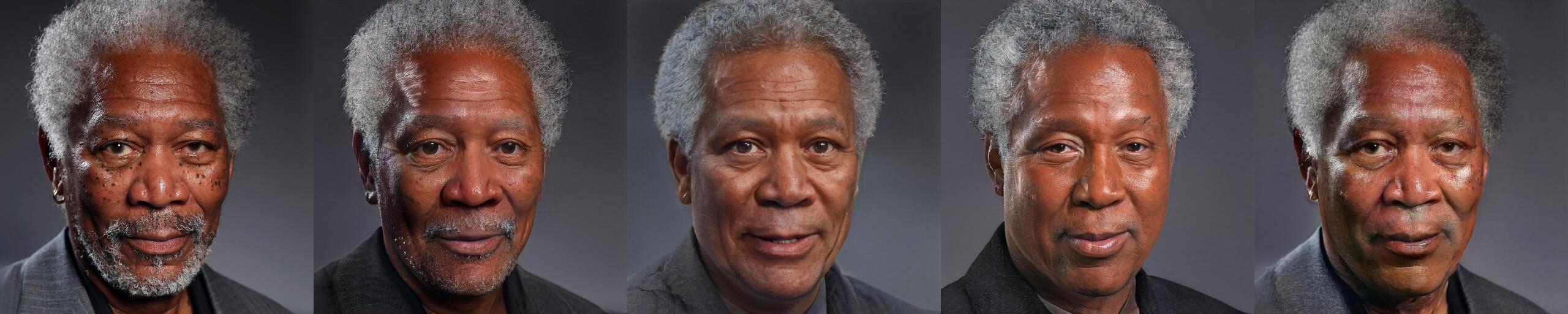}} \\
\noalign{\vskip 0.1cm}
\rotatebox[origin=t]{90}{Rotation} &
\raisebox{-.5\totalheight}{\includegraphics[width=0.9\textwidth]{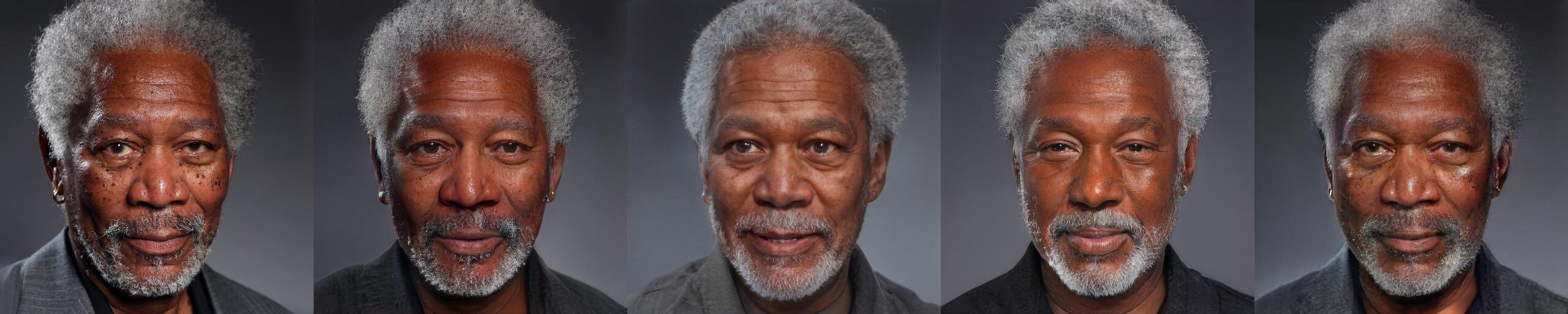}} \\
\end{tabular}
\vspace{0.15cm}
\caption{Comparison of various edits applied on the same challenging image. As can be seen, our method performs meaningful editing while surpassing other methods in preserving fine details. }
\label{fig:ood_edit5}
\end{figure*}

\begin{figure*}
\centering
\begin{tabular}{cc}
\hspace{0.3cm}  \textbf{Original} \hspace{1.3cm} \textbf{StyleClip}  \hspace{0.9cm} \textbf{PTI+StyleClip} &
\hspace{0.3cm}  \textbf{Original} \hspace{1.3cm} \textbf{StyleClip}  \hspace{0.9cm} \textbf{PTI+StyleClip} \\
\raisebox{-.5\totalheight} {\includegraphics[width=\columnwidth]{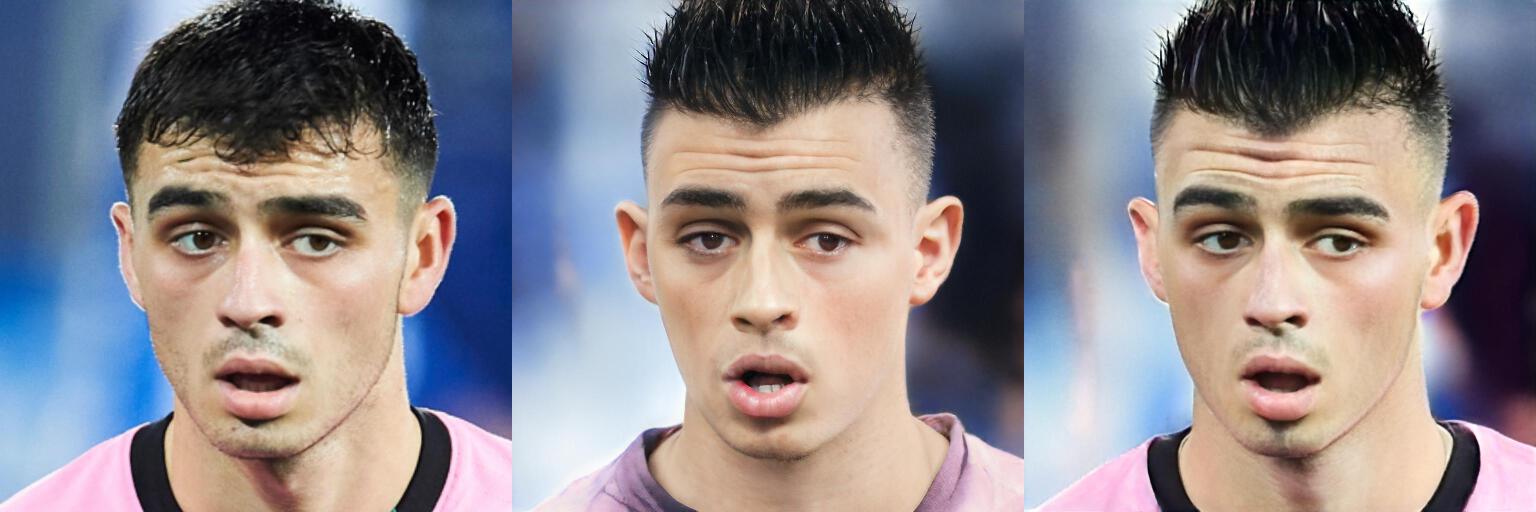}} &
\raisebox{-.5\totalheight} {\includegraphics[width=\columnwidth]{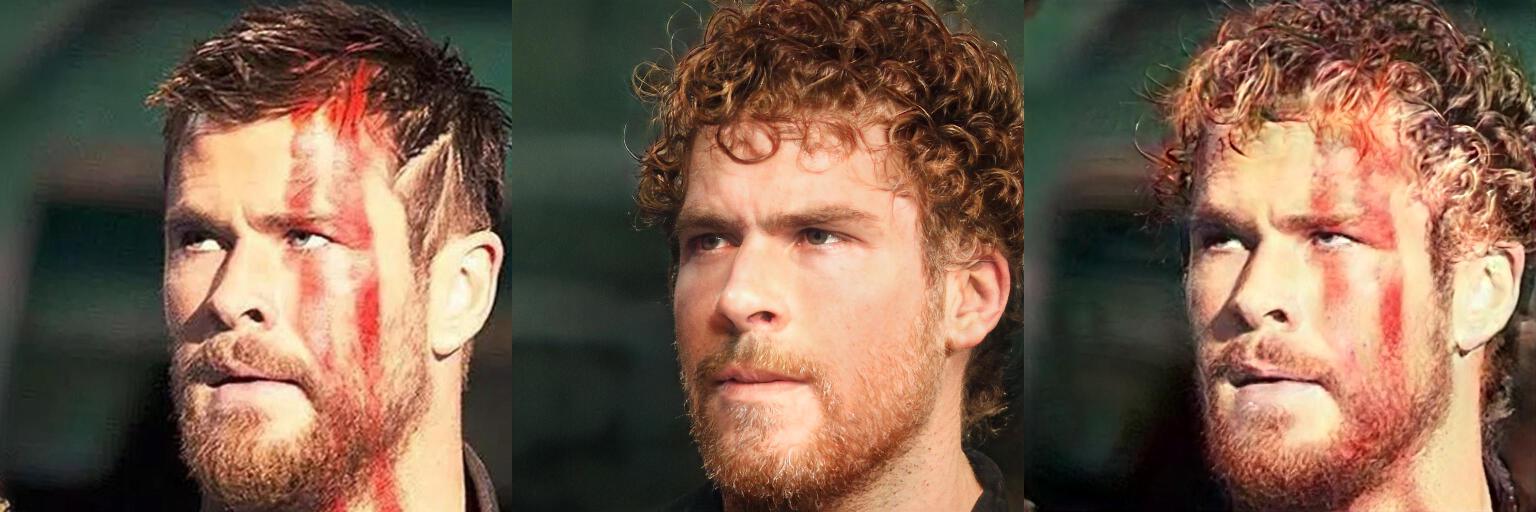}} \\
\raisebox{-.5\totalheight} {\includegraphics[width=\columnwidth]{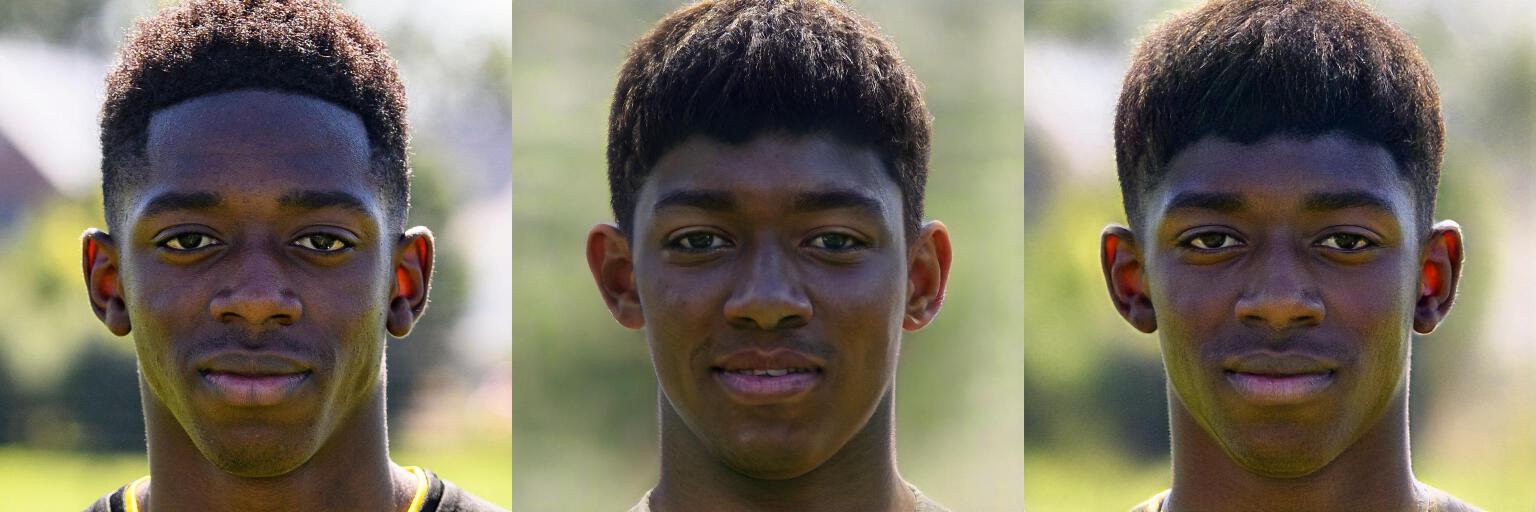}} &
\raisebox{-.5\totalheight} {\includegraphics[width=\columnwidth]{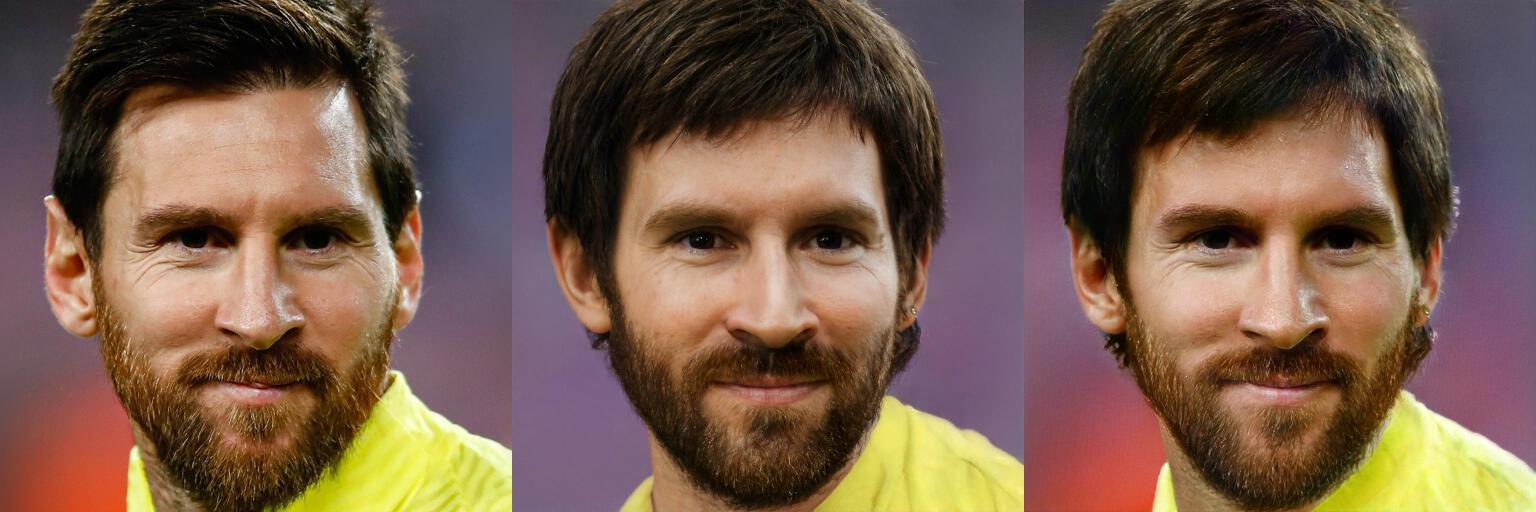}} \\
\raisebox{-.5\totalheight} {\includegraphics[width=\columnwidth]{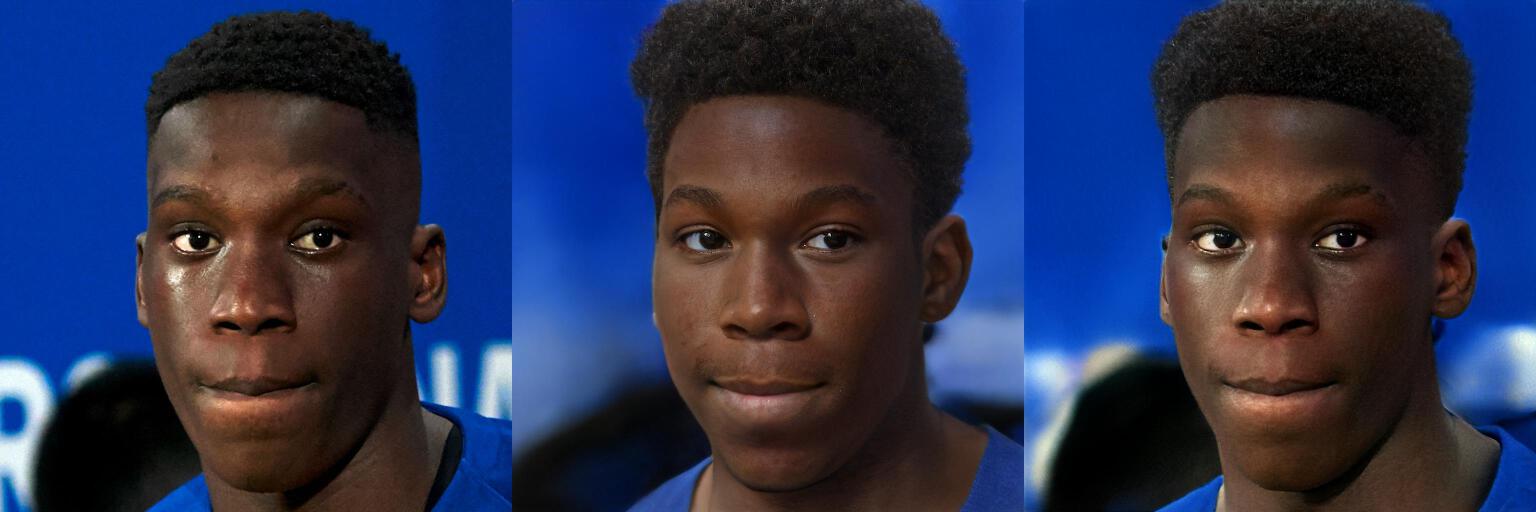}} &
\raisebox{-.5\totalheight} {\includegraphics[width=\columnwidth]{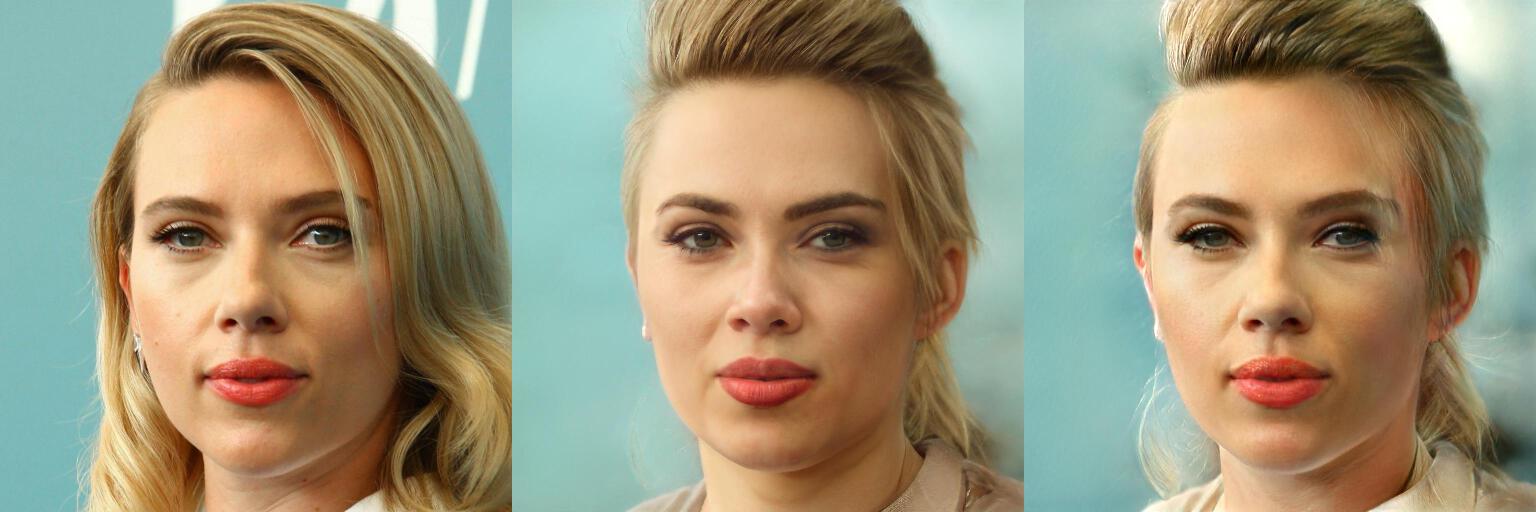}} \\

\end{tabular}
\vspace{0.15cm}
\caption{Additional StyleClip comparison results. We perform various hair edits using StyleClip with and without pivotal tuning inversion.}
\label{fig:styleClip_inv2}
\end{figure*}

Figures~\ref{fig:barcelona_multi_id} and ~\ref{fig:modern_family} demonstrate the inversion of multiple identities. To prevent the suspicion of cherry-picking, we provide uncurated editing comparison results of the first $18$ images from CelebA-HQ test set, provided in Figures~\ref{fig:celeba_supp} to \ref{fig:celeba_supp3}. To further avoid picking, we perform the same three edits recurrently. Figures~\ref{fig:ood_edit2} to ~\ref{fig:ood_edit5} present further comparisons of editing quality over real images of recognizable characters, and Figure~\ref{fig:ood_rec2} depicts reconstruction results. Finally, additional StyleClip editing results can be found Figure~\ref{fig:styleClip_inv2}. All additional results show that our method achieves higher reconstruction and editing quality, even for challenging images. This enables us to preserve the original identity successfully, while still maintaining high editability.

\newcommand{\modenSize}{0.18}

\begin{figure*}[h]

\centering
\begin{tabular}{ccccc}
\rotatebox[origin=t]{90}{Real Image} & \raisebox{-.5\totalheight}{\includegraphics[width=\modenSize\textwidth]{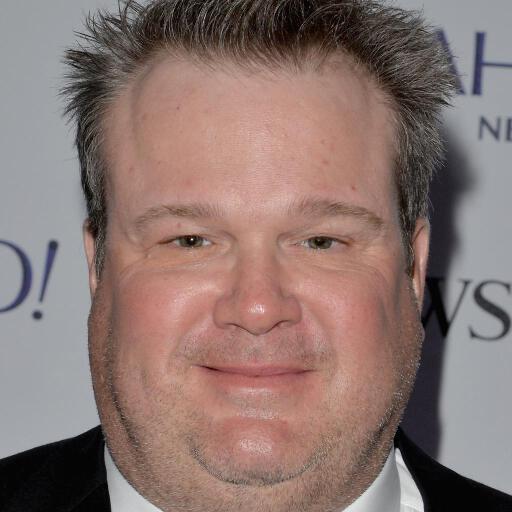}} &
\raisebox{-.5\totalheight}{\includegraphics[width=\modenSize\textwidth]{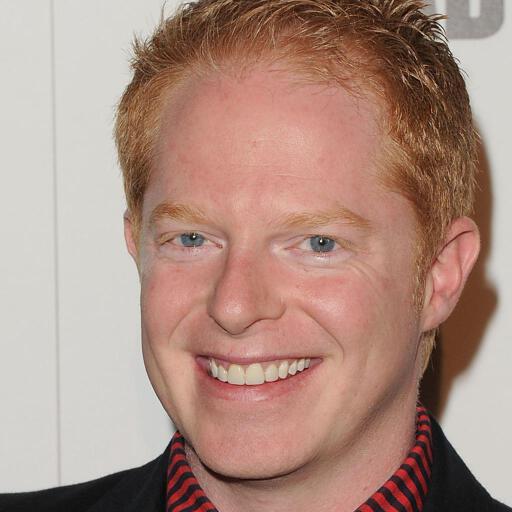}} &
\raisebox{-.5\totalheight}{\includegraphics[width=\modenSize\textwidth]{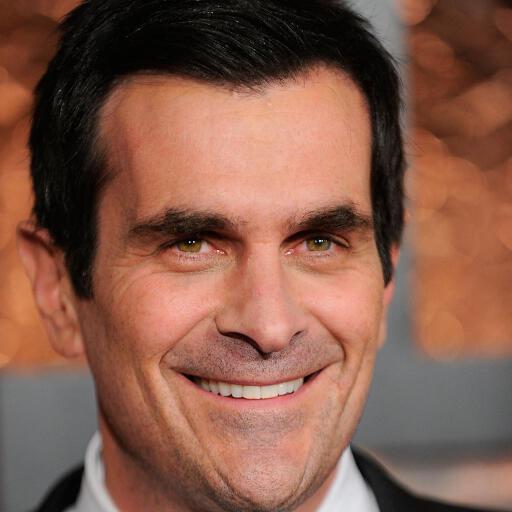}} &
\raisebox{-.5\totalheight}{\includegraphics[width=\modenSize\textwidth]{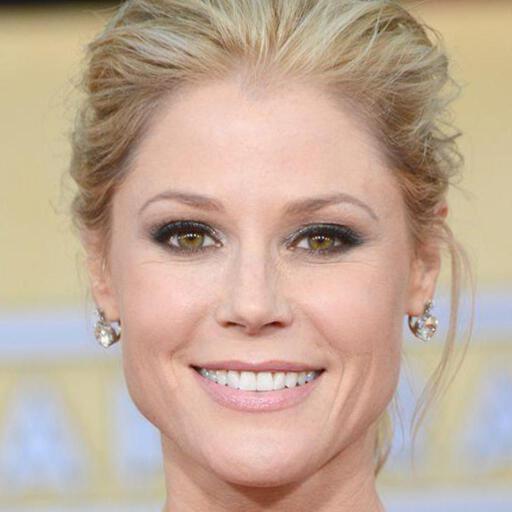}} \\
\noalign{\vskip 1mm}
\rotatebox[origin=t]{90}{Inversion} &
\raisebox{-.5\totalheight}{\includegraphics[width=\modenSize\textwidth]{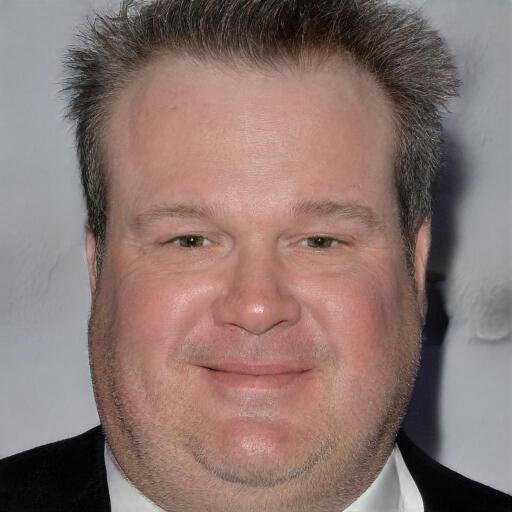}} &
\raisebox{-.5\totalheight}{\includegraphics[width=\modenSize\textwidth]{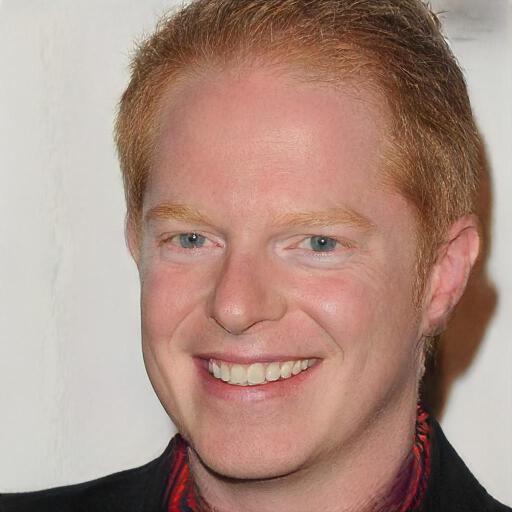}} &
\raisebox{-.5\totalheight}{\includegraphics[width=\modenSize\textwidth]{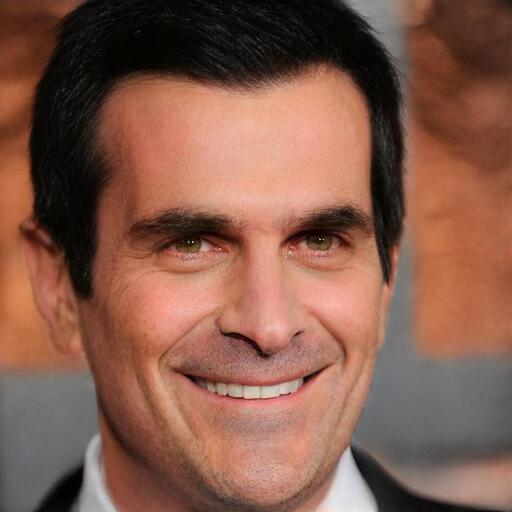}} &
\raisebox{-.5\totalheight}{\includegraphics[width=\modenSize\textwidth]{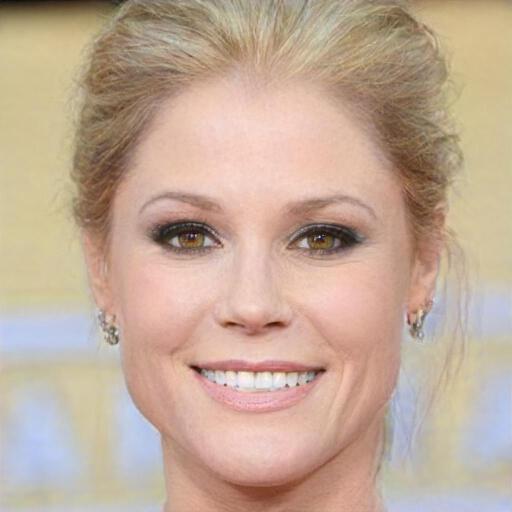}} \\ 
\noalign{\vskip 1mm}
\rotatebox[origin=t]{90}{Age} &
\raisebox{-.5\totalheight}{\includegraphics[width=\modenSize\textwidth]{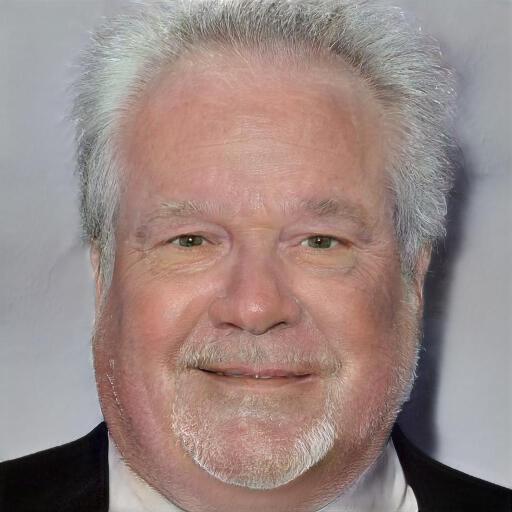}} &
\raisebox{-.5\totalheight}{\includegraphics[width=\modenSize\textwidth]{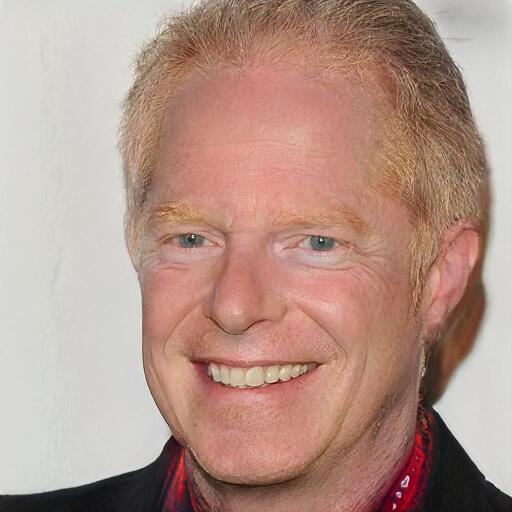}} &
\raisebox{-.5\totalheight}{\includegraphics[width=\modenSize\textwidth]{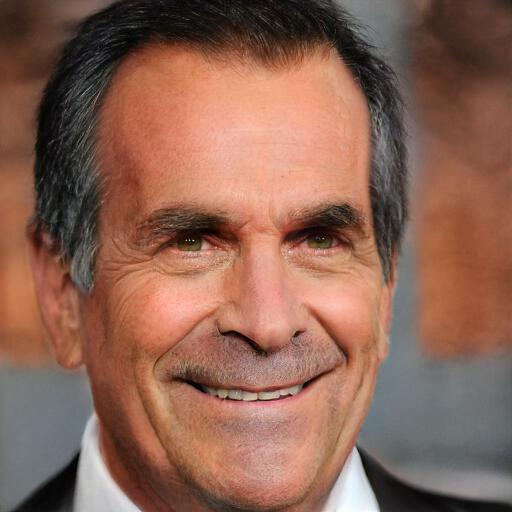}} &
\raisebox{-.5\totalheight}{\includegraphics[width=\modenSize\textwidth]{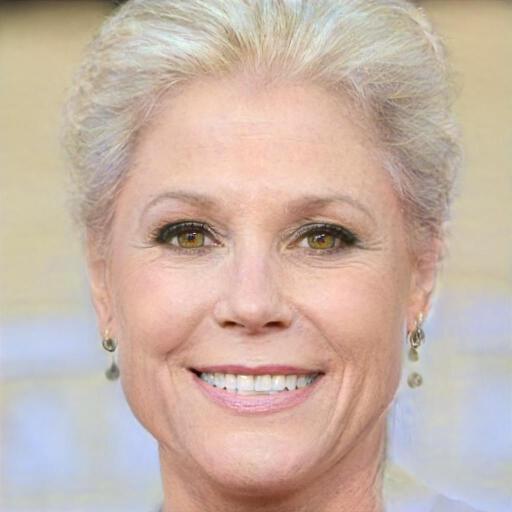}} \\
\noalign{\vskip 1mm}
\rotatebox[origin=t]{90}{Smile} &
\raisebox{-.5\totalheight}{\includegraphics[width=\modenSize\textwidth]{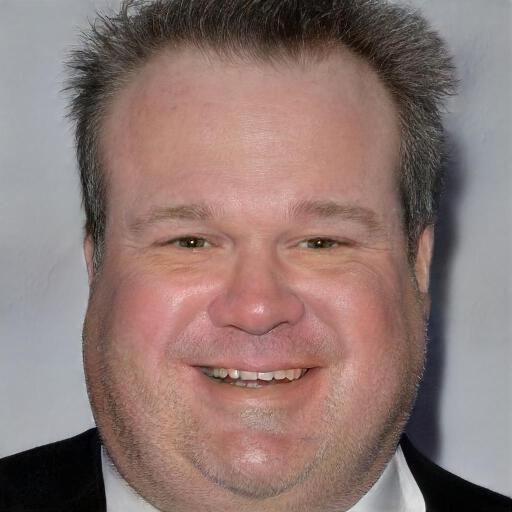}} &
\raisebox{-.5\totalheight}{\includegraphics[width=\modenSize\textwidth]{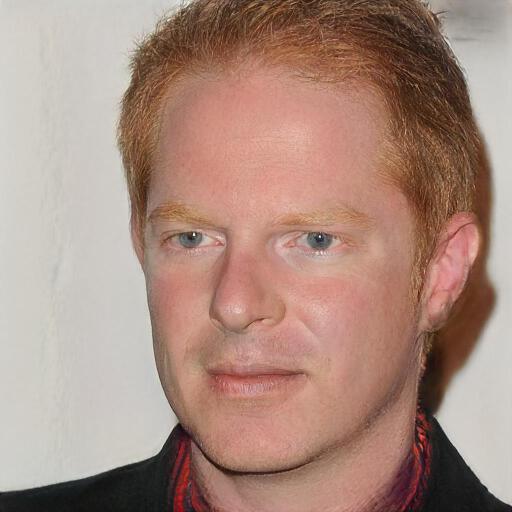}} &
\raisebox{-.5\totalheight}{\includegraphics[width=\modenSize\textwidth]{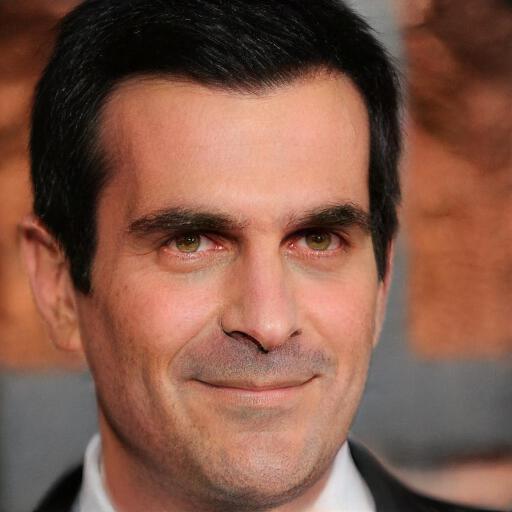}} &
\raisebox{-.5\totalheight}{\includegraphics[width=\modenSize\textwidth]{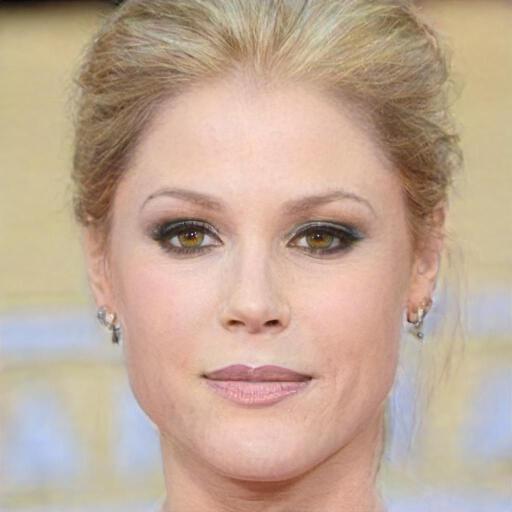}} \\
\noalign{\vskip 0.1cm}
\rotatebox[origin=t]{90}{Rotation} &
\raisebox{-.5\totalheight}{\includegraphics[width=\modenSize\textwidth]{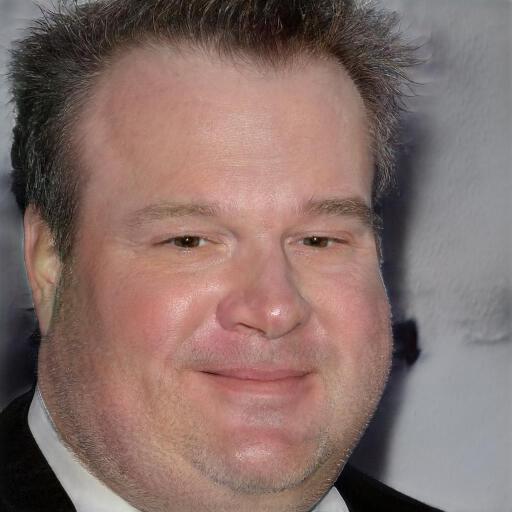}} &
\raisebox{-.5\totalheight}{\includegraphics[width=\modenSize\textwidth]{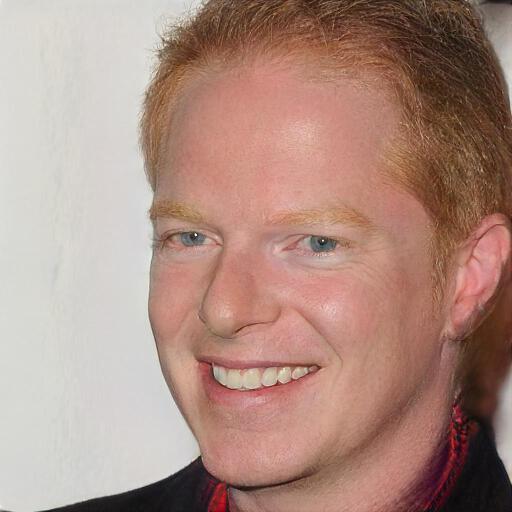}} &
\raisebox{-.5\totalheight}{\includegraphics[width=\modenSize\textwidth]{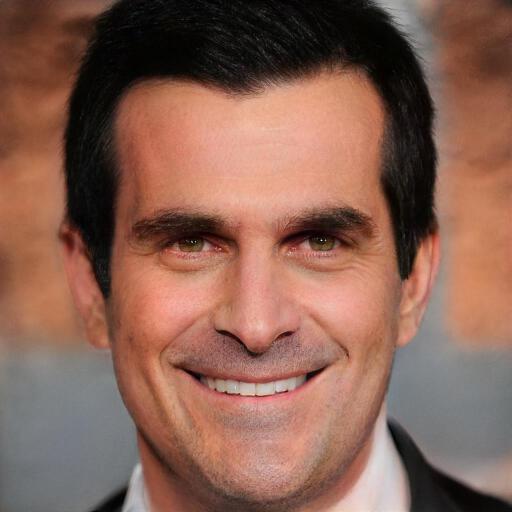}} &
\raisebox{-.5\totalheight}{\includegraphics[width=\modenSize\textwidth]{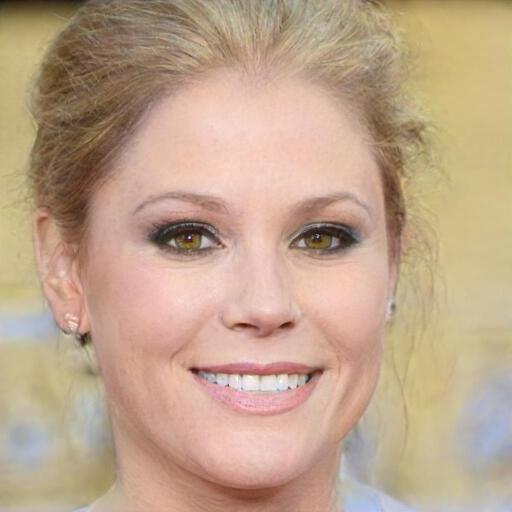}}\\
\noalign{\vskip 0.1cm}
\rotatebox[origin=t]{90}{Hair style} &
\raisebox{-.5\totalheight}{\includegraphics[width=\modenSize\textwidth]{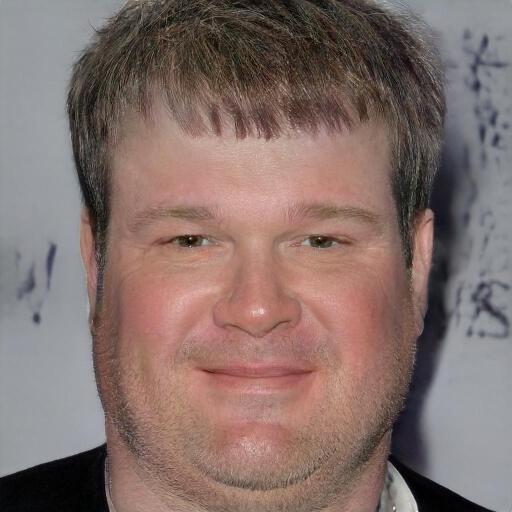}} &
\raisebox{-.5\totalheight}{\includegraphics[width=\modenSize\textwidth]{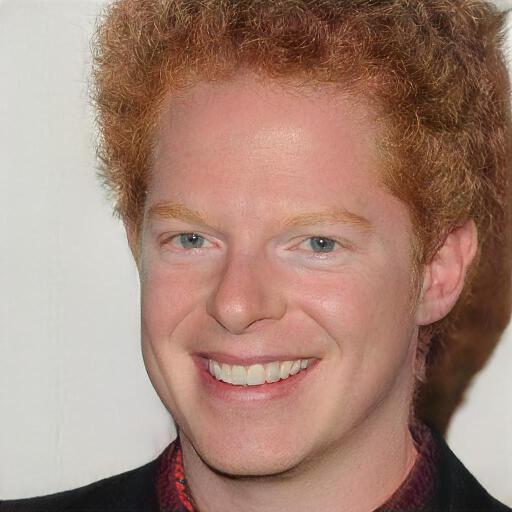}} &
\raisebox{-.5\totalheight}{\includegraphics[width=\modenSize\textwidth]{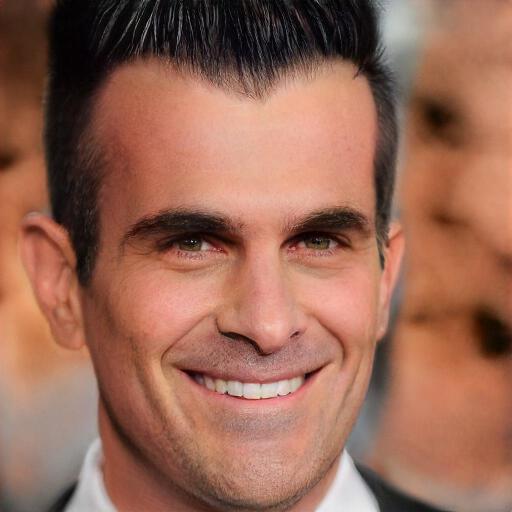}} &
\raisebox{-.5\totalheight}{\includegraphics[width=\modenSize\textwidth]{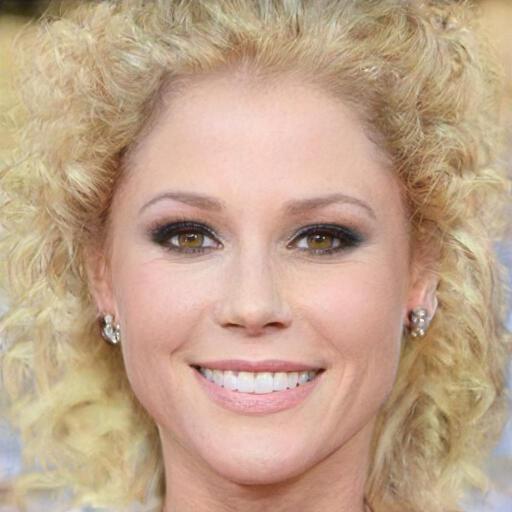}}\\
\end{tabular}
\vspace{0.15cm}
\caption{Modern-Family StyleGAN. We invert multiple images of the television show "Modern Family" cast into a single StyleGAN latent space, and demonstrate both high reconstruction and editing quality for the inverted identities.}
\label{fig:modern_family}
\end{figure*}

\newcommand{\celebaSize}{0.9}

\begin{figure*}[h]

\centering
\begin{tabular}{c}
 \hspace{-0.5cm} \textbf{Original}  \hspace{1.7cm} \textbf{SG2 $\mathcal{W+}$} \hspace{2cm} \textbf{e4e}  \hspace{2.3cm} \textbf{SG2} \hspace{2.2cm} \textbf{Ours} \\
\raisebox{-.5\totalheight}{\includegraphics[width=\celebaSize\textwidth]{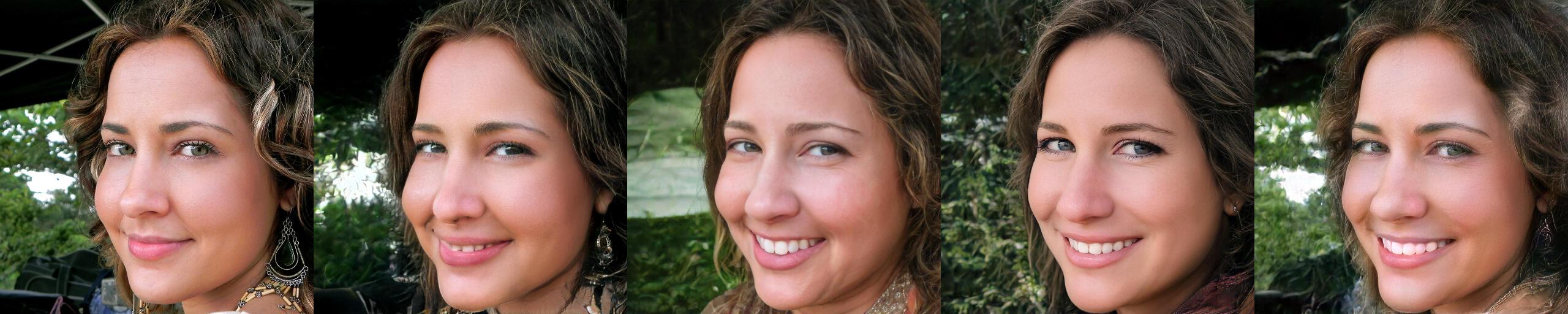}} \\
\noalign{\vskip 1mm}
\raisebox{-.5\totalheight}{\includegraphics[width=\celebaSize\textwidth]{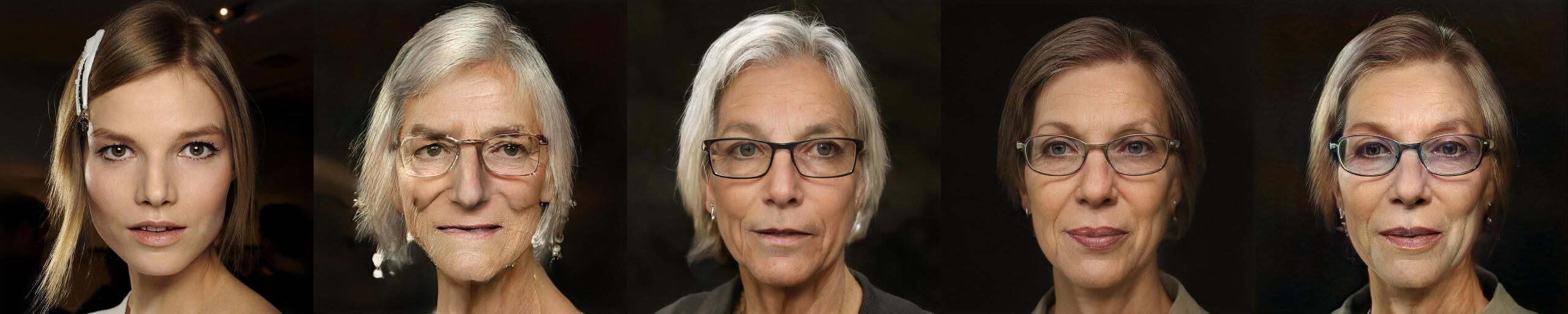}} \\
\noalign{\vskip 1mm}
\raisebox{-.5\totalheight}{\includegraphics[width=\celebaSize\textwidth]{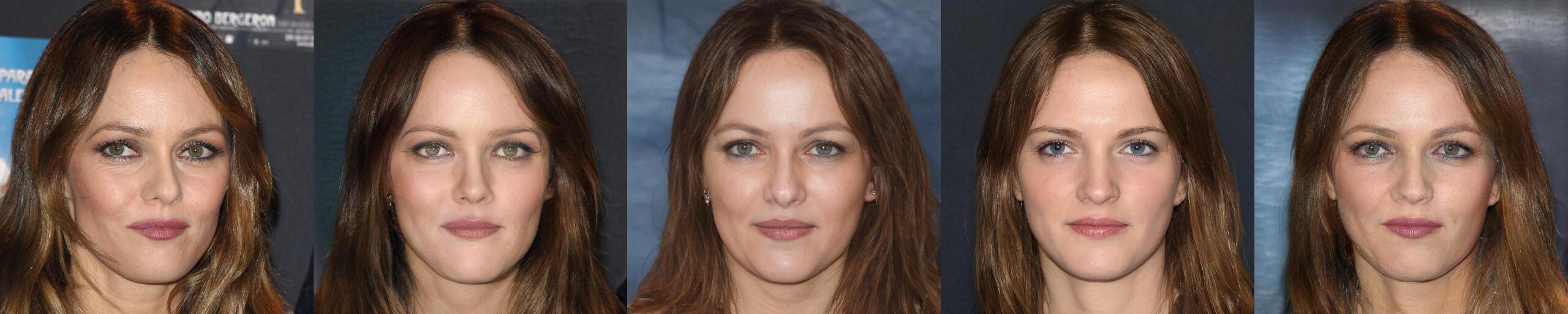}} \\
\noalign{\vskip 1mm}
\raisebox{-.5\totalheight}{\includegraphics[width=\celebaSize\textwidth]{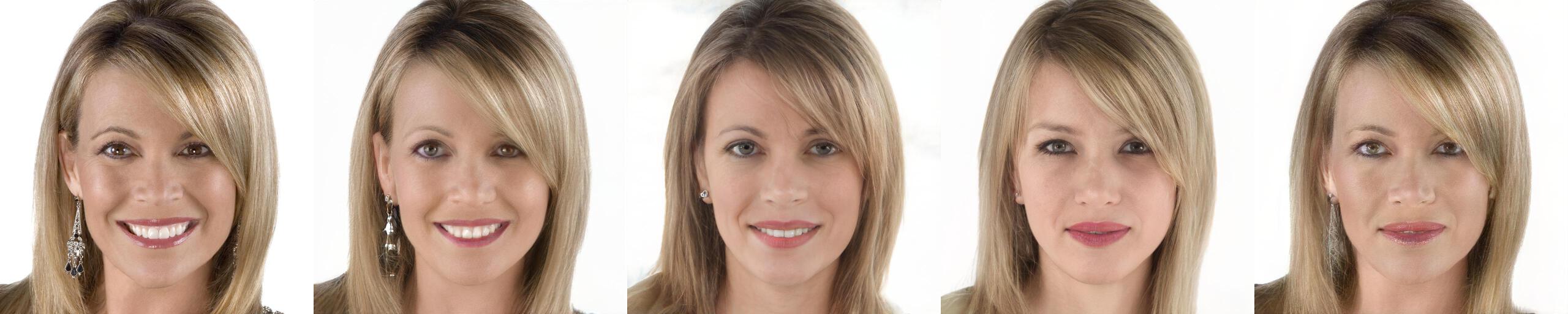}} \\
\noalign{\vskip 1mm}
\raisebox{-.5\totalheight}{\includegraphics[width=\celebaSize\textwidth]{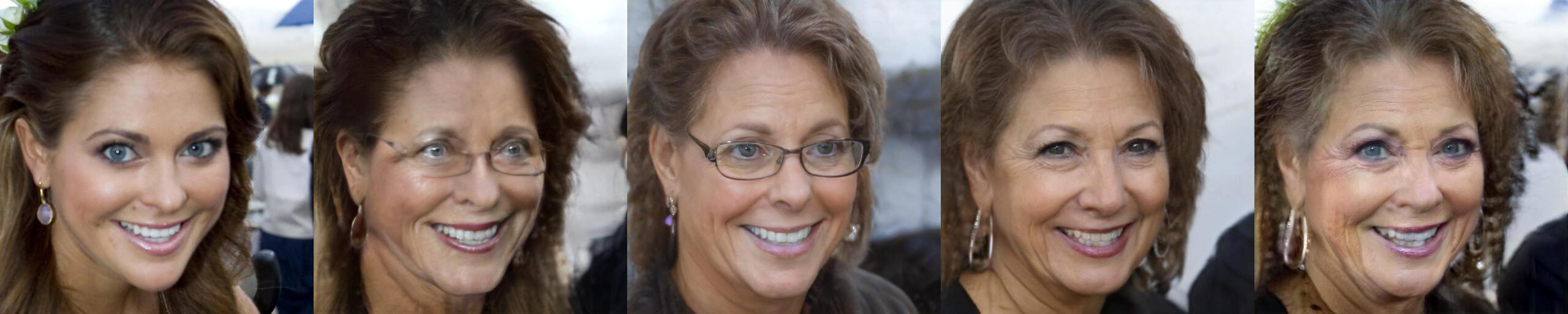}} \\
\noalign{\vskip 1mm}
\raisebox{-.5\totalheight}{\includegraphics[width=\celebaSize\textwidth]{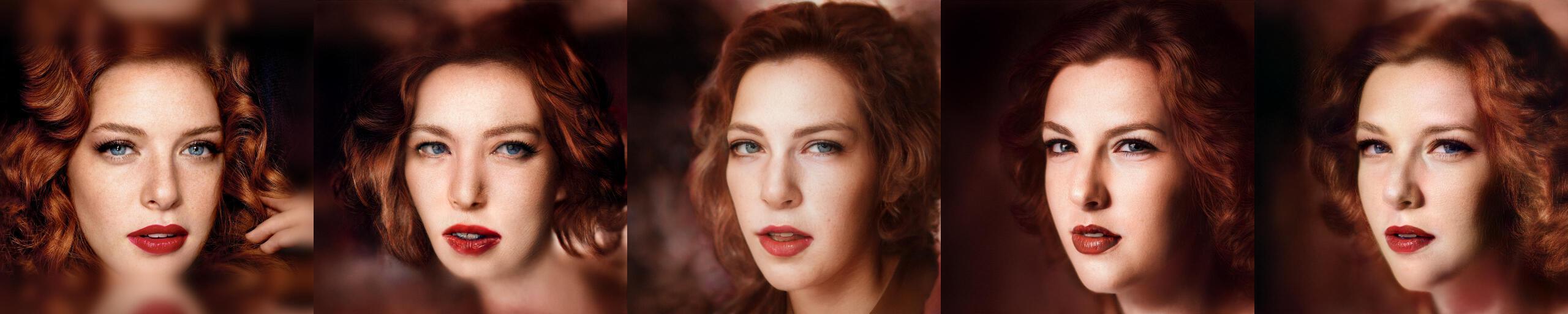}} \\

\end{tabular}
\vspace{0.15cm}
\caption{Uncurated CelebA-HQ editing results. We evaluate the first identities in the test dataset using recurrent edits: smile, age, and rotation. Here identities $0-5$ are depicted.}
\label{fig:celeba_supp}
\end{figure*}

\begin{figure*}[h]

\centering
\begin{tabular}{c}
 \hspace{-0.5cm} \textbf{Original}  \hspace{1.7cm} \textbf{SG2 $\mathcal{W+}$} \hspace{2cm} \textbf{e4e}  \hspace{2.3cm} \textbf{SG2} \hspace{2.2cm} \textbf{Ours} \\
\raisebox{-.5\totalheight}{\includegraphics[width=\celebaSize\textwidth]{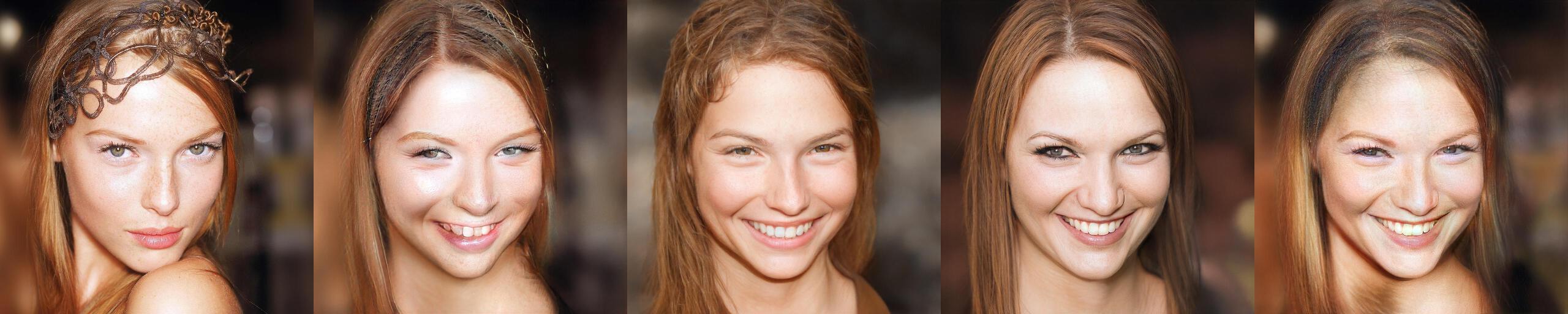}} \\
\noalign{\vskip 1mm}
\raisebox{-.5\totalheight}{\includegraphics[width=\celebaSize\textwidth]{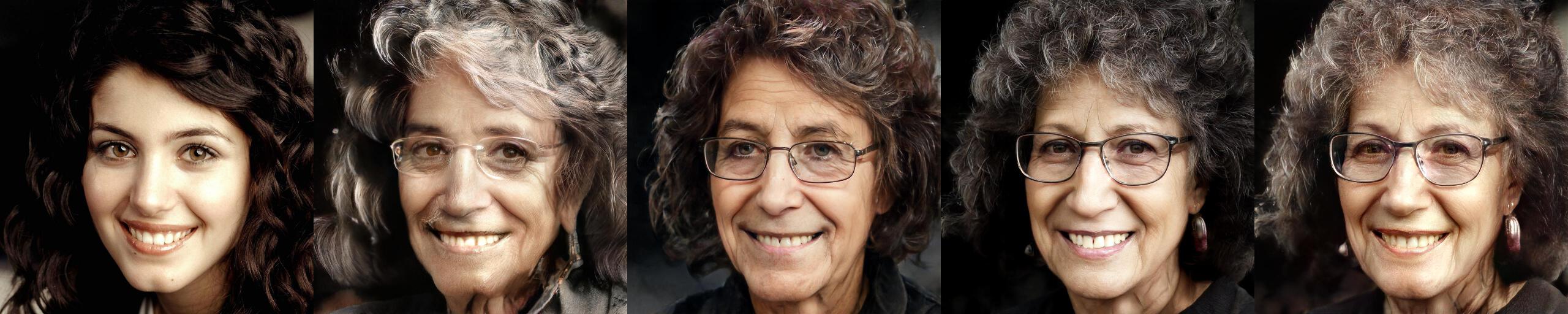}} \\
\noalign{\vskip 1mm}
\raisebox{-.5\totalheight}{\includegraphics[width=\celebaSize\textwidth]{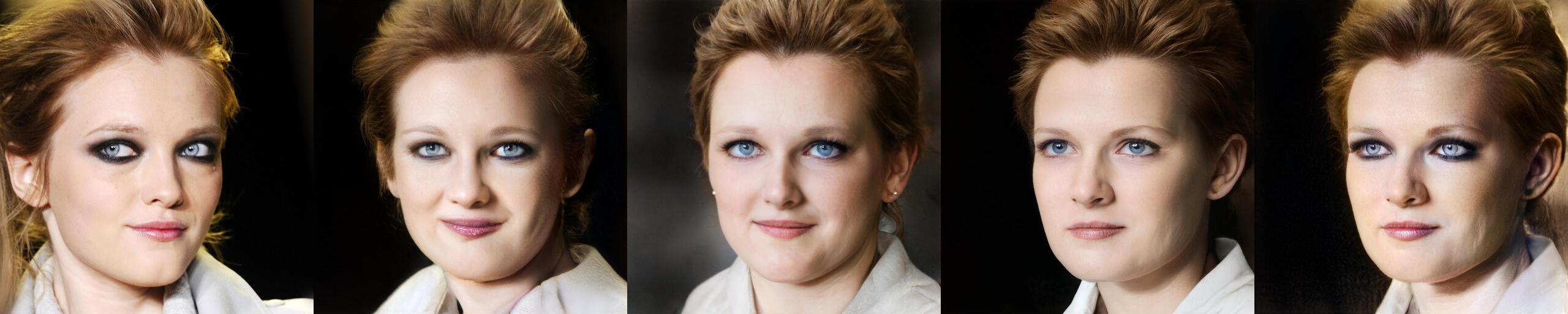}} \\
\noalign{\vskip 1mm}
\raisebox{-.5\totalheight}{\includegraphics[width=\celebaSize\textwidth]{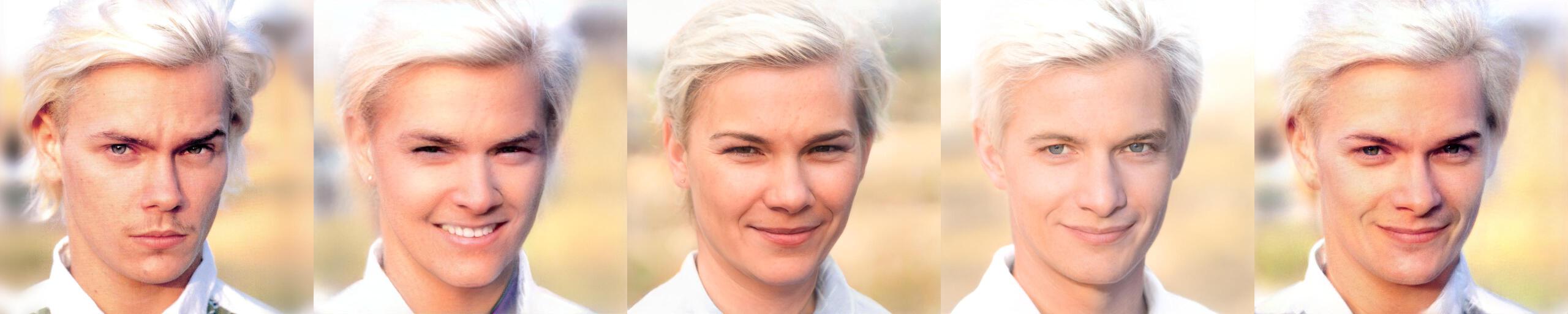}} \\
\noalign{\vskip 1mm}
\raisebox{-.5\totalheight}{\includegraphics[width=\celebaSize\textwidth]{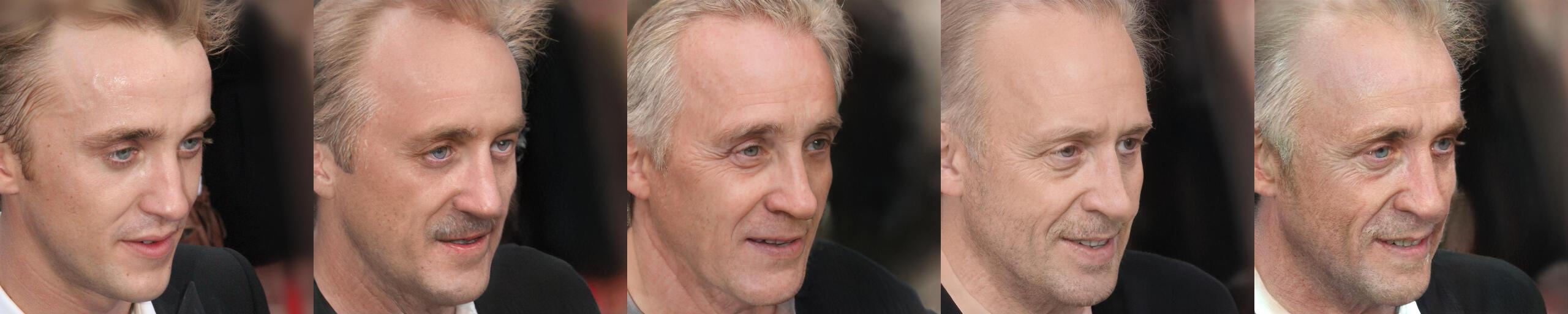}} \\
\noalign{\vskip 1mm}
\raisebox{-.5\totalheight}{\includegraphics[width=\celebaSize\textwidth]{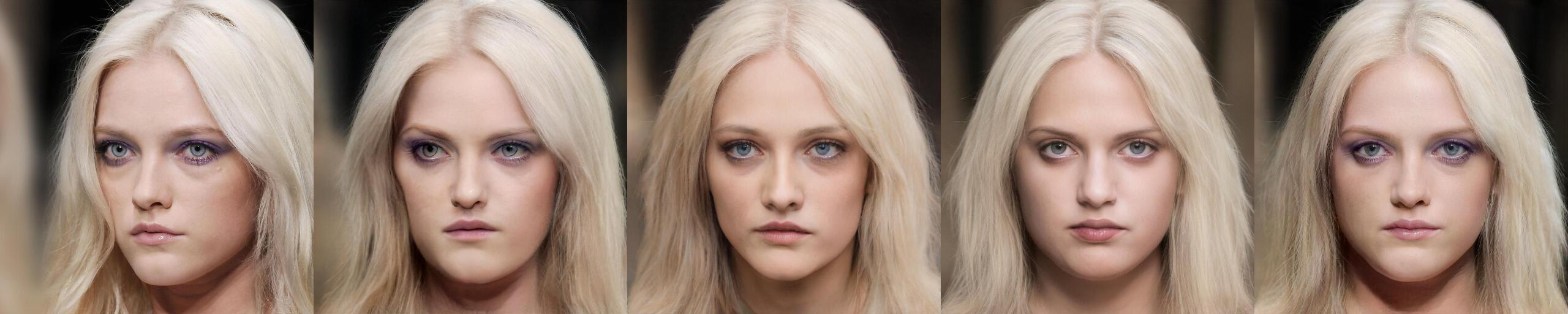}} \\

\end{tabular}
\vspace{0.15cm}
\caption{uncurated CelebA-HQ editing results. We evaluate the first identities in the test dataset using recurrent edits: smile, age, and rotation. Here identities 6-11 are depicted.}
\label{fig:celeba_supp2}
\end{figure*}

\begin{figure*}[h]

\centering
\begin{tabular}{c}
 \hspace{-0.5cm} \textbf{Original}  \hspace{1.7cm} \textbf{SG2 $\mathcal{W+}$} \hspace{2cm} \textbf{e4e}  \hspace{2.3cm} \textbf{SG2} \hspace{2.2cm} \textbf{Ours} \\
\raisebox{-.5\totalheight}{\includegraphics[width=\celebaSize\textwidth]{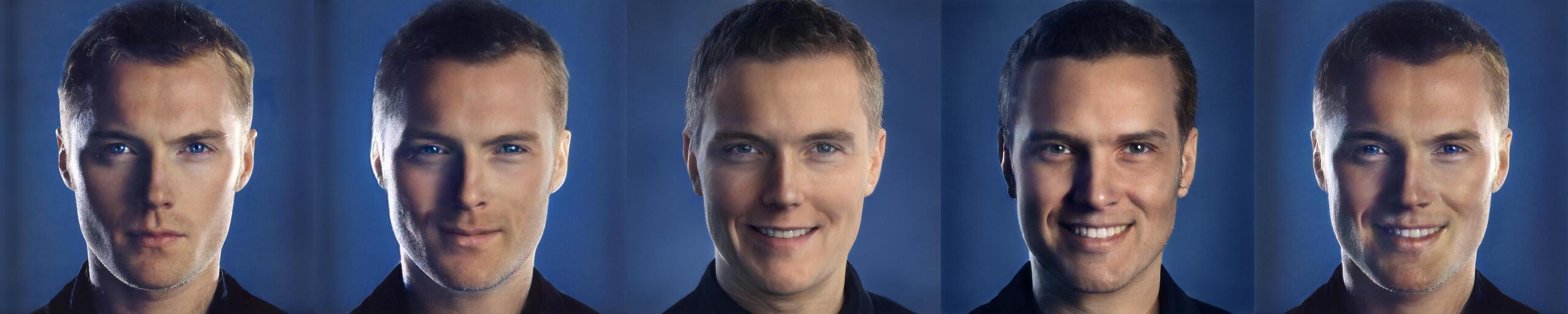}} \\
\noalign{\vskip 1mm}
\raisebox{-.5\totalheight}{\includegraphics[width=\celebaSize\textwidth]{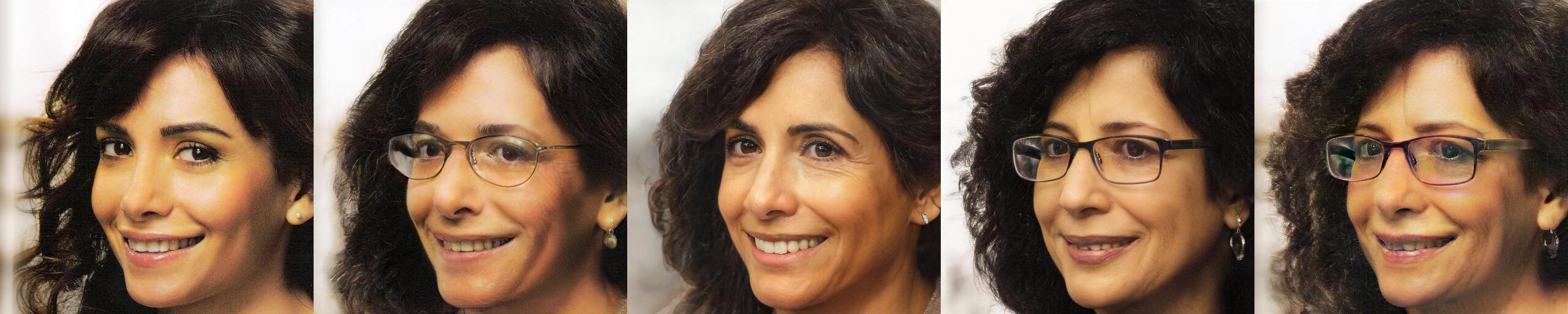}} \\
\noalign{\vskip 1mm}
\raisebox{-.5\totalheight}{\includegraphics[width=\celebaSize\textwidth]{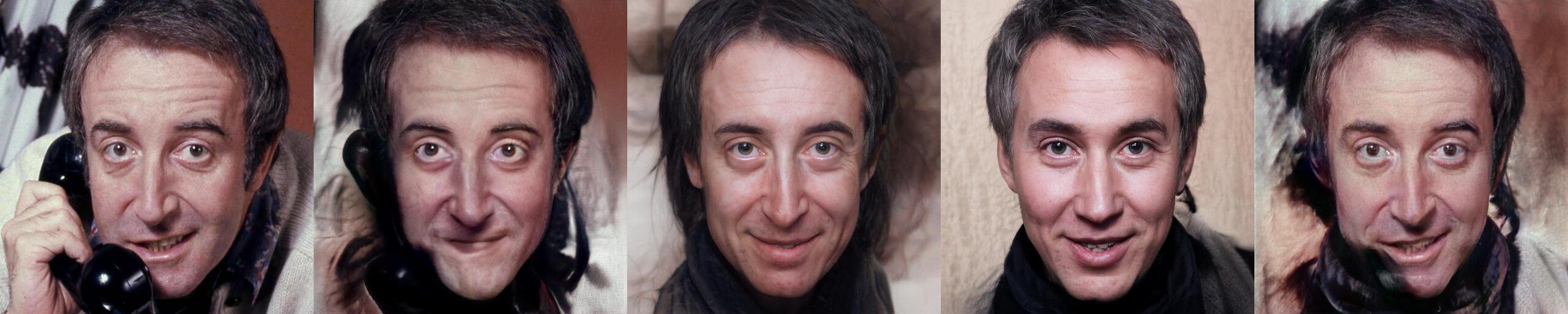}} \\
\noalign{\vskip 1mm}
\raisebox{-.5\totalheight}{\includegraphics[width=\celebaSize\textwidth]{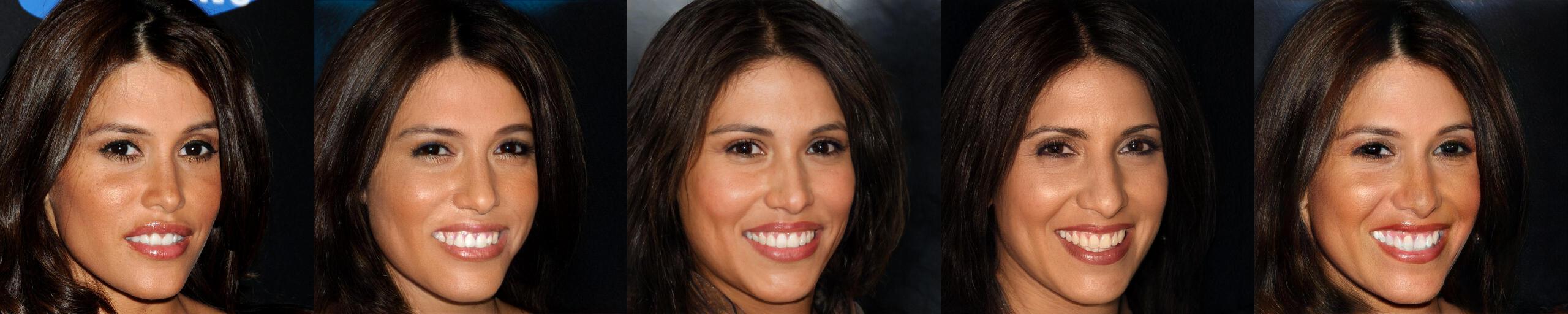}} \\
\noalign{\vskip 1mm}
\raisebox{-.5\totalheight}{\includegraphics[width=\celebaSize\textwidth]{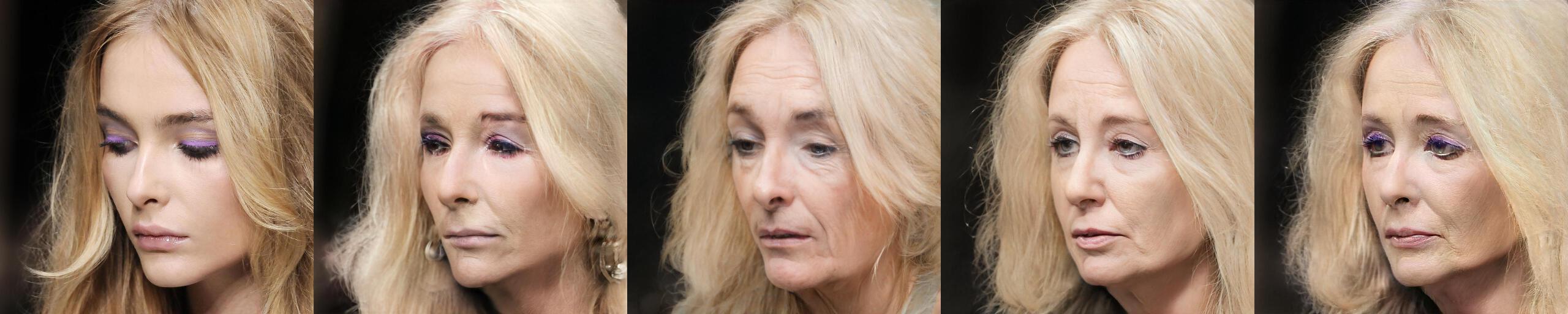}} \\
\noalign{\vskip 1mm}
\raisebox{-.5\totalheight}{\includegraphics[width=\celebaSize\textwidth]{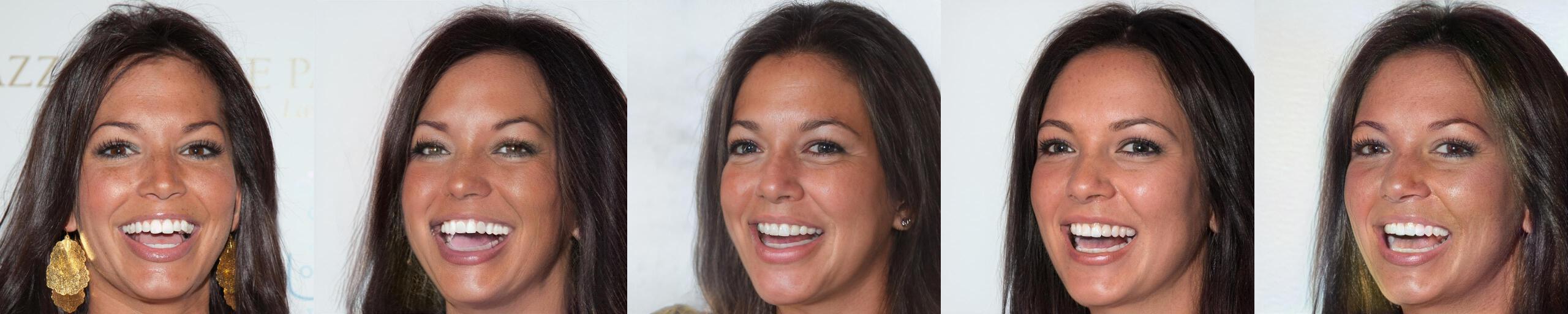}} \\

\end{tabular}
\vspace{0.15cm}
\caption{uncurated CelebA-HQ editing results. We evaluate the first identities in the test dataset using recurrent edits: smile, age, and rotation. Here identities 12-17 are depicted.}
\label{fig:celeba_supp3}
\end{figure*}

\begin{figure*}[h]

\centering
\begin{tabular}{c}
 \hspace{-0.5cm} \textbf{Original}  \hspace{1.7cm} \textbf{SG2 $\mathcal{W+}$} \hspace{2cm} \textbf{e4e}  \hspace{2.3cm} \textbf{SG2} \hspace{2.2cm} \textbf{Ours} \\
\raisebox{-.5\totalheight}{\includegraphics[width=\celebaSize\textwidth]{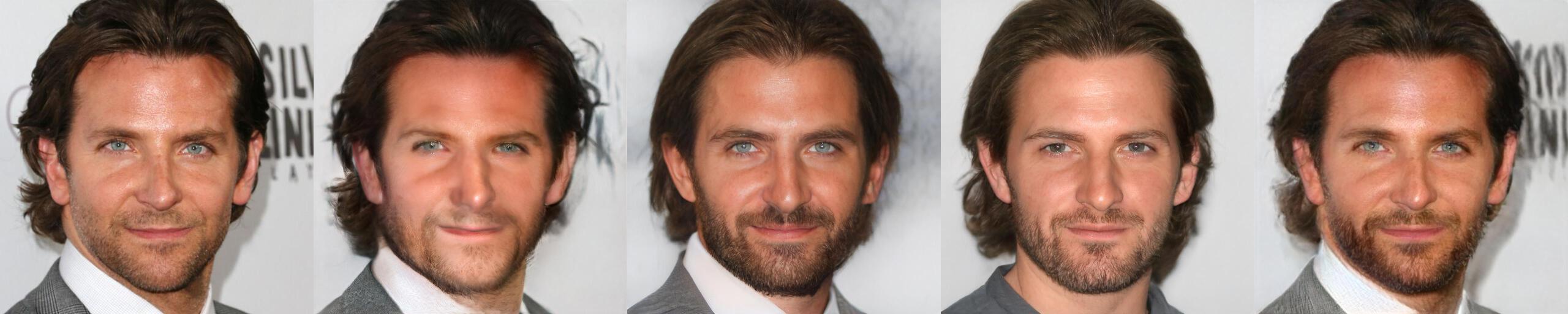}} \\
\noalign{\vskip 1mm}
\raisebox{-.5\totalheight}{\includegraphics[width=\celebaSize\textwidth]{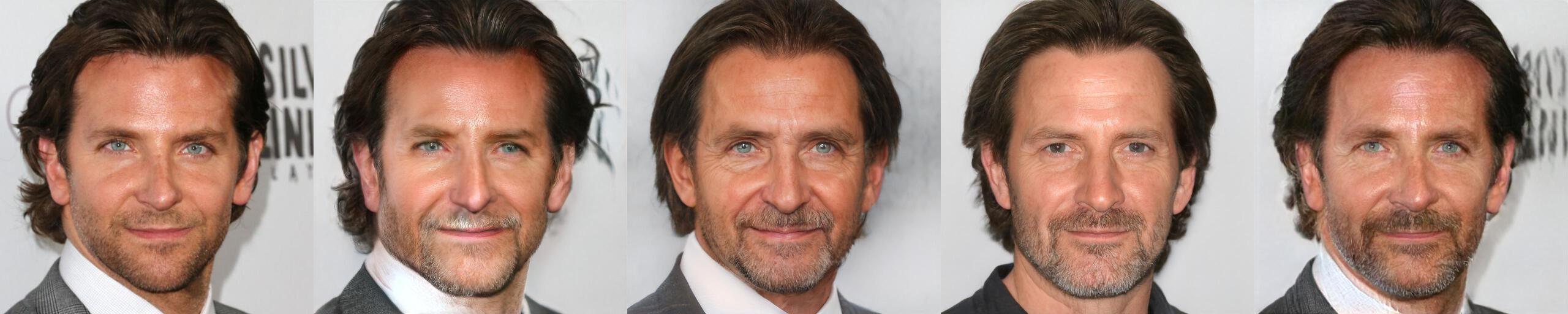}} \\
\noalign{\vskip 1mm}
\raisebox{-.5\totalheight}{\includegraphics[width=\celebaSize\textwidth]{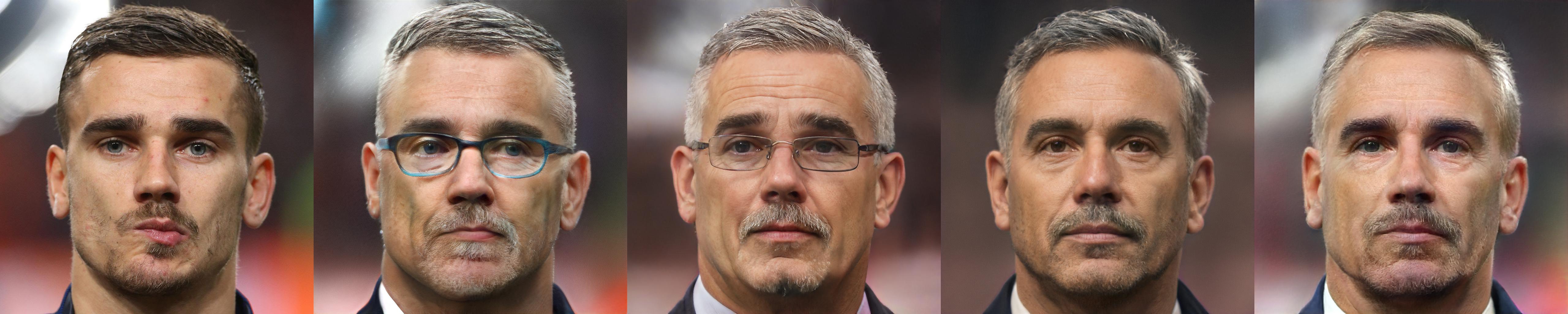}} \\
\noalign{\vskip 1mm}
\raisebox{-.5\totalheight}{\includegraphics[width=\celebaSize\textwidth]{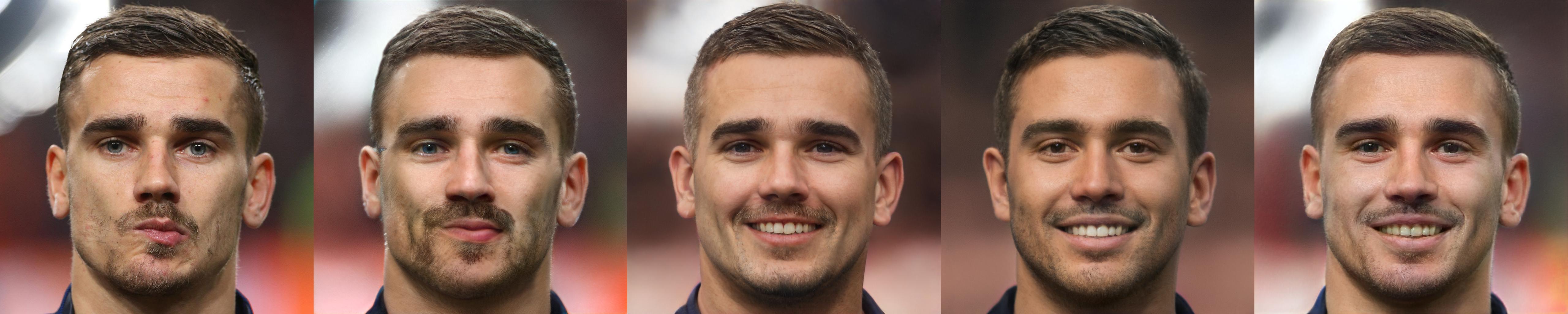}} \\
\noalign{\vskip 1mm}
\raisebox{-.5\totalheight}{\includegraphics[width=\celebaSize\textwidth]{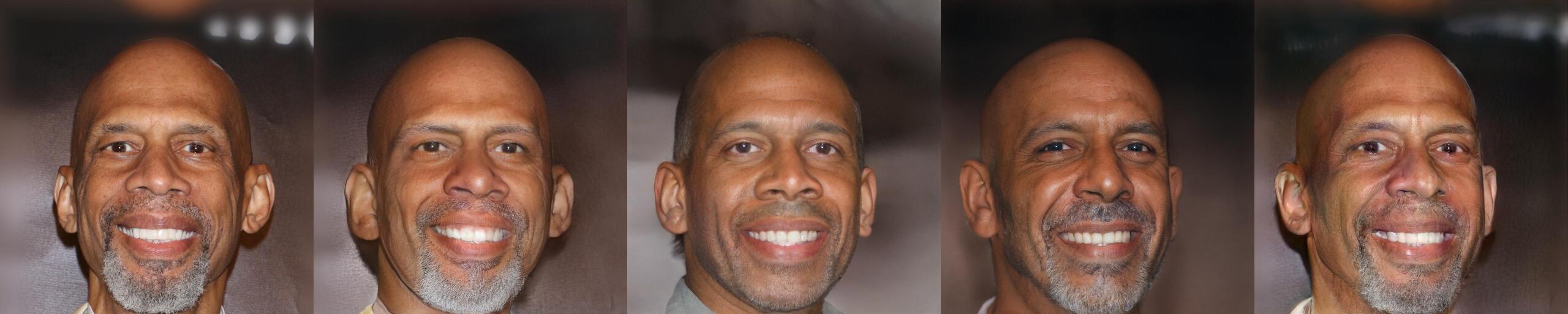}} \\
\noalign{\vskip 1mm}
\raisebox{-.5\totalheight}{\includegraphics[width=\celebaSize\textwidth]{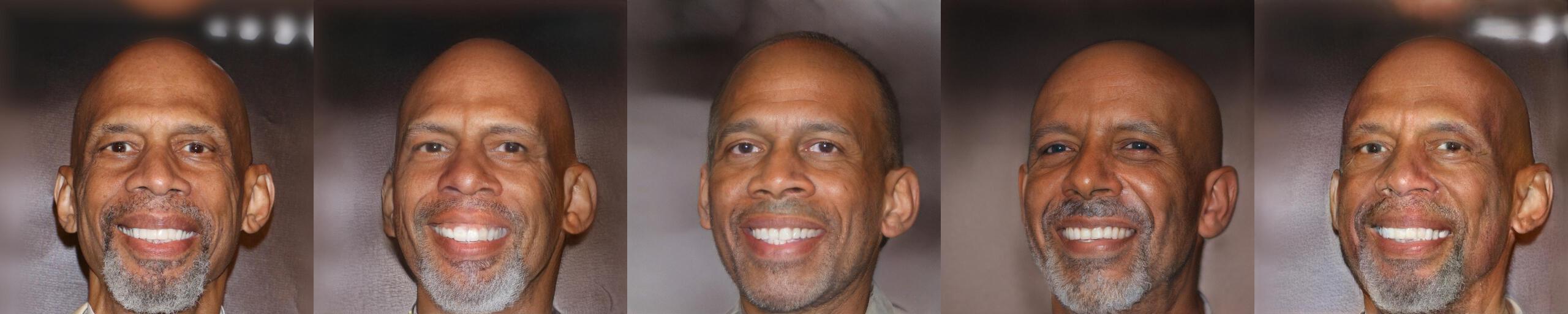}} \\
\end{tabular}
\vspace{0.15cm}
\caption{Additional editing comparison over real images.}
\label{fig:ood_edit2}
\end{figure*}

\begin{figure*}[h]

\centering
\begin{tabular}{c}
 \hspace{-0.5cm} \textbf{Original}  \hspace{1.7cm} \textbf{SG2 $\mathcal{W+}$} \hspace{2cm} \textbf{e4e}  \hspace{2.3cm} \textbf{SG2} \hspace{2.2cm} \textbf{Ours} \\
\raisebox{-.5\totalheight}{\includegraphics[width=\celebaSize\textwidth]{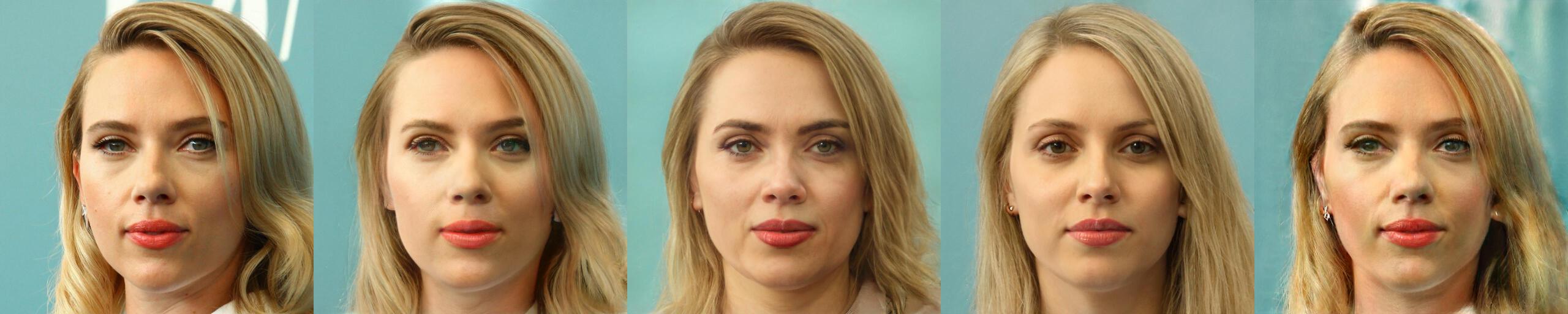}} \\
\noalign{\vskip 1mm}
\raisebox{-.5\totalheight}{\includegraphics[width=\celebaSize\textwidth]{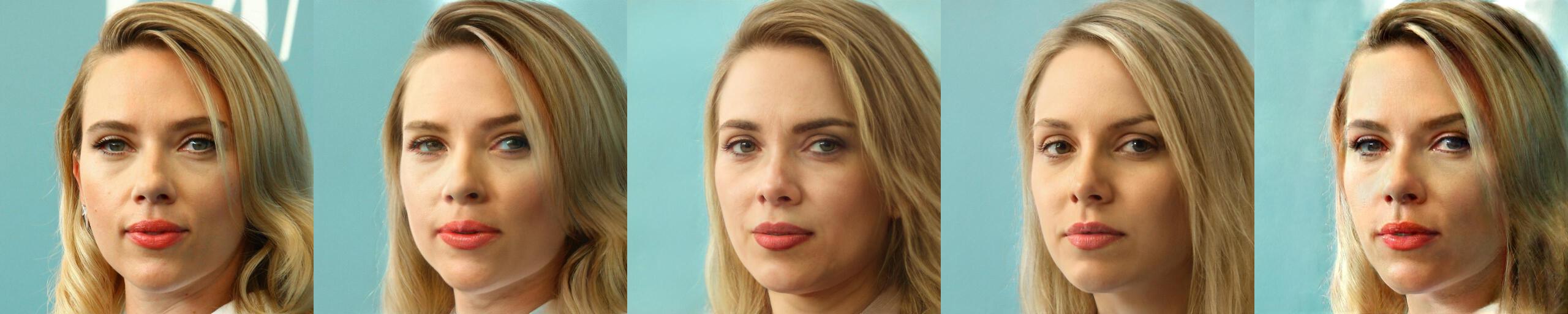}} \\
\noalign{\vskip 1mm}
\raisebox{-.5\totalheight}{\includegraphics[width=\celebaSize\textwidth]{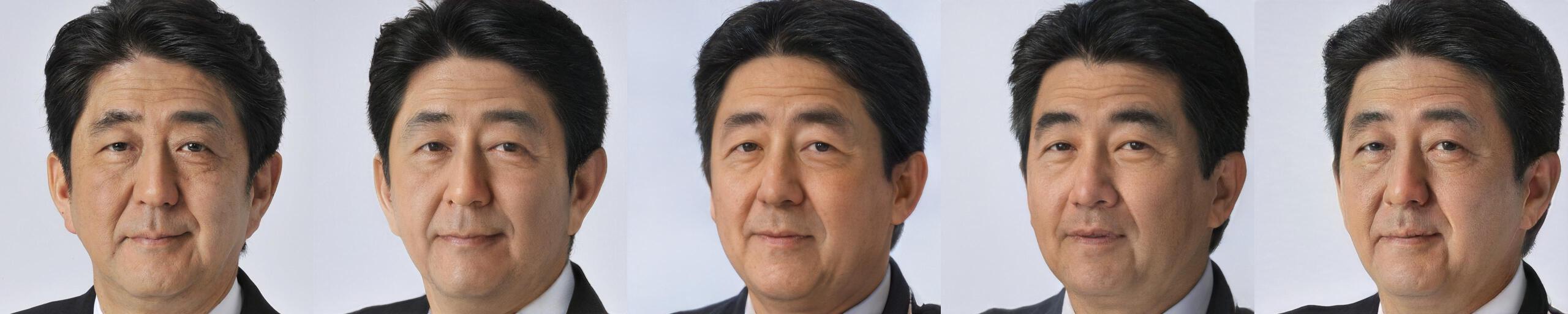}} \\
\noalign{\vskip 1mm}
\raisebox{-.5\totalheight}{\includegraphics[width=\celebaSize\textwidth]{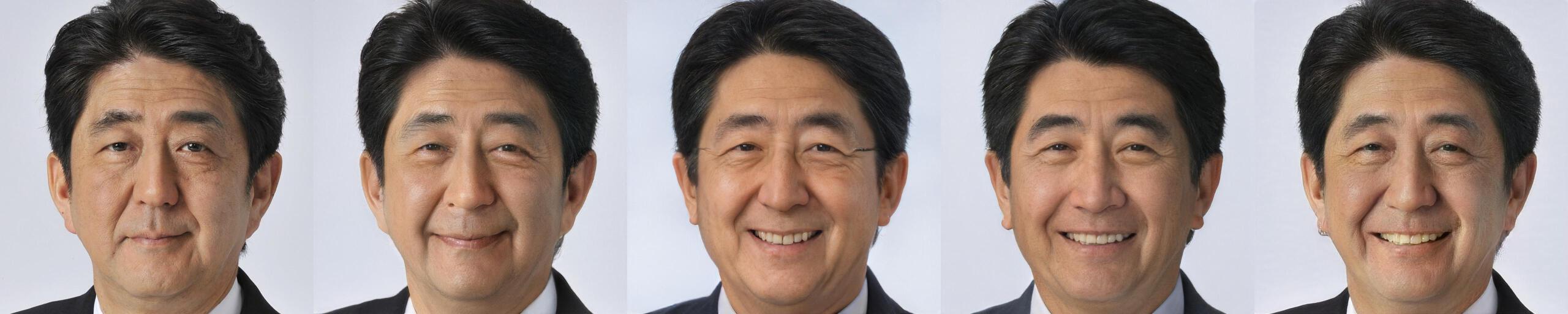}} \\
\noalign{\vskip 1mm}
\raisebox{-.5\totalheight}{\includegraphics[width=\celebaSize\textwidth]{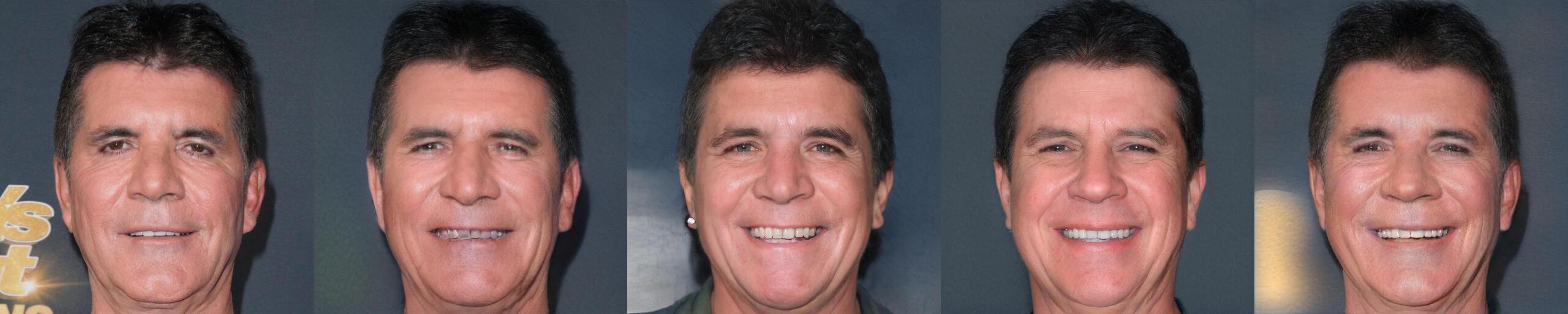}} \\
\noalign{\vskip 1mm}
\raisebox{-.5\totalheight}{\includegraphics[width=\celebaSize\textwidth]{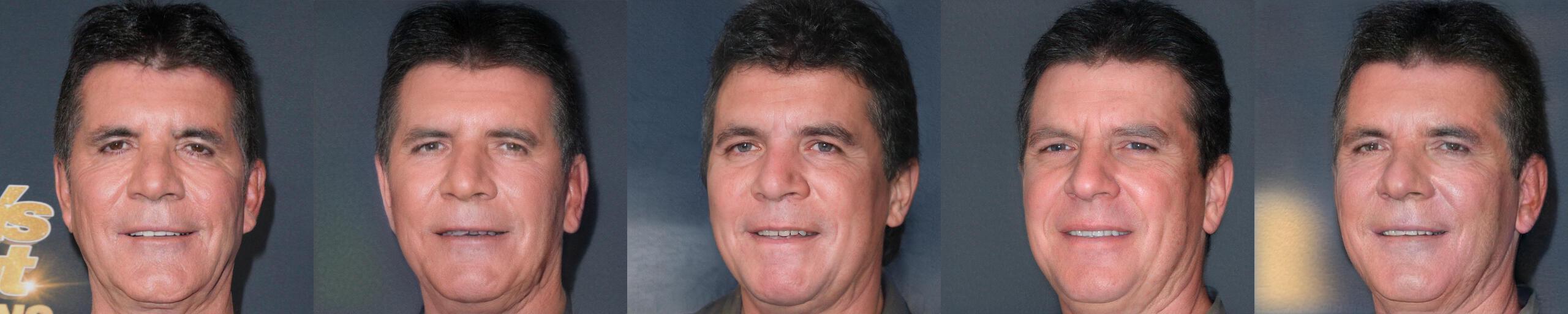}} \\

\end{tabular}
\vspace{0.15cm}
\caption{Additional editing comparison over real images.}
\label{fig:ood_edit3}
\end{figure*}

\begin{figure*}[h]

\centering
\begin{tabular}{c}
 \hspace{-0.5cm} \textbf{Original}  \hspace{1.7cm} \textbf{SG2 $\mathcal{W+}$} \hspace{2cm} \textbf{e4e}  \hspace{2.3cm} \textbf{SG2} \hspace{2.2cm} \textbf{Ours} \\
\raisebox{-.5\totalheight}{\includegraphics[width=\celebaSize\textwidth]{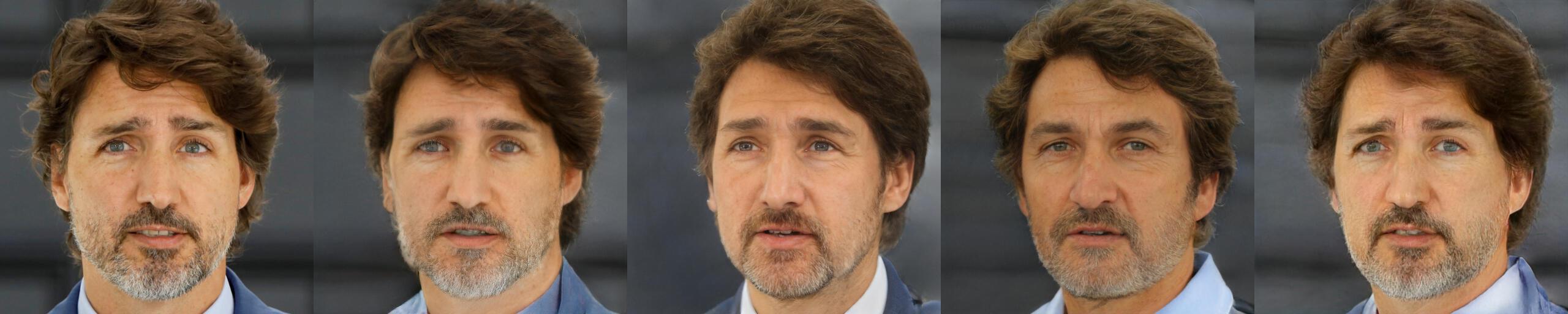}} \\
\noalign{\vskip 1mm}
\raisebox{-.5\totalheight}{\includegraphics[width=\celebaSize\textwidth]{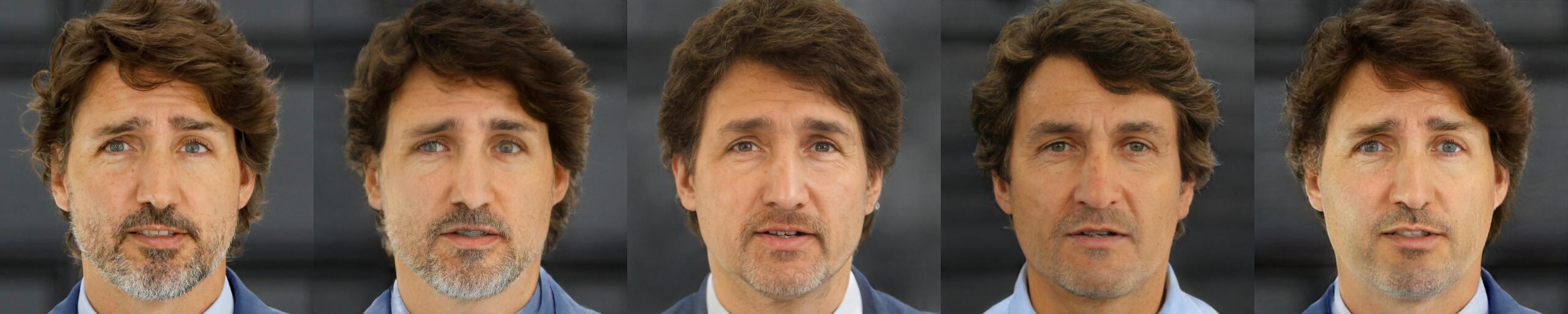}} \\
\noalign{\vskip 1mm}
\raisebox{-.5\totalheight}{\includegraphics[width=\celebaSize\textwidth]{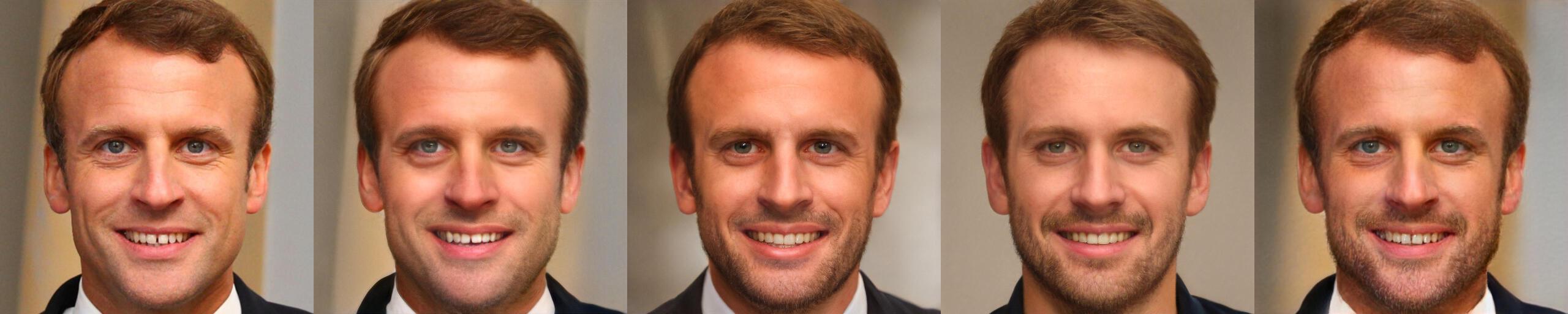}} \\
\noalign{\vskip 1mm}
\raisebox{-.5\totalheight}{\includegraphics[width=\celebaSize\textwidth]{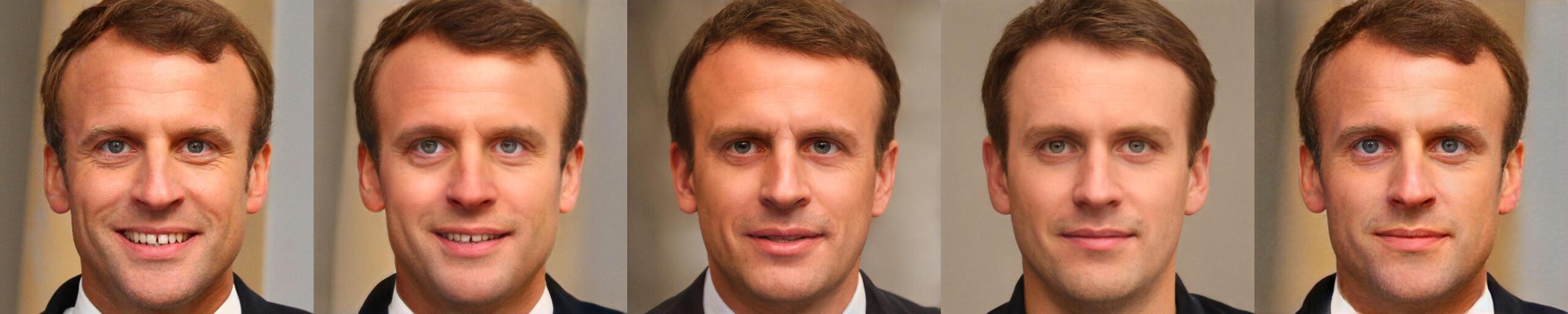}} \\
\noalign{\vskip 1mm}
\raisebox{-.5\totalheight}{\includegraphics[width=\celebaSize\textwidth]{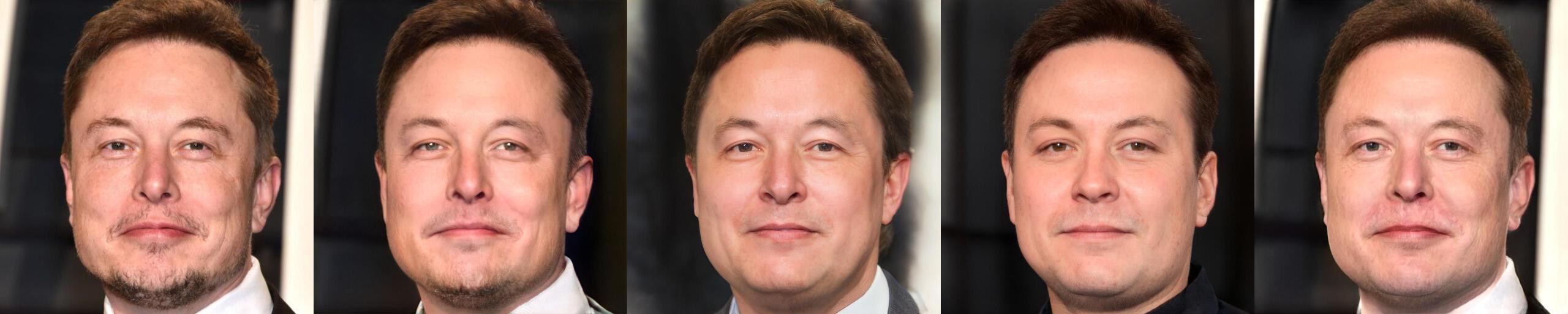}} \\
\noalign{\vskip 1mm}
\raisebox{-.5\totalheight}{\includegraphics[width=\celebaSize\textwidth]{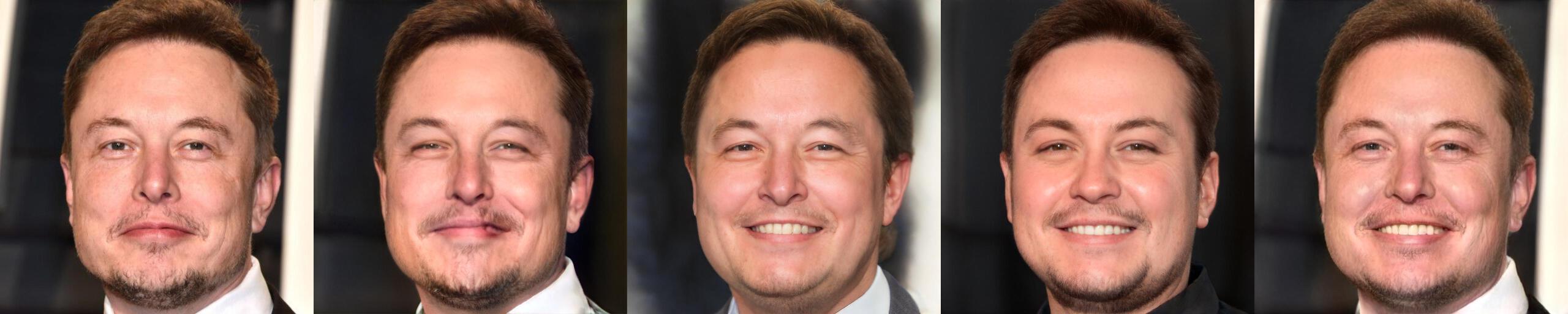}} \\

\end{tabular}
\vspace{0.15cm}
\caption{Additional editing comparison over real images.}
\label{fig:ood_edit4}
\end{figure*}

\begin{figure*}[h]

\centering
\begin{tabular}{c}
 \hspace{-0.5cm} \textbf{Original}  \hspace{1.7cm} \textbf{SG2 $\mathcal{W+}$} \hspace{2cm} \textbf{e4e}  \hspace{2.3cm} \textbf{SG2} \hspace{2.2cm} \textbf{Ours} \\
\raisebox{-.5\totalheight}{\includegraphics[width=\celebaSize\textwidth]{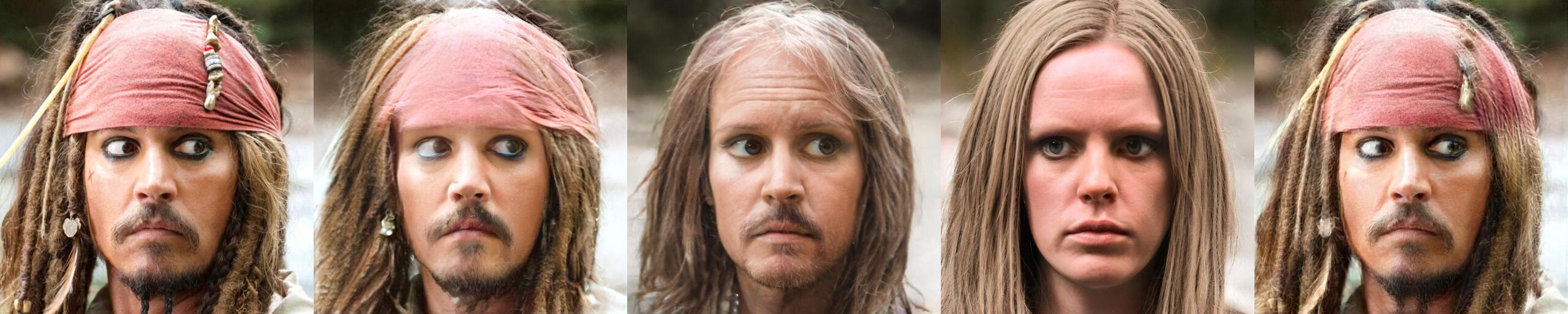}} \\
\noalign{\vskip 1mm}
\raisebox{-.5\totalheight}{\includegraphics[width=\celebaSize\textwidth]{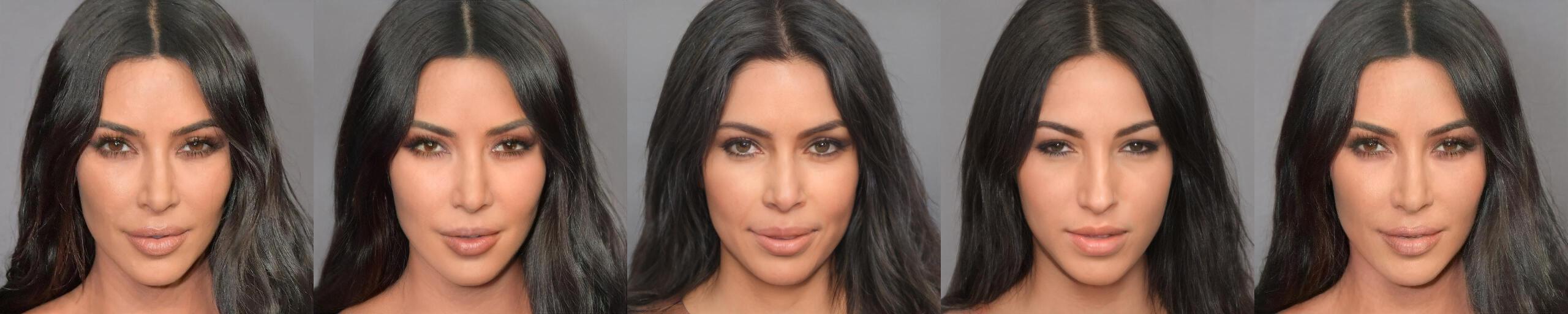}} \\
\noalign{\vskip 1mm}
\raisebox{-.5\totalheight}{\includegraphics[width=\celebaSize\textwidth]{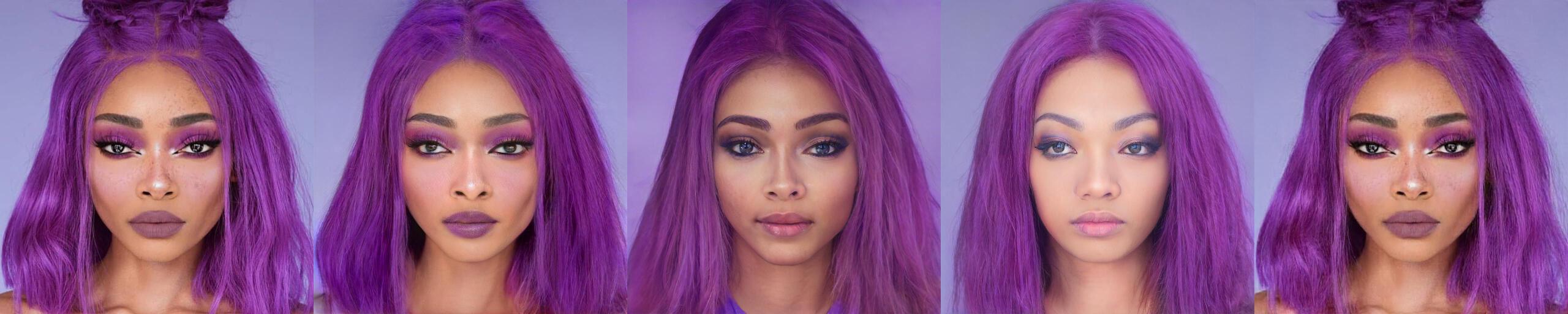}} \\
\noalign{\vskip 1mm}
\raisebox{-.5\totalheight}{\includegraphics[width=\celebaSize\textwidth]{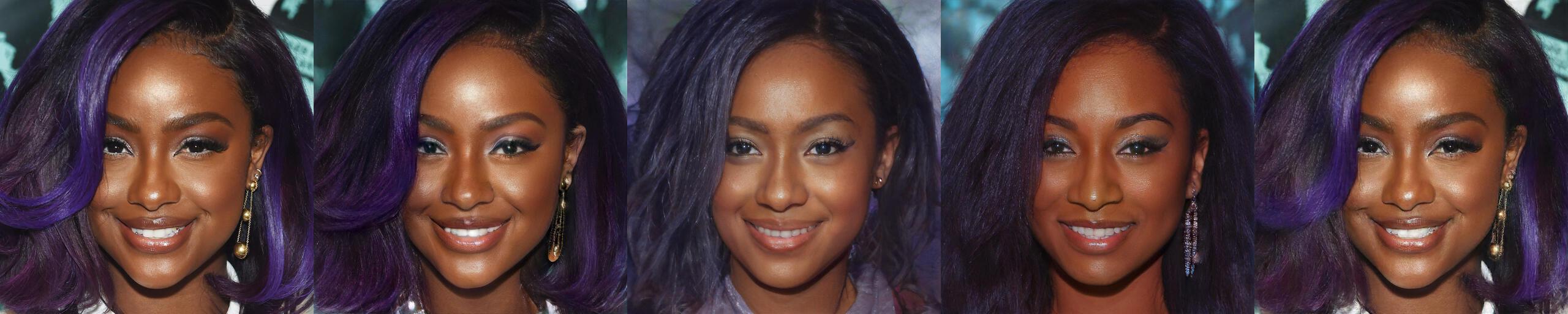}} \\
\noalign{\vskip 1mm}
\raisebox{-.5\totalheight}{\includegraphics[width=\celebaSize\textwidth]{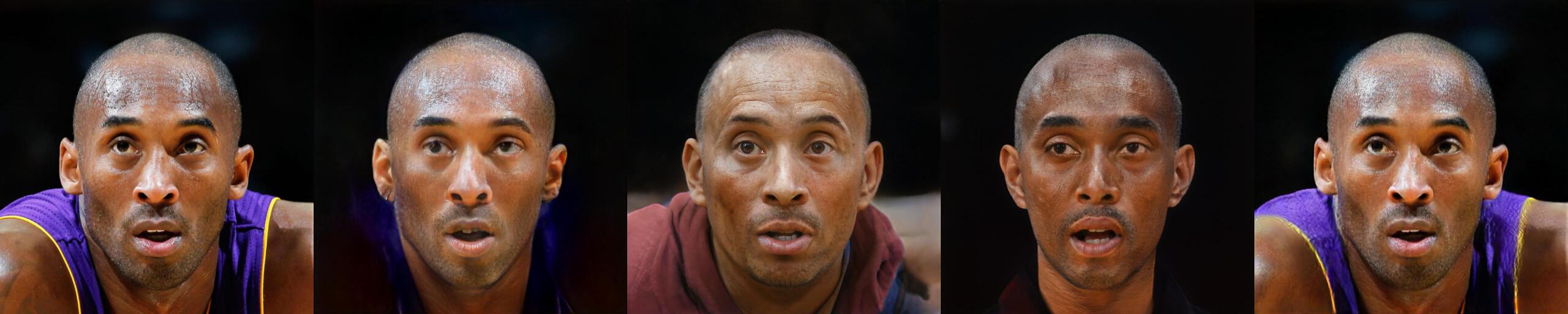}} \\
\noalign{\vskip 1mm}
\raisebox{-.5\totalheight}{\includegraphics[width=\celebaSize\textwidth]{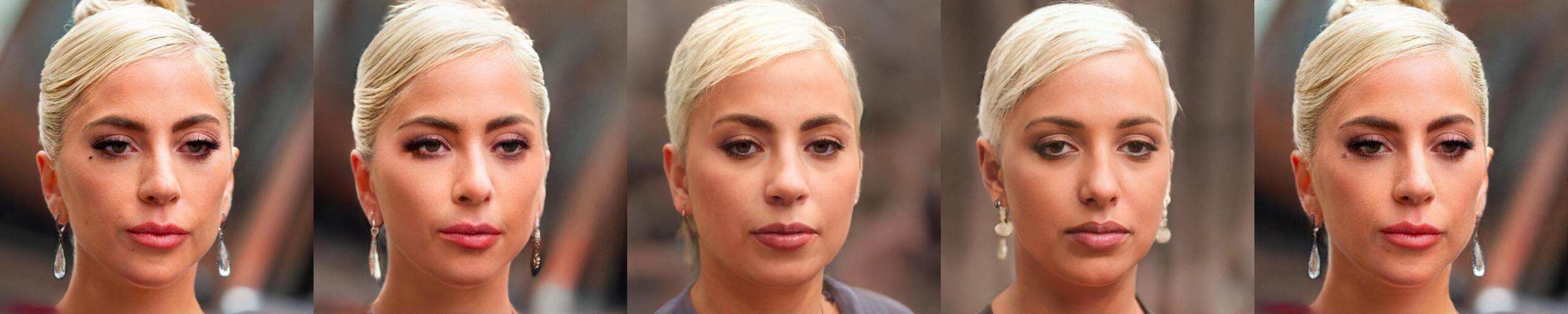}} \\

\end{tabular}
\vspace{0.15cm}
\caption{Additional reconstruction results over real images. Our method preserves out-of-distribution details, such as earrings or complicated make-up.}
\label{fig:ood_rec2}
\end{figure*}

\end{document}